\documentclass[letterpaper]{article} 
\usepackage{aaai24}  
\usepackage{times}  
\usepackage{helvet}  
\usepackage{courier}  
\usepackage[hyphens]{url}  
\usepackage{graphicx} 
\usepackage{natbib}  
\usepackage{caption} 
\usepackage[dvipsnames]{xcolor}
\usepackage{paralist}
\usepackage{enumitem}
\usepackage{multirow}
\usepackage{amsmath}
\usepackage{amsthm}
\usepackage{amssymb}
\usepackage{graphicx}
\usepackage{subcaption}
\usepackage{comment}
\usepackage{nicefrac}
\usepackage{booktabs}
\newtheorem{prop}{Proposition}

\allowdisplaybreaks

\usepackage{amsmath,amsfonts,bm}

\def\eqref#1{equation~\ref{#1}}

\def\1{\bm{1}}

\DeclareMathAlphabet{\mathsfit}{\encodingdefault}{\sfdefault}{m}{sl}
\SetMathAlphabet{\mathsfit}{bold}{\encodingdefault}{\sfdefault}{bx}{n}

\def\ba{\mathbf{a}}

\newcommand{\norm}[1]{\left\|#1\right\|}

\newcommand{\size}[1]{\left|#1\right|}

\title{Rethinking Robustness of Model Attributions}
\author {
    Sandesh Kamath\textsuperscript{\rm 1},
    Sankalp Mittal\textsuperscript{\rm 1},
    Amit Deshpande\textsuperscript{\rm 2},
    Vineeth N Balasubramanian\textsuperscript{\rm 1}
}
\affiliations {
    \textsuperscript{\rm 1}Indian Institute of Technology, Hyderabad\\
    \textsuperscript{\rm 2}Microsoft Research, Bengaluru\\
    sandesh.kamath@gmail.com,
}

\begin{document}

\maketitle

\begin{abstract}
For machine learning models to be reliable and trustworthy, their decisions must be interpretable. As these models find increasing use in safety-critical applications, it is important that not just the model predictions but also their explanations (as feature attributions) be robust to small human-imperceptible input perturbations. Recent works have shown that many attribution methods are fragile and have proposed improvements in either these methods or the model training. We observe two main causes for fragile attributions: first, the existing metrics of robustness (e.g., top-$k$ intersection) overpenalize even reasonable local shifts in attribution, thereby making random perturbations to appear as a strong attack, and second, the attribution can be concentrated in a small region even when there are multiple important parts in an image. To rectify this, we propose simple ways to strengthen existing metrics and attribution methods that incorporate locality of pixels in robustness metrics and diversity of pixel locations in attributions. Towards the role of model training in attributional robustness, we empirically observe that adversarially trained models have more robust attributions on smaller datasets, however, this advantage disappears in larger datasets. Code is made available\footnote{https://github.com/ksandeshk/LENS}.

\end{abstract}

\section{Introduction}
\label{sec_intro}
The explosive increase in the use of deep neural network (DNN)-based models for applications across domains has resulted in a very strong need to find ways to interpret the decisions made by these models \citep{GadeGKMT20,Tang2021shap,Yap2021rna,Oviedo2022,OhJ20a}. Interpretability is an important aspect of responsible and trustworthy AI, and model explanation methods (also known as attribution methods) are an important aspect of the community's efforts towards explaining and debugging real-world AI/ML systems. 
Attribution methods \citep{zeiler2015deconv,simonyan2014deep,bach2015lrp,Selvaraju_2019,Chattopadhay_2018,SundararajanTY17,shrikumar2017just,smilkov2017smoothgrad,lundberg2017unified} attempt to explain the decisions made by DNN models through input-output attributions or saliency maps. \citep{lipton2018mythos,samek_explainable_2019,fan_interpretability_2021,zhang_survey_2020} present detailed surveys on these methods. 
Recently, the growing numbers of attribution methods has led to a concerted focus on studying the robustness of attributions to input perturbations to handle potential security hazards \citep{chen2019robust,sarkar2020enhanced, Wang22Kendall, agarwal2022rethinking}. One could view these efforts as akin to adversarial robustness that focuses on defending against attacks on model predictions, whereas \textit{attributional robustness} focuses on defending against attacks on model explanations.
For example, an explanation for a predicted credit card failure cannot change significantly for a small human-imperceptible change in input features, or the saliency maps explaining the COVID risk prediction from a chest X-ray should not change significantly with a minor human-imperceptible change in the image. 

\begin{figure}[!ht]
  \begin{center}
    \includegraphics[width=0.15\textwidth]{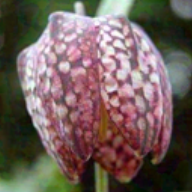}
    \includegraphics[width=0.15\textwidth]{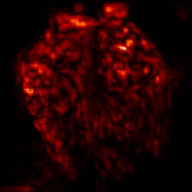}
    \includegraphics[width=0.15\textwidth]{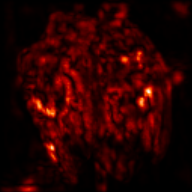}\\    
    \includegraphics[width=0.15\textwidth]{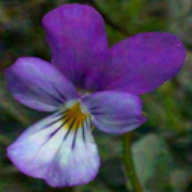}
    \includegraphics[width=0.15\textwidth]{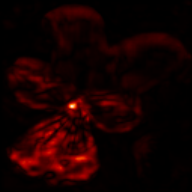}
    \includegraphics[width=0.15\textwidth]{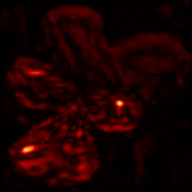}\\
    \includegraphics[width=0.15\textwidth]{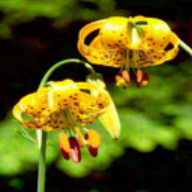}
    \includegraphics[width=0.15\textwidth]{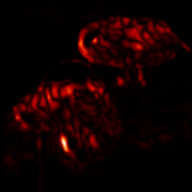}
    \includegraphics[width=0.15\textwidth]{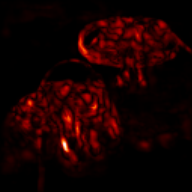}
  \end{center}
  \caption{Sample images from Flower dataset with Integrated Gradients (IG) before and after attributional attack. The attack used here is the top-$k$ attributional attack of \citet{ghorbani2018interpretation} on a ResNet model. Robustness of attribution measured by top-$k$ intersection is small, and ranges from 0.04 (first image) to 0.45 (third image) as it penalizes for both local changes in attribution and concentration of top pixels in a small region. Visually, we can observe that such overpenalization leads to a wrong sense of robustness as the changes are within the object of importance.}  
  \label{fig_intro}
\end{figure}

DNN-based models are known to have a vulnerability to imperceptible adversarial perturbations \citep{Biggio_2013,szegedy2014intriguing,goodfellow15fgsm}, which make them misclassify input images. 
Adversarial training \citep{madry2019deep} is widely understood to provide a reasonable degree of robustness to such perturbation attacks. 
While adversarial robustness has received significant attention over the last few years \citep{ozdag2018advsurvey,silva2020advrobustsurvey}, the need for stable and robust attributions, corresponding explanation methods and their awareness are still in their early stages at this time \citep{ghorbani2018interpretation,chen2019robust,slack2020fooling,sarkar2020enhanced,lakkaraju2020icml,slack2021neurips-a,slack2021neurips-b}. In an early effort, \citep{ghorbani2018interpretation} provided a method to construct a small imperceptible perturbation which when added to an input $x$ results in a change in attribution map of the original map to that of the perturbed image. This is measured through top-$k$ intersection, Spearman's rank-order correlation or Kendall's rank-order correlation between the two attribution maps (of original and perturbed images). See Figure \ref{fig_intro} for an example. Defenses proposed against such attributional attacks 
\citep{chen2019robust,singh2020attributional,wang2020smoothed,sarkar2020enhanced}
also leverage the same metrics to evaluate the robustness of attribution methods.

While these efforts have showcased the need and importance of studying the robustness of attribution methods, we note in this work that the metrics used, and hence the methods, can be highly sensitive to minor local changes in attributions (see Fig \ref{fig_intro} \textit{row 2}).
We, in fact, show (in Appendix \ref{app:subsec:rand-vec}) that under existing metrics to evaluate robustness of attributions, a random perturbation can be as strong an attributional attack as existing benchmark methods. This may not be a true indicator of the robustness of a model's attributions, and can mislead further research efforts in the community. We hence focus our efforts in this work on rethinking metrics and methods to study the robustness of model attributions (in particular, we study image-based attribution methods to have a focused discussion and analysis). Beyond highlighting this important issue, we propose locality-sensitive improvements of the above metrics that incorporate the locality of attributions along with their rank order. We show that such a locality-sensitive distance is upper-bounded by a metric based on symmetric set difference. We also introduce a new measure {\bf top-$k$-div} that incorporates diversity of a model's attributions. 
Our key contributions are summarized below:
\begin{itemize}[leftmargin=*]
\setlength\itemsep{-0.06em}
\item Firstly, we observe that existing robustness metrics for model attributions overpenalize minor drifts in attribution, leading to a false sense of fragility. 
\item In order to address this issue, we propose Locality-sENSitive (LENS) improvements of existing metrics, namely, LENS-top-$k$, LENS-Spearman and LENS-Kendall, that incorporate the locality of attributions along with their rank order. Besides avoiding overpenalizing attribution methods for minor local drifts, we show that our proposed LENS variants are well-motivated by metrics defined on the space of attributions. 
\item We subsequently introduce a second measure based on diversity that enriches model attributions by preventing the localized grouping of top model attributions. LENS can be naturally applied to this measure, thereby giving a method to incorporate both diversity and locality in measuring attributional robustness. 
\item Our comprehensive empirical results on benchmark datasets and models used in existing work clearly support our aforementioned observations, as well as the need to rethink the evaluation of the robustness of model attributions using locality and diversity. 
\item Finally, we also show that existing methods for robust attributions implicitly support such a locality-sensitive metric for evaluating progress in the field. 
\end{itemize}
%
\section{Background and Related Work}
\label{sec_related_work}
We herein discuss background literature from three different perspectives that may be related to our work: model explanation/attribution methods, efforts on attributional robustness (both attacks and defenses), and other recent related work.
\noindent \textbf{Attribution Methods.} Existing efforts on explainability in DNN models can be broadly categorized as: local and global methods, model-agnostic and model-specific methods, or as post-hoc and ante-hoc (intrinsically interpretable) methods \citep{molnar2019,xaitutorial}. Most existing methods in use today -- including methods to visualize weights and neurons \citep{simonyan2014deep,ZeilerF14}, guided backpropagation \citep{springenberg2015striving}, CAM \citep{ZhouKLOT16}, GradCAM \citep{Selvaraju_2019}, Grad-CAM++ \citep{Chattopadhay_2018}, LIME \citep{ribeiro2016why}, DeepLIFT \citep{shrikumar2017just,shrikumar2019learning}, LRP \citep{bach2015lrp}, Integrated Gradients \citep{SundararajanTY17}, SmoothGrad \citep{smilkov2017smoothgrad}), DeepSHAP \citep{lundberg2017unified} and TCAV \citep{kim2018interpretability} -- are post-hoc methods, which are used on top of a pre-trained DNN model to explain its predictions. We focus on such post-hoc attribution methods in this work. For a more detailed survey of explainability methods for DNN models, please see \citep{xaitutorial,molnar2019,samek_explainable_2019}. 
%

\noindent \textbf{Robustness of Attributions.} The growing numbers of attribution methods proposed has also led to efforts on identifying the desirable characteristics of such methods \citep{alvarez2018robustness,AdebayoGMGHK18,yeh2019fidelity,ChalasaniC00J20,TomsettHCGP20,Boggust22Faithful,agarwal2022rethinking}. A key desired trait that has been highlighted by many of these efforts is robustness or stability of attributions, i.e., the explanation should not vary significantly within a small local neighborhood of the input \citep{alvarez2018robustness,ChalasaniC00J20}. \citet{ghorbani2018interpretation} showed that well-known methods such as gradient-based attributions, DeepLIFT \citep{shrikumar2019learning} and Integrated Gradients (IG) \citep{SundararajanTY17} are vulnerable to such input perturbations, and also provided an algorithm to construct a small imperceptible perturbation which when added to the input results in changes in the attribution.  \citet{slack2020fooling} later showed that methods like LIME \citep{ribeiro2016why} and DeepSHAP \citep{lundberg2017unified} are also vulnerable to such manipulations. The identification of such vulnerability and potential for attributional attacks has since led to multiple research efforts to make a model's attributions robust. \citet{chen2019robust} proposed a regularization-based approach, where an explicit regularizer term is added to the loss function to maintain the model gradient across input (IG, in particular) while training the DNN model. This was subsequently extended by \citep{sarkar2020enhanced,singh2020attributional,wang2020smoothed}, all of whom provide different training strategies and regularizers to improve attributional robustness of models. Each of these methods including \citet{ghorbani2018interpretation} measures change in attribution before and after input perturbation using the same metrics: top-$k$ intersection, and/or rank correlations like Spearman's $\rho$ and Kendall' $\tau$. Such metrics have recently, in fact, further been used to understand issues surrounding attributional robustness \citep{Wang22Kendall}. Other efforts that quantify stability of attributions in tabular data also use Euclidean distance (or its variants) between the original and perturbed attribution maps  \citep{alvarez2018robustness,yeh2019fidelity,agarwal2022rethinking}. Each of these metrics look for dimension-wise correlation or pixel-level matching between attribution maps before and after perturbation, and thus penalize even a minor change in attribution (say, even by one pixel coordinate location). This results in a false sense of fragility, and could even be misleading. In this work, we highlight the need to revisit such metrics, and propose variants based on locality and diversity that can be easily integrated into existing metrics. 

\noindent \textbf{Other Related Work.} In other related efforts that have studied similar properties of attribution-based explanations, \citep{carvalho2019machine,bhatt2020evaluating} stated that stable explanations should not vary too much between similar input samples, unless the model's prediction changes drastically. The abovementioned attributional attacks and defense methods \citep{ghorbani2018interpretation,sarkar2020enhanced,singh2020attributional,wang2020smoothed} maintain this property, since they focus on input perturbations that change the attribution without changing the model prediction itself. Similarly, \citet{arun2020assessing} and \citet{fel2022good} introduced the notions of repeatability/reproducibility and generalizability respectively, both of which focus on the desired property that a trustworthy explanation must point to similar evidence across similar input images. In this work, we provide a practical metric to study this notion of similarity by considering locality-sensitive metrics.
\section{{\bf L}ocality-s{\bf ENS}itive Metrics (LENS) for Attributional Robustness} \label{sec:lens}
 \begin{figure}[h]
 \centering
     \includegraphics[width=0.22\textwidth]{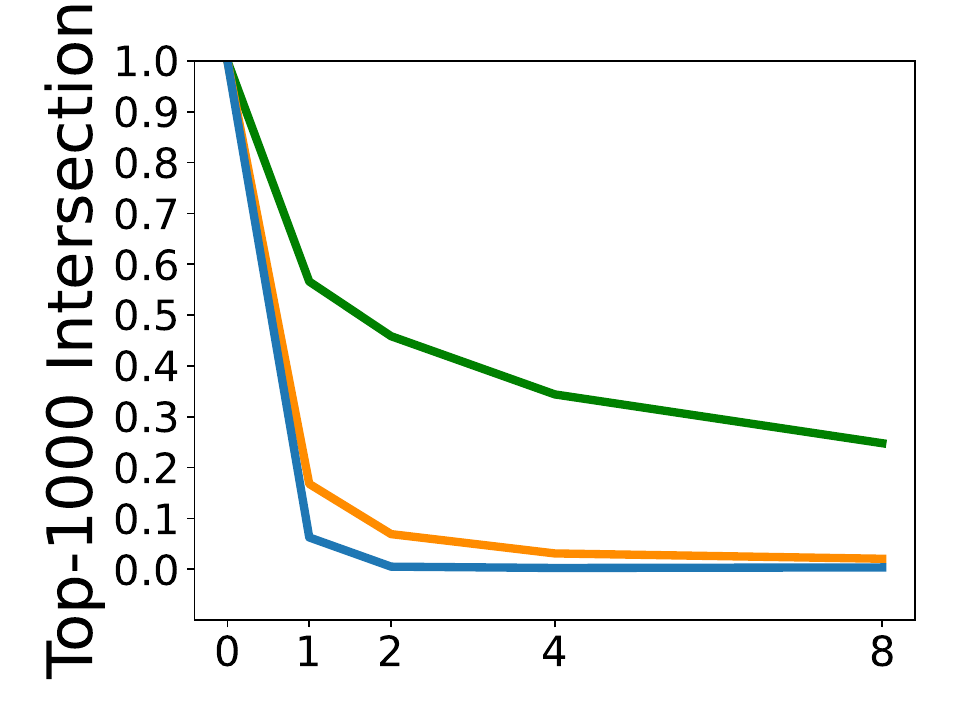}
     \includegraphics[width=0.22\textwidth]{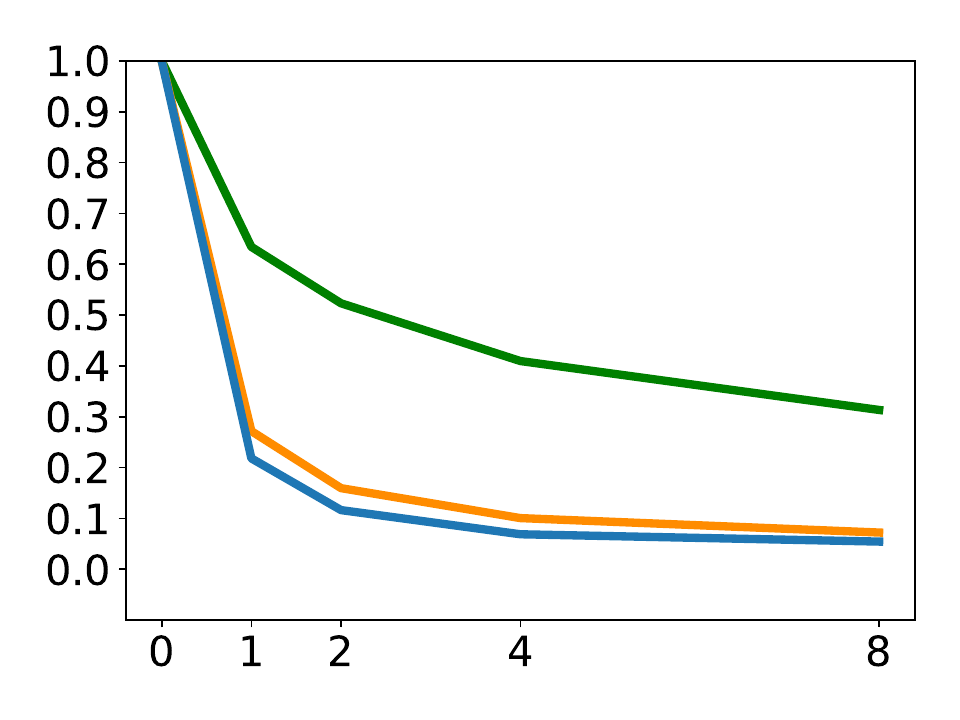}\\
     \includegraphics[width=0.22\textwidth]{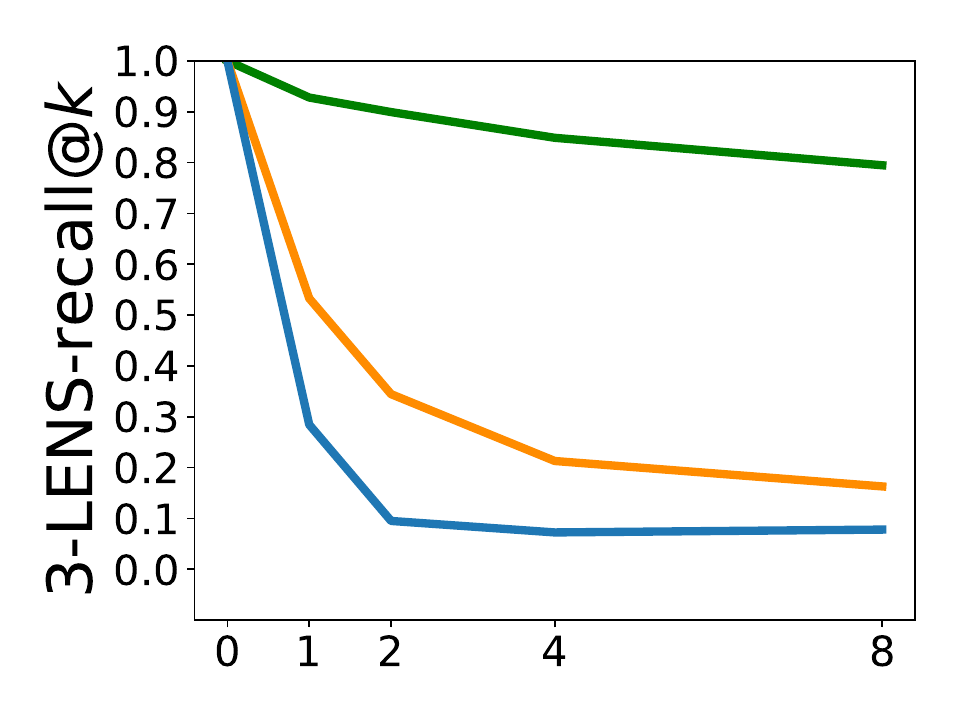}
     \includegraphics[width=0.22\textwidth]{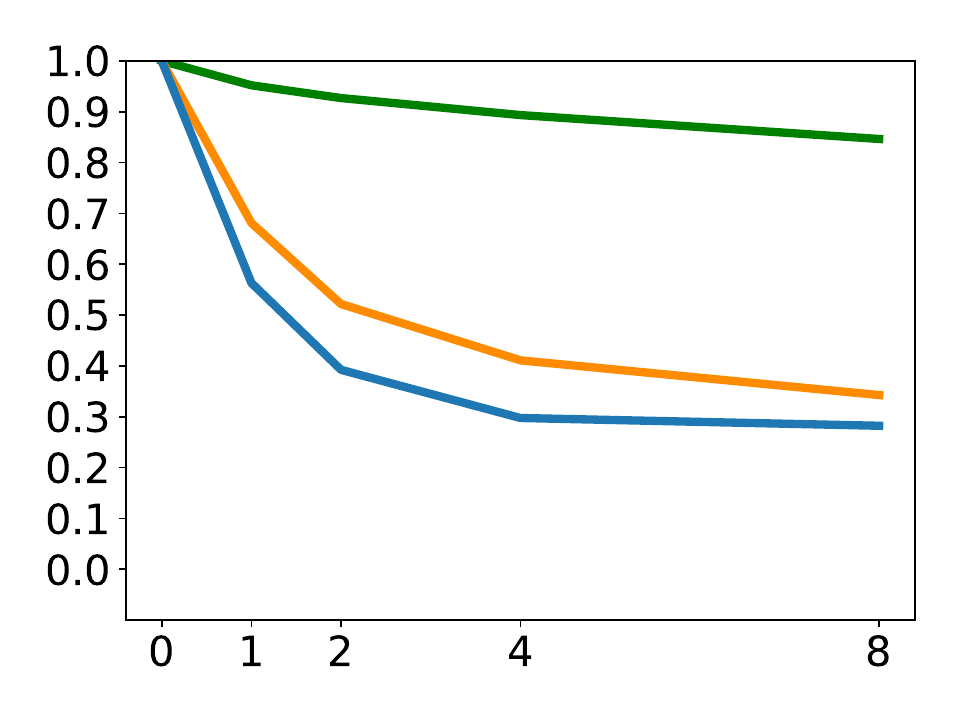}\\         
     \includegraphics[width=0.22\textwidth]{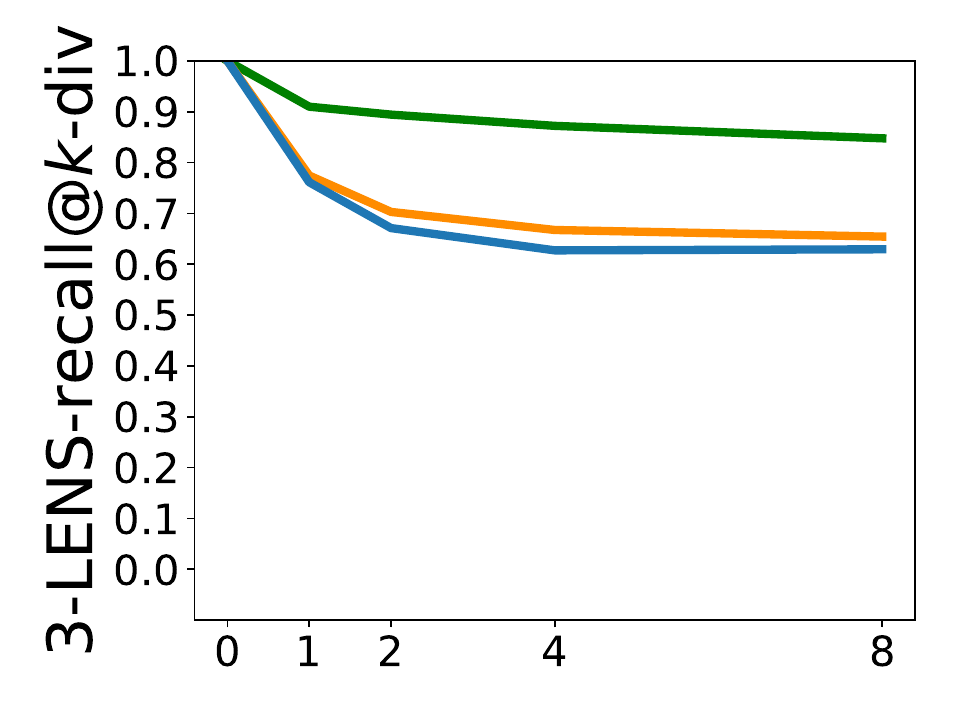}
     \includegraphics[width=0.22\textwidth]{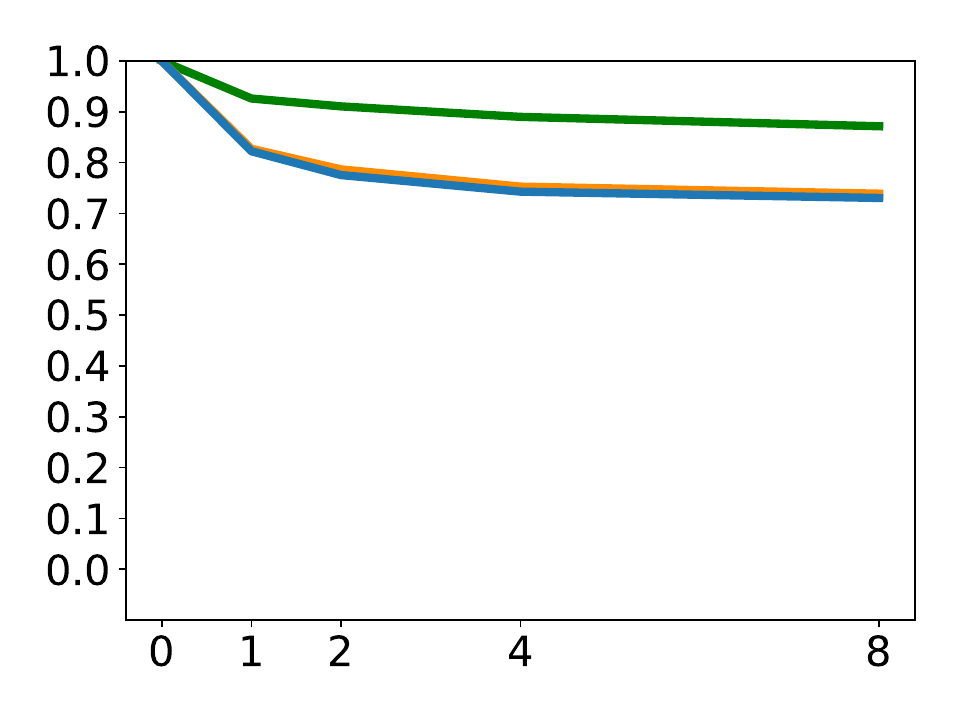}    
    \caption{From top to bottom, we plot average top-$k$ intersection (currently used metric), $3$-LENS-recall@$k$ and $3$-LENS-recall@$k$-div (proposed metrics) against the $\ell_{\infty}$-norm of  attributional attack perturbations for Simple Gradients (SG) (left) and Integrated Gradients (IG) (right) of a SqueezeNet model on Imagenet. We use $k=1000$ and three attributional attack variants proposed by \citet{ghorbani2018interpretation}. Evidently, the proposed metrics show more robustness under the same attacks.}
\label{imagenet-ghorbani-with-lens-div}     
\end{figure}

As a motivating example, Figure \ref{imagenet-ghorbani-with-lens-div} presents the results obtained using \cite{ghorbani2018interpretation} with Simple Gradients (SG) and Integrated Gradients (IG) of an NN model trained on ImageNet. The top row, which reports the currently followed top-$k$ intersection measure of attribution robustness, shows a significant drop in robustness performance even for the random sign attack (green line). The subsequent rows, which report our metrics for the same experiments, show significant improvements in robustness -- especially when combining the notions of locality and diversity. Observations made on current metrics could lead to a false sense of fragility, which overpenalizes even an attribution shift by 1-2 pixels. A detailed description of our experimental setup for these results is available in Appendix \ref{app:exp-setup}. Motivated by these observations, we explore improved measures for attributional robustness that maintain the overall requirements of robustness, but do not overpenalize minor deviations.
\begin{figure}[!ht]
  \begin{center}
    \includegraphics[width=0.11\textwidth]{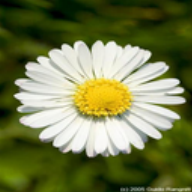}
    \includegraphics[width=0.11\textwidth]{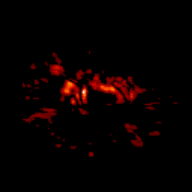}
    \includegraphics[width=0.11\textwidth]{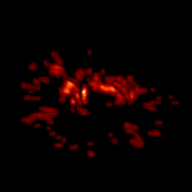}
    \includegraphics[width=0.11\textwidth]{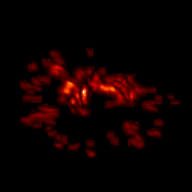}\\
    \includegraphics[width=0.11\textwidth]{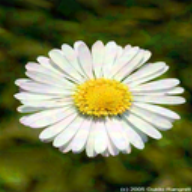}
    \includegraphics[width=0.11\textwidth]{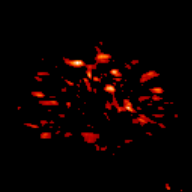}
    \includegraphics[width=0.11\textwidth]{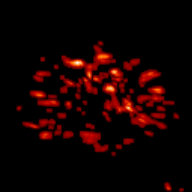}
    \includegraphics[width=0.11\textwidth]{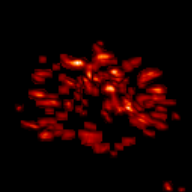}\\
  \end{center}
      \caption{A sample image from Flower dataset before (top) and after (bottom) the top-$k$ attributional attack of \cite{ghorbani2018interpretation} on a ResNet model for Integrated Gradients (IG) attribution method. From left to right: the image, its top-$k$ pixels as per IG, the union of the $3 \times 3$-pixel neighborhoods and $5 \times 5$-pixel neighborhoods of the top-$k$ pixels, respectively, for $k=1000$. Quantitatively, top-$k$ intersection: 0.14, 1-LENS-recall@$k$: 0.25, 1-LENS-pre@$k$: 0.37, 2-LENS-recall@$k$: 0.40, 2-LENS-pre@$k$: 0.62.} 
    \label{fig_intro_lens-19}
\end{figure}
%
%
\subsection{Defining LENS Metrics for Attributions} \label{subsec:define-lens}
To begin with, we propose an extension of existing similarity measures to incorporate the locality of pixel attributions in images to derive more practical and useful measures of attributional robustness. Let $a_{ij}(x)$ denote the attribution value or importance assigned to the $(i, j)$-th pixel in an input image $x$, and let $S_{k}(x)$ denote the set of $k$ pixel positions with the largest attribution values. Let $N_{w}(i, j) = \{(p, q) \;:\; i-w \leq p \leq i+w,~ j-w \leq q \leq j+w\}$ be the neighboring pixel positions within a $(2w+1) \times (2w+1)$ window around the $(i, j)$-th pixel. By a slight abuse of notation, we use $N_{w}(S_{k}(x))$ to denote $\bigcup_{(i, j) \in S_{k}(x)} N_{w}(i, j)$, that is, the set of all pixel positions that lie in the union of $(2w+1) \times (2w+1)$ windows around the top-$k$ pixels. 

For a given attributional perturbation $\mathrm{Att}(\cdot)$, let $T_{k} = S_{k}(x + \mathrm{Att}(x))$ denote the top-$k$ pixels in attribution values after applying the attributional perturbation $\mathrm{Att}(x)$. The currently used top-$k$ intersection metric is then computed as: $\size{S_{k}(x) \cap T_{k}(x)}/k$. To address the abovementioned issues, we instead propose \emph{Locality-sENSitive top-$k$} metrics (LENS-top-$k$) as $\size{N_{w}(S_{k}(x)) \cap T_{k}(x)}/k$ and $\size{S_{k}(x) \cap N_{w}(T_{k}(x))}/k$, which are also closer to more widely used metrics such as precision and recall in ranking methods. We simiarly define Locality-sENSitive Spearman's $\rho$ (LENS-Spearman) and Locality-sENSitive Kendall's $\tau$ (LENS-Kendall) metrics as rank correlation coefficients for the smoothed ranking orders according to $\tilde{a}_{ij}(x)$'s and $\tilde{a}_{ij}(x + \mathrm{Att}(x))$'s, respectively. These can be used to compare two different attributions for the same image, the same attribution method on two different images, or even two different attributions on two different images, as long as the attribution vectors lie in the same space, e.g., images of the same dimensions where attributions assign importance values to pixels. Figure \ref{fig_intro_lens-19} provides the visualization of the explanation map of a sample from the Flower dataset with the top-1000 pixels followed by the corresponding maps with 1-LENS@$k$ and 2-LENS@$k$.

We show that the proposed locality-sensitive variants of the robustness metrics also possess some theoretically interesting properties. Let $\ba_{1}$ and $\ba_{2}$ be two attribution vectors for two images, and let $S_{k}$ and $T_{k}$ be the set of top $k$ pixels in these images according to $\ba_{1}$ and $\ba_{2}$, respectively. We define a locality-sensitive top-$k$ distance between two attribution vectors $\ba_{1}$ and $\ba_{2}$ as $d_{k}^{(w)}(\ba_{1}, \ba_{2})\overset{\text{def}}{=} \text{prec}_{k}^{(w)}(\ba_{1}, \ba_{2}) + \text{recall}_{k}^{(w)}(\ba_{1}, \ba_{2})$, where
$
\text{prec}_{k}^{(w)}(\ba_{1}, \ba_{2}) \overset{\text{def}}{=} \frac{\size{S_{k} \setminus N_{w}(T_{k})}}{k} \quad \text{and} 
$
$
\quad \text{recall}_{k}^{(w)}(\ba_{1}, \ba_{2}) \overset{\text{def}}{=} \frac{\size{T_{k} \setminus N_{w}(S_{k})}}{k},
$
similar to precision and recall used in ranking literature, with the key difference being the inclusion of neighborhood items based on locality. Below we state a monotonicity property of $d_{k}^{(w)}(\ba_{1}, \ba_{2})$ and upper bound it in terms of the symmetric set difference of top-$k$ attributions. 
\begin{prop} \label{prop:monotone}
For any $w_{1} \leq w_{2}$, we have $d_{k}^{(w_{2})}(\ba_{1}, \ba_{2}) \leq d_{k}^{(w_{1})}(\ba_{1}, \ba_{2}) \leq \size{S_{k} \triangle T_{k}}/k$, where $\triangle$ denotes the symmetric set difference, i.e., $A \triangle B = (A \setminus B) \cup (B \setminus A)$.
\end{prop}
Combining $d_{k}^{(w)}(\ba_{1}, \ba_{2})$ across different values of $k$ and $w$, we can define a distance 
\[
d(\ba_{1}, \ba_{2}) = \sum_{k=1}^{\infty} \alpha_{k} \sum_{w=0}^{\infty} \beta_{w}~ d_{k}^{(w)}(\ba_{1}, \ba_{2}),
\] 
where $\alpha_{k}$ and $\beta_{w}$ be non-negative weights, monotonically decreasing in $k$ and $w$, respectively, such that $\sum_{k} \alpha_{k} < \infty$ and $\sum_{w} \beta_{w} < \infty$. We show that the distance defined above is upper-bounded by a metric similar to those proposed in \cite{fagin2003comparing} based on symmetric set difference of top-$k$ ranks to compare two rankings.
\begin{prop} \label{prop:metric}
$d(\ba_{1}, \ba_{2})$ defined above is upper-bounded by $u(\ba_{1}, \ba_{2})$ given by
\[
u(\ba_{1}, \ba_{2}) = \sum_{k=1}^{\infty} \alpha_{k} \sum_{w=0}^{\infty} \beta_{w}~ \frac{\size{S_{k} \triangle T_{k}}}{k},
\]
and $u(\ba_{1}, \ba_{2})$ defines a bounded metric on the space of attribution vectors.
\end{prop}
Note that top-$k$ intersection, Spearman's $\rho$ and Kendall's $\tau$ do not take the attribution values $a_{ij}(x)$'s into account but only the rank order of pixels according to these values. We also define a locality-sensitive $w$-smoothed attribution as follows.
\[
\tilde{a}_{ij}^{(w)}(x) = \frac{1}{(2w+1)^{2}} \sum_{\substack{(p, q) \in N_{w}(i, j), \\ 1 \leq p, q \leq n}} a_{pq}(x)
\]
We show that the $w$-smoothed attribution leads to a contraction in the $\ell_{2}$ norm commonly used in theoretical analysis of simple gradients as attributions. 
\begin{prop} \label{prop:w-smoothed}
For any inputs $x, y$ and any $w \geq 0$, $\norm{\tilde{\ba}^{(w)}(x) - \tilde{\ba}^{(w)}(y)}_{2} \leq \norm{\ba(x) - \ba(y)}_{2}$.
\end{prop}
Thus, any theoretical bounds on the attributional robustness of simple gradients in $\ell_{2}$ norm proved in previous works continue to hold for locality-sensitive $w$-smoothed gradients. For example, \cite{wang2020smoothed} show the following Hessian-based bound on simple gradients. For an input $x$ and a classifier or model defined by $f$, let $\nabla_{x}(f)$ and $\nabla_{y}(f)$ be the simple gradients w.r.t. the inputs at $x$ and $y$. Theorem 3 in \cite{wang2020smoothed} upper bounds the $\ell_{2}$ distance between the simple gradients of nearby points $\norm{x - y}_{2} \leq \delta$ as  $\norm{\nabla_{x}(f) - \nabla_{y}(f)}_{2} \lesssim \delta~ \lambda_{\text{max}}(H_{x}(f))$, where $H_{x}(f)$ is the Hessian of $f$ w.r.t. the input at $x$ and $\lambda_{\text{max}}(H_{x}(f))$ is its maximum eigenvalue. By Proposition \ref{prop:w-smoothed} above, the same continues to hold for $w$-smoothed gradients, i.e., $\norm{\tilde{\nabla}_{x}^{(w)}(f) - \tilde{\nabla}_{y}^{(w)}(f)}_{2} \lesssim \delta~ \lambda_{\text{max}}(H_{x}(f))$. The proofs of all the propositions above are included in Appendix \ref{app:proofs}.

\subsection{Relevance to Attributional Robustness} \label{subsec:lens-relevance-robustness}
The top-$k$ intersection is a measure of similarity instead of distance. Therefore, in our experiments for attributional robustness, we use locality-sensitive similarity measures $w\text{-LENS-prec@}k$ and $w\text{-LENS-recall@}k$ to denote $1 - \text{prec}_{k}^{(w)}(\ba_{1}, \ba_{2})$ and $1 - \text{recall}_{k}^{(w)}(\ba_{1}, \ba_{2})$, respectively, where $\ba_{1}$ is the attribution of the original image and $\ba_{2}$ is the attribution of the perturbed image. For rank correlation coefficients such as Kendall's $\tau$ and Spearman's $\rho$, we compute $w$-LENS-Kendall and $w$-LENS-Spearman as the same Kendall's $\tau$ and Spearman's $\rho$ but computed on the locality-sensitive $w$-smoothed attribution map $\tilde{\ba}^{(w)}$ instead of the original attribution map $\ba$. We also study how these similarity measures and their resulting attributional robustness measures change as we vary $w$.
%
In this section, we measure the attributional robustness of Integrated Gradients (IG) on naturally trained models as top-$k$ intersection, $w$-LENS-prec@$k$ and $w$-LENS-recall@$k$ between the IG of the original images and the IG of their perturbations obtained by various attacks. The attacks we consider are the top-$t$ attack and the mass-center attack of \citet{ghorbani2018interpretation} as well as random perturbation. All perturbations have $\ell_{\infty}$ norm bounded by $\delta=0.3$ for MNIST, $\delta=0.1$ for Fashion MNIST, and $\delta=8/255$ for GTSRB and Flower datasets. 

The values of $t$ used to construct top-$t$ attacks of \citet{ghorbani2018interpretation} are $t=200$ on MNIST, $t=100$ on Fashion MNIST and GTSRB, $t=1000$ on Flower. In the robustness evaluations for a fixed $k$, we use $k=100$ on MNIST, Fashion MNIST, GTSRB, and $k=1000$ on Flower.

\begin{figure}[!ht]
\centering
    \includegraphics[width=0.25\textwidth]{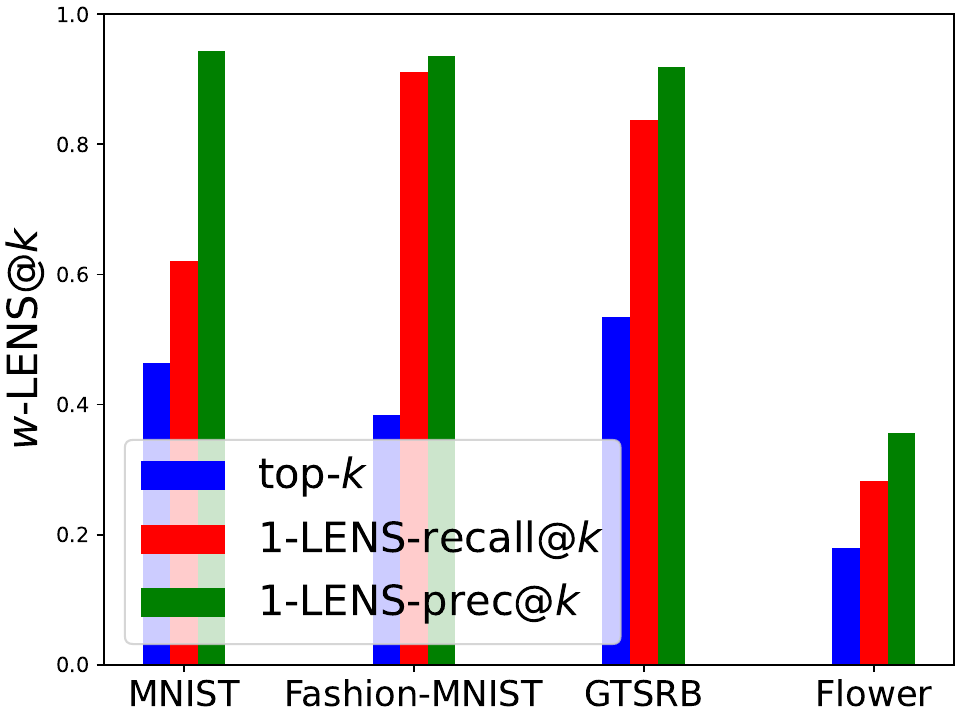}
\caption{Attributional robustness of IG on naturally trained models measured as average top-$k$ intersection, $1$-LENS-prec@$k$ and $1$-LENS-recall@$k$ between IG(original image) and IG(perturbed image) obtained by the top-$t$ attack \citep{ghorbani2018interpretation} across different datasets.} 
\label{dataset-topk-new-metric-st-nat}
\end{figure}

\paragraph{Comparison of top-$k$ intersection, $1$-LENS-prec@$k$ and $1$-LENS-recall@$k$.} 
Figure \ref{dataset-topk-new-metric-st-nat} shows that top-$k$ intersection penalizes IG even for small, local changes. $1$-LENS-prec@$k$ and $1$-LENS-recall@$k$ values are always higher in comparison across all datasets in our experiments. Moreover, on both MNIST and Fashion MNIST, $1$-LENS-prec@$k$ is roughly $2$x higher (above $90\%$) compared to top-$k$ intersection (near $40\%$). In other words, an attack may appear stronger under a weaker measure of attributional robustness, if it ignores locality. This increase clearly shows that the top-$k$ attack of \citet{ghorbani2018interpretation} appears to be weaker on these datasets as the proportional increase by using locality indicates that the attack is only creating a local change than previously thought. We can see that for MNIST, Fashion-MNIST and GTSRB for $<$ 20\% of the samples, the top-$k$ attack was able to make changes larger than what 1-LENS@$k$ could measure.

\begin{figure}[!ht]
\centering
\includegraphics[width=0.22\textwidth]{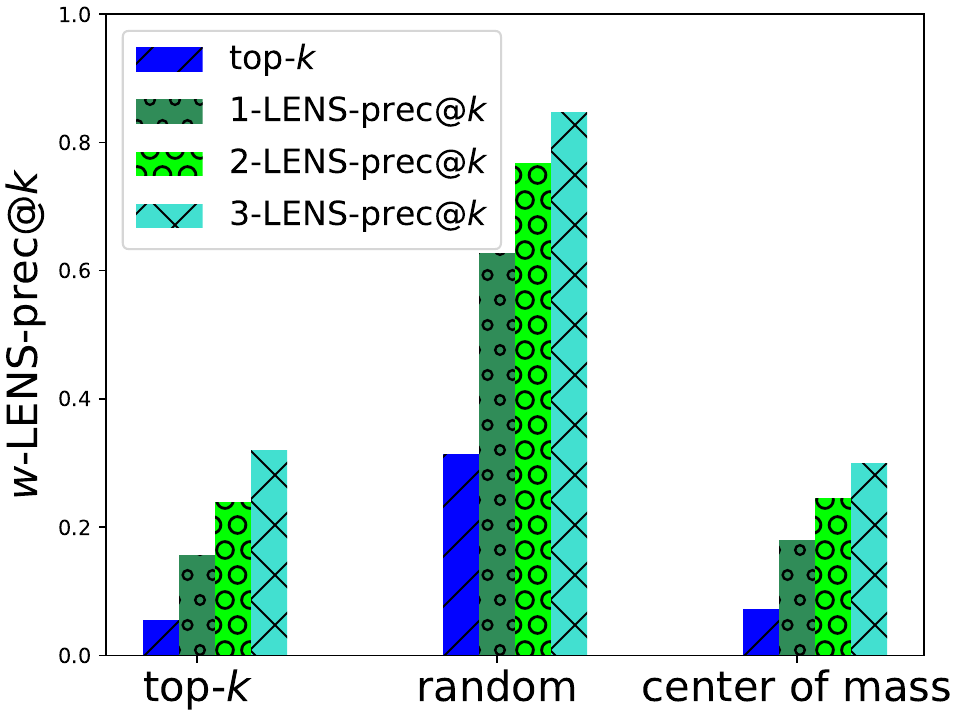}
\includegraphics[width=0.219\textwidth]{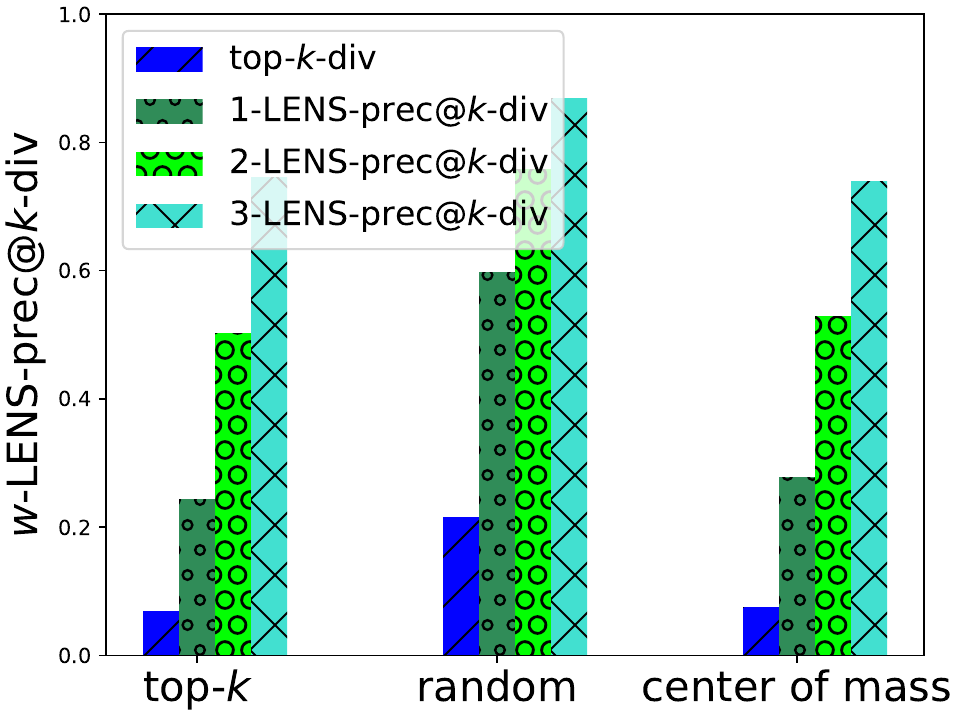}
\caption{Effect of increasing $w$ on average $w$-LENS-prec@$k$ and $w$-LENS-prec@$k$-div in comparison with top-$k$ intersection for IG map on ImageNet using a SqueezeNet model, when attacked with three attributional attacks (viz., top-$k$, random sign perturbation and mass center) of \citet{ghorbani2018interpretation}.}
\label{imagenet-topk-window-new-metric-nat-ig-lens} 
\end{figure}

\paragraph{$w$-LENS-prec@$k$ for varying $w$.} In Figure \ref{imagenet-topk-window-new-metric-nat-ig-lens}(left) $w$-LENS-prec@$k$ increases as we increase $w$ to consider larger neighborhoods around the pixels with top attribution values. This holds for multiple perturbations, namely, top-$t$ attack and mass-center attack by \citet{ghorbani2018interpretation} as well as a random perturbation. Notice that the top-$t$ attack of \citet{ghorbani2018interpretation} is constructed specifically for the top-$t$ intersection objective, and perhaps as a result, shows larger change when we increase local-sensitivity by increasing $w$ in the robustness measure. 

Due to space constraint and purposes of coherence, we present few results with IG here; we present similar results on other explanation methods in the Appendix \ref{app:more-lens-results}. Refer to Appendix \ref{app:all-ig-exp} for similar plots with random sign perturbation and mass center attack of \citet{ghorbani2018interpretation}. Appendix \ref{app:all-sg-exp} contains additional results with similar conclusions when Simple Gradients are used instead of Integrated Gradients (IG) for obtaining the attributions. 

As a natural follow-up question we present in Appendix \ref{app:vary-k-sp-ken} results obtained by modifying the similarity objective of top-$k$ attack of \citet{ghorbani2018interpretation} with $1$-LENS-prec@$k$ with the assumption to obtain a stronger attack. But surprisingly, we notice that it leads to a \emph{worse} attributional attack, if we measure its effectiveness using the top-$k$ intersections and $1$-LENS-prec@$k$. In other words, attributional attacks against locality-sensitive measures of attributional robustness are non-trivial and may require fundamentally different ideas.

\subsection{Alignment of attributional robustness metrics to human perception}
\label{subsec:survey-human}

We conducted a survey with human participants, where we presented images from the Flower dataset and a pair of attribution maps---an attribution map of the original image alongside an attribution map of their random perturbation or attributional attacked version \citet{ghorbani2018interpretation}, in a random order and without revealing this information to the participants. The survey participants were asked whether the two maps were relatable to the image and if one of them was different than the other. In Table \ref{table-survey-results} we summarize the results obtained from the survey. We simplify the choices presented to the user into 2 final categories - \begin{inparaenum}[(1)] \item Agree with w-LENS-prec@$k$ \item Agree with top-$k$ metric \end{inparaenum} Category (1) includes all results where the user found the maps the same, relatable to the image but dissimilar or the perturbed map was preferred over the original map. Category (2) was the case where the user preferred the original map over the perturbed map. Refer to Appendix \ref{app:survey-human} for more details.

\begin{table}[!ht]
\vspace{-9pt}
\begin{center}
\resizebox{0.47\textwidth}{!}{ 
\begin{tabular}{|c|c|}
\hline 
{\bf Agree with 3-LENS-prec@$k$ metric(\%)} & {\bf Agree with top-$k$ metric(\%)} \\
\hline
70.37 & 29.63 \\ 
\hline
\end{tabular}
}
\vspace{-6pt}
\end{center}
\caption{Survey results based on humans ability to relate the explanation map to the original image with or without noise using the Flower dataset.}
\label{table-survey-results}
\end{table}

\begin{figure}[!ht]
  \begin{center}
    \includegraphics[width=0.11\textwidth]{./images/flower_19_image.pdf}
    \includegraphics[width=0.11\textwidth]{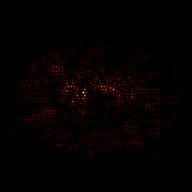}
    \includegraphics[width=0.11\textwidth]{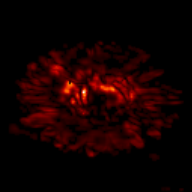}
    \includegraphics[width=0.11\textwidth]{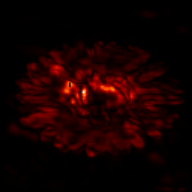}\\
    \includegraphics[width=0.11\textwidth]{./images/flower_19_image_pert.pdf}
    \includegraphics[width=0.11\textwidth]{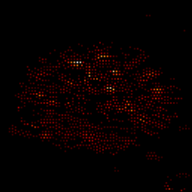}
    \includegraphics[width=0.11\textwidth]{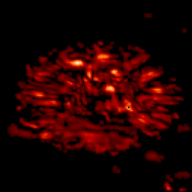}
    \includegraphics[width=0.11\textwidth]{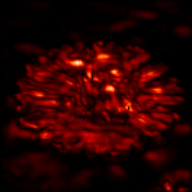}\\
  \end{center}
      \caption{A sample image from Flower dataset before (top) and after (bottom) the top-$k$ attributional attack of \citet{ghorbani2018interpretation} on a ResNet model. For both, we show from left to right: the image, its top-$k$ diverse pixels as per IG, the union of $3 \times 3$-pixel neighborhoods and $5 \times 5$-pixel neighborhoods of the top-$k$ diverse pixels, respectively, for $k=1000$. Quantitatively, improved overlap is captured by top-$k$-div intersection: 0.22, $1$-LENS-recall@$k$-div: 0.87, $1$-LENS-pre@$k$-div: 0.86, $2$-LENS-recall@$k$-div: 0.95, $2$-LENS-pre@$k$-div: 0.93. \emph{Zoom in required to see the diverse pixels.}}
  \label{fig_intro_div-19}
\end{figure}

\section{Diverse Attribution for Robustness} \label{sec:div-lens} 
Column 1 of Figure \ref{fig_intro_div-19} shows a typical  image from Flower dataset whose top-1000 pixels according to IG are concentrated in a small region. As seen in this illustrative example, when an image has multiple important parts, concentration of top attribution pixels in a small region increases vulnerability to attributional attacks. To alleviate this vulnerability, we propose post-processing any given attribution method to output top-$k$ \emph{diverse} pixels instead of just the top-$k$ pixels with the highest attribution scores. We use a natural notion of $w$-diversity based on pixel neighborhoods, so that these diverse pixels can be picked by a simple greedy algorithm. Starting with $S \leftarrow \emptyset$, repeat for $k$ steps: Pick the pixel of highest attribution score or importance outside $S$, add it to $S$ and disallow the $(2w+1) \times (2w+1)$-pixel neighborhood around it for future selection. The set of $k$ diverse pixels picked as above contains no two pixels within $(2w+1) \times (2w+1)$-pixel neighborhood of each other, and moreover, has the highest total importance (as the sum of pixel-wise attribution scores) among all such sets of $k$ pixels. The sets of $k$ pixels where no two pixels lie in $(2w+1) \times (2w+1)$-pixel neighborhood of each other form a matroid, where the optimality of greedy algorithm is well-known; see \citet{KorteLovasz1981}. 

Once we have the top-$k$ diverse pixels as described above, we can extend our locality-sensitive robustness metrics from the previous section to $w$-LENS-prec@$k$-div and $w$-LENS-recall@$k$-div, defined analogously using the union of $(2w+1) \times (2w+1)$-pixel neighborhoods of top-$k$ diverse pixels. In other words, define $\tilde{S}_{k}(x)$ as the top-$k$ diverse pixels for image $x$ and $\tilde{T}_{k} = \tilde{S}_{k}(x  + \text{Att}(x))$, and use $\tilde{S_{k}}$ and $\tilde{T}_{k}$ to replace $S_{k}$ and $T_{k}$ used in Subsection \ref{subsec:define-lens}. 

For $k=1000$, Figure \ref{fig_intro_div-19} shows a sample image from Flower dataset before and after the top-$k$ attributional attack of \citet{ghorbani2018interpretation}. Figure \ref{fig_intro_div-19} visually shows the top-$k$ diverse pixels in the Integrated Gradients (IG) and the union of their $(2w+1) \times (2w+1)$-pixel neighborhoods, for $w=\{1, 2\}$, for this image before and after the attributional attack. The reader may be required to zoom in to see the top-$k$ diverse pixels. See Appendix \ref{app:sample-fragility-examples} for more examples. Note that $0$-LENS-prec@$k$ and $0$-LENS-recall@$k$ are both the same and equivalent to top-$k$ intersection. However, a combined effect of locality and diversity can show a drastic leap from top-$k$ intersection value 0.14 to $2$-LENS-recall@$k$-div value 0.95 (see Fig.\ref{fig_intro_lens-19} and Fig.\ref{fig_intro_div-19}). Fig. \ref{imagenet-topk-window-new-metric-nat-ig-lens}(right) shows the effect of increasing $w$ on the $w$-LENS-prec@$k$-div metric on ImageNet.

\section{A Stronger Model for Attributional Robustness} \label{sec:adv-robust}

\begin{figure}[!ht]
  \begin{center}
    \includegraphics[width=0.15\textwidth]{./images/flower_19_image.pdf}
    \includegraphics[width=0.15\textwidth]{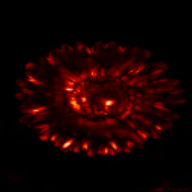}
    \includegraphics[width=0.15\textwidth]{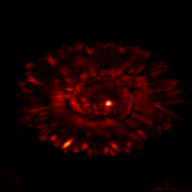}\\
    \includegraphics[width=0.15\textwidth]{./images/flower_19_image.pdf}
    \includegraphics[width=0.15\textwidth]{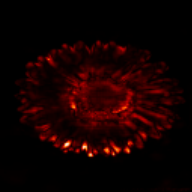}
    \includegraphics[width=0.15\textwidth]{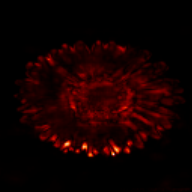}
  \end{center}
      \caption{From left to right: a sample image from Flower dataset and Integrated Gradients (IG) before and after the top-$k$ attributional attack of \citet{ghorbani2018interpretation}. The top row uses PGD-trained model whereas the bottom row uses IG-SUM-NORM-trained model.}
\label{flower-sample-pgd-rar-topk-19}
\end{figure}

%
%
%
%
A common approach to get robust attributions is to keep the attribution method unchanged but train the models differently in a way that the resulting attributions are more robust to small perturbations of inputs. \citet{chen2019robust} proposed the first defense against the attributional attack of \citet{ghorbani2018interpretation}. \citet{wang2020smoothed} also find that IG-NORM based training of \citet{chen2019robust} gives models that exhibit attributional robustness against the top-$k$ attack of \citet{ghorbani2018interpretation} along with adversarially trained models. Figure \ref{flower-sample-pgd-rar-topk-19} shows a sample image from the Flower dataset, where the Integrated Gradients (IG) of the original image and its perturbation by the top-$k$ attack are visually similar for models that are either adversarially trained (trained using Projected Gradient Descent or PGD-trained, as proposed by \citep{madry2019deep}) or IG-SUM-NORM trained as in \citet{chen2019robust}. In other words, these differently trained model guard the sample image against the attributional top-$k$ attack. Recent work by \citet{nourelahi2022} has empirically studied the effectiveness of adversarially (PGD) trained models in obtaining better attributions, e.g., Figure \ref{flower-sample-pgd-rar-topk-19}(center) shows sharper attributions to features highlighting the ground-truth class.

\begin{figure}[!ht]
\centering
\includegraphics[width=0.219\textwidth]{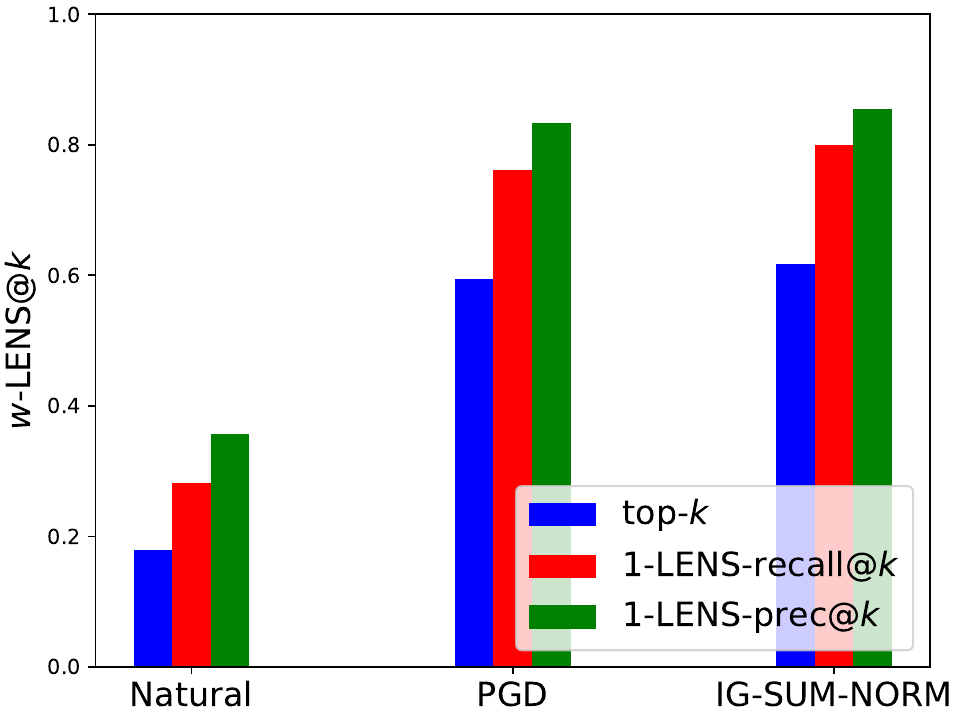}
\caption{For Flower dataset, average top-$k$ intersection, $1$-LENS-prec@$k$, $1$-LENS-recall@$k$ measured between IG(original image) and IG(perturbed image) for models that are naturally trained, PGD-trained and IG-SUM-NORM trained. The perturbation used is the top-$t$ attack of \cite{ghorbani2018interpretation}. Note top-$k$ is equivalent to $0$-LENS-prec@$k$, $0$-LENS-recall@$k$.}
\label{flower-nat-pgd-rar-ig-diff-lens} 
\end{figure}

Figure \ref{flower-nat-pgd-rar-ig-diff-lens} shows that PGD-trained and IG-SUM-NORM trained models have more robust Integrated Gradients (IG) in comparison to their naturally trained counterparts, and this holds for the previously used measures of attributional robustness (e.g., top-$k$ intersection) as well as the new locality-sensitive measures we propose (e.g., $1$-LENS-prec@$k$, $1$-LENS-recall@$k$) across all datasets in \citet{chen2019robust} experiments (Refer Appendix \ref{app:all-ig-exp} and \ref{app:all-sg-exp}). The top-$k$ attack of \citet{ghorbani2018interpretation} is not a threat to IG if we simply measure its effectiveness using $1$-LENS-prec@$k$ (Appendix \ref{app:all-ig-exp}, \ref{app:all-sg-exp} for MNIST, Fashion MNIST and GTSRB). The above observation about robustness of Integrated Gradients (IG) for PGD-trained and IG-SUM-NORM trained models holds even when we use $1$-LENS-Spearman and $1$-LENS-Kendall measures to quantify the attributional robustness to the top-$k$ attack of \citet{ghorbani2018interpretation}, and it holds across the datasets used by \citet{chen2019robust} in their study; see Appendix \ref{app:more-lens-results}.


\citet{ChalasaniC00J20} show theoretically that $\ell_{\infty}$-adversarial training (PGD-training) leads to stable Integrated Gradients (IG) under $\ell_{1}$ norm. They also show empirically that PGD-training leads to sparse attributions (IG \& DeepSHAP) when sparseness in measured indirectly as the change in Gini index. Our empirical results extend their theoretical observation about stability of IG for PGD-trained models, as we measure local stability in terms of both the top attribution values and their positions in the image.

Table \ref{table-topk-div-results-rand-imagenet} obtains the top-$k$ intersection, 3-LENS-recall@$k$, and 3-LENS-recall@$k$-div of different attribution methods on ImageNet for naturally trained and PGD-trained ResNet50 models. We observe that for random sign attack the improvement obtained on top-$k$ intersection is reduced for a large dataset like ImageNet. Still our conclusions about locality and diversity in attribution robustness in comparison with the top-$k$ intersection baseline holds as we observe improvements in using diversity and locality. 
More results about incorporating diversity in the attribution and the resulting robustness metrics are available in Appendix \ref{app:extra-div-results}.
%

\begin{table}[!ht]    
\begin{center}
\resizebox{0.47\textwidth}{!}{ 
\begin{tabular}{|l|l|c|c|c|}
\hline 
{\bf Training} & {\bf Attribution method} & {\bf top-$k$} & {\bf 3-LENS-recall@$k$} & {\bf 3-LENS-recall@$k$-div} \\
\hline
Natural & Simple Gradient       & 0.3825 & 0.7875 & 0.8290 \\
Natural & Image $\times$ Gradient  & 0.3316 & 0.7765 & 0.8655 \\
Natural & LRP [Bach 2015]     & 0.1027 & 0.2487 & 0.7518 \\
Natural & DeepLIFT [Shrikumar 2017] & 0.2907 & 0.7641 & 0.8504 \\
Natural & GradSHAP [Lundberg 2017] & 0.2290 & 0.6513 & 0.8099 \\
Natural & IG [Sundararajan 2017]      & 0.2638 & 0.7148 & 0.8380 \\
\hline 
PGD & Simple Gradient       & 0.1725 & 0.7245 & 0.8004 \\
PGD & Image $\times$ Gradient      & 0.1714 & 0.7269 & 0.8552 \\
PGD & LRP [Bach 2015] & 0.2374 & 0.4147 & 0.8161 \\
PGD & DeepLIFT [Shrikumar 2017] & 0.5572 & 0.9746 & 0.8977 \\
PGD & GradSHAP [Lundberg 2017] & 0.1714 & 0.7270 & 0.8552 \\
PGD & IG [Sundararajan 2017]      & 0.1947 & 0.7335 & 0.8584 \\
\hline
\end{tabular}
}
\caption{Average top-$k$ intersection, 3-LENS-prec@$k$ and 3-LENS-prec@$k$-div for random sign perturbation attack applied to different attribution methods on ImageNet for naturally and adversarially(PGD)-trained ResNet50 models.}
\label{table-topk-div-results-rand-imagenet}
\end{center}
\end{table}

\section{Conclusion and Future Work} \label{sec:conclusion}
We show that the fragility of attributions is an effect of using fragile robustness metrics such as top-$k$ intersection that only look at the rank order of attributions and fail to capture the locality of pixel positions with high attributions. We highlight the need for locality-sensitive metrics for attributional robustness and propose natural locality-sensitive extensions of existing metrics. We introduce another method of picking diverse top-$k$ pixels that can be naturally extended with locality to obtain improved measure of attributional robustness. Theoretical understanding of locality-sensitive metrics of attributional robustness, constructing stronger attributional attacks for these metrics, and using them to build attributionally robust models are important future directions. 


\bibliography{anonymous-submission-latex-2024}
\bibliographystyle{aaai24}

\clearpage

\appendix
\section{Supplementary : Rethinking Robustness of Model Attributions}

The Appendix contains proofs, additional experiments to show that the trends hold across different datasets and other ablation studies which could not be included in the main paper due to space constraints. 

\section{Attributional Robustness Metrics are Weak} \label{app:random-ext-fragile}


In this section, we empirically show that the existing metrics for attributional robustness are weak and inadequate as they allow even a small random perturbation to appear like a decent attributional attack.

\subsection{Random Vectors are Attributional Attacks under Existing Metrics}
\label{app:subsec:rand-vec}
Random vectors of a small $\ell_{\infty}$ norm are often used as baselines of input perturbations (both in adversarial robustness \citep{silva2020advrobustsurvey} and attributional robustness literature \cite{ghorbani2018interpretation}), since it is known that predictions of neural network models are known to be resilient to random perturbations of inputs. 
Previous work by \citet{ghorbani2018interpretation} has shown random perturbations to be a reasonable baseline to compare against their attributional attack. Extending it further, we show that a single input-agnostic random perturbation happens to be an effective \emph{universal} attributional attack if we measure attributional robustness using a weak metric based on top-$k$ intersection. In other words, considering even a random perturbation happens to be a good attributional attack under such metrics, we show that existing metrics for attributional robustness such as top-$k$ intersection are extremely fragile, i.e., they would unfairly deem many attribution methods as fragile. 

\begin{figure}[!ht]
\centering
    \begin{subfigure}[b]{0.18\textwidth}
    \includegraphics[width=\textwidth]{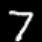}
    \caption{Original image}
    \label{mnist-sample-fig:a}
    \end{subfigure}
    \begin{subfigure}[b]{0.18\textwidth}
    \includegraphics[width=\textwidth]{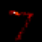}
    \caption{IG(original image)}
    \label{mnist-sample-fig:b}
    \end{subfigure}\\
    \begin{subfigure}[b]{0.18\textwidth}
    \includegraphics[width=\textwidth]{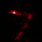}
    \caption{IG after random}
    \label{mnist-sample-fig:c}
    \end{subfigure}
    \begin{subfigure}[b]{0.18\textwidth}
    \includegraphics[width=\textwidth]{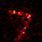}
    \caption{IG after universal}
    \label{mnist-sample-fig:d}
    \end{subfigure}
    \caption{Sample image from MNIST on LeNet based model shows that the Integrated Gradients (IG) after a universal random perturbation are more dissimilar than IG after a simple, independent random perturbation for each input. All perturbations have random $\pm 1$ coordinates, scaled down to have $\ell_{\infty}$ norm $\epsilon=0.3$. (c) has a top-$k$ intersection of 0.68, while (d) has a top-$k$ intersection of 0.62. With our locality-sensitive metric, (c) has 1-LENS@$k$ of 0.99 and (d) has 1-LENS@$k$ of 1.0.}
\label{mnist-sample-nat-rand-topk}
\end{figure}

Integrated Gradients (IG) is a well-known attribution method based on well-defined axiomatic foundations \citep{SundararajanTY17}, which is commonly used in attributional robustness literature \citep{chen2019robust,sarkar2020enhanced}. 
We take a naturally trained CNN model on MNIST and perturb the images using a random perturbation (an independent random perturbation per input image) as well as a single, input-agnostic or \emph{universal} random perturbation for all images. Figure \ref{mnist-sample-nat-rand-topk} shows a sample image from the MNIST dataset and the visual difference between the IG of the original image, the IG after adding a random perturbation, and the IG after adding a \emph{universal} random perturbation. The IG after the universal random attack (Figure \ref{mnist-sample-fig:d}) is visually more dissimilar to the IG of the original image (Figure \ref{mnist-sample-fig:b}) than the IG of a simple random perturbation (Figure \ref{mnist-sample-fig:c}). (Note that top-$k$ intersection between Figure \ref{mnist-sample-fig:b} and \ref{mnist-sample-fig:c} is only 0.62, although the two look similar. As stated in the caption, a locality-sensitive metric shows them to be closer in attribution however.)


Similarly, Table \ref{table-rand-metric} shows that under existing metrics to quantify attributional robustness of IG on a naturally trained CNN model, even a single, input-agnostic or \emph{universal} random perturbation can sometimes be a more effective attributional attack than using an independent random perturbation for each input. 
\begin{table}[!ht]
\begin{center}
\resizebox{0.47\textwidth}{!}{ 
\begin{tabular}{|c|c|c|c|c|}
\hline 
{\bf Dataset} & {\bf Perturbation} & {\bf top-$k$ intersection} & {\bf Spearman's $\rho$} & {\bf Kendall's $\tau$} \\
\hline
MNIST & random & 0.7500 & 0.5347 & 0.4337 \\
& universal random & 0.5855 & 0.4831 & 0.4063 \\
\hline
Fashion MNIST & random & 0.5385 & 0.6791 & 0.5152 \\ 
& universal random & 0.5280 & 0.7154 & 0.5688 \\
\hline 
GTSRB & random & 0.8216 & 0.9433 & 0.8136 \\
& universal random & 0.9293 & 0.9887 & 0.9243 \\
\hline
Flower & random & 0.8202 & 0.9562 & 0.8340 \\ 
& universal random & 0.9344 & 0.9908 & 0.9321 \\ 
\hline
\end{tabular}
}
\end{center}
\caption{Attributional robustness of IG on naturally trained models measured using average top-$k$ intersection, Spearman's $\rho$ and Kendall's $\tau$ between IG(original image) and IG(perturbed image). $k=100$ for MNIST, Fashion MNIST, GTSRB and $k=1000$ for Flower. 
}
\label{table-rand-metric}
\end{table}


\section{Details of Experimental Setup} \label{app:exp-setup}
The detailed description of the setup used in our experiments.

\textit{Datasets:} We use the standard benchmark train-test split of all the datasets used in this work, that is publicly available. MNIST dataset consists of $70,000$ images of $28 \times 28$ size, divided into $10$ classes: $60,000$ used for training and $10,000$ for testing. Fashion MNIST dataset consists of $70,000$ images of $28 \times 28$ size, divided into $10$ classes: $60,000$ used for training and $10,000$ for testing. GTSRB dataset consists of $51,739$ images of $32 \times 32$ size, divided into $43$ classes: $34,699$ used for training, $4,410$ for validation and $12,630$ for testing. Flower dataset consist of $1,360$ images of $128 \times 128$ size, divided into $17$ classes: $1,224$ used for training and $136$ for testing. GTSRB and Flower datasets were preprocessed exactly as given in \cite{chen2019robust}[Appendix C] for consistency of results. ImageNet dataset consists of images of $227 \times 227$ size, divided into $1000$ classes. $50,000$ for validation were used to obtain samples for testing.

\textit{Architectures:} For MNIST, Fashion MNIST, GTSRB and Flower datasets we use the exact architectures as used by \citet{chen2019robust}, ImageNet dataset we use SqueezeNet\citep{IandolaMAHDK16} as given by \citet{ghorbani2018interpretation} and ResNet50\citep{HeZRS15}.

\textit{Attribution robustness metrics:} We use the same comparison metrics as used by \citet{ghorbani2018interpretation} and \citet{chen2019robust} like top-$k$ pixels intersection, Spearman's $\rho$ and Kendall's $\tau$ rank correlation to compare attribution maps of the original and perturbed images. The $k$ value for top-$k$ attack along with settings like step size, number of steps and number of times attack is to be applied is as used by \citet{chen2019robust} for the attack construction : MNIST(200,0.01,100,3), Fashion MNIST(100,0.01,100,3), GTSRB(100,1.0,50,3), Flower(1000,1.0,100,3) and ImageNet(1000,1.0,100,3) \citep{ghorbani2018interpretation}.

\textit{Sample sizes for attribution robustness evaluations:} \textit{IG based experiments} For MNIST, Fashion MNIST and Flower with fixed top-$k$ attack similar to \citet{chen2019robust} the complete test set were used to obtain the results. For GTSRB a random sample of size 1000 was used for all the experiments.  \textit{Simple gradient based experiments} For MNIST and Fashion MNIST a random sample of 2500/1000 from the test set. For GTSRB, a random sample of size 1000 and the complete test set for the Flower dataset. For ImageNet, we used the samples provide by \citet{ghorbani2018interpretation}. We used around 500 random samples for the random sign perturbation results using Pytorch/Captum.

\textit{Adversarial training:} We use the standard setup as used by \citep{chen2019robust}. We perform PGD based adversarial training with the provided $\epsilon$ budget using the following settings (number of steps, step size) for PGD : MNIST (40,0.01), Fashion MNIST(20,0.01), GTSRB(7,2/255), Flower(7,2/255). For ImageNet, we used the PGD based pre-trained ResNet50 model with $\ell_{\infty}$-norm of $\epsilon = 8/255$ provided in the robustness package\footnote{https://github.com/MadryLab/robustness/}.

\textit{Training for Attributional Robustness:} We use the IG-SUM-NORM objective function for all the datasets study based on \citep{chen2019robust} based training. With the exact setting as given in paper with code \footnote{https://github.com/jfc43/robust-attribution-regularization}.

\textit{Hardware Configuration:} We used a server with 4 Nvidia GeForce GTX 1080i GPU and a server with 8 Nvidia Tesla V100 GPU to run the experiments in the paper.

\textit{Explanation methods:} The Tensorflow \footnote{https://www.tensorflow.org} code for  IG, SG, DeepLIFT we used \citep{chen2019robust} and \citet{ghorbani2018interpretation} depending on the dataset. We used Pytorch\footnote{https://github.com/pytorch}- Captum\footnote{https://github.com/pytorch/captum} to obtain the various explanation maps with the random sign perturbation experiments.

\section{Proofs from Section \ref{subsec:define-lens}} \label{app:proofs}
We restate and prove Proposition \ref{prop:monotone} below.
\begin{prop} 
For any $w_{1} \leq w_{2}$, we have $d_{k}^{(w_{2})}(\ba_{1}, \ba_{2}) \leq d_{k}^{(w_{1})}(\ba_{1}, \ba_{2}) \leq \size{S_{k} \triangle T_{k}}/k$, where $\triangle$ denotes the symmetric set difference, i.e., $A \triangle B = (A \setminus B) \cup (B \setminus A)$.
\end{prop}
\begin{proof}
The inequalities follows immediately using $S \subseteq N_{w_{1}}(S) \subseteq N_{w_{2}}(S)$, for any $S$, and hence, $\size{S \setminus N_{w}(T)} \leq \size{S \setminus T}$, for any $S, T$ and $w$.
\end{proof}
We restate and prove Proposition \ref{prop:metric} below.
\begin{prop}
$d(\ba_{1}, \ba_{2})$ defined above is upper bounded by $u(\ba_{1}, \ba_{2})$ given by
\[
u(\ba_{1}, \ba_{2}) = \sum_{k=1}^{\infty} \alpha_{k} \sum_{w=0}^{\infty} \beta_{w}~ \frac{\size{S_{k} \triangle T_{k}}}{k},
\]
and $u(\ba_{1}, \ba_{2})$ defines a bounded metric on the space of attribution vectors.
\end{prop}
\begin{proof}
Proof follows from Proposition \ref{prop:monotone} and using the fact that symmetric set difference satisfies triangle inequality.
\end{proof}
We restate and prove Proposition \ref{prop:w-smoothed} below.
\begin{prop}
For any inputs $x, y$ and any $w \geq 0$, $\norm{\tilde{\ba}^{(w)}(x) - \tilde{\ba}^{(w)}(y)}_{2} \leq \norm{\ba(x) - \ba(y)}_{2}$.
\end{prop}
\begin{proof}

$\norm{\tilde{\ba}^{(w)}(x) - \tilde{\ba}^{(w)}(y)}_{2}^{2}$ 
\begin{align*}
&= \sum_{1 \leq i, j \leq n} \left(\tilde{a}_{ij}^{(w)}(x) - \tilde{a}_{ij}^{(w)}(y)\right)^{2} \\ 
& = \sum_{1 \leq i, j \leq n} \frac{1}{(2w+1)^{4}} \left(\sum_{\substack{(p, q) \in N_{w}(i, j), \\ 1 \leq p, q \leq n}} \left(a_{pq}(x) - a_{pq}(y)\right)\right)^{2} \\
& \leq \sum_{1 \leq i, j \leq n} \frac{(2w+1)^{2}}{(2w+1)^{4}} \sum_{\substack{(p, q) \in N_{w}(i, j), \\ 1 \leq p, q \leq n}} \left(a_{pq}(x) - a_{pq}(y)\right)^{2}~ 
\end{align*}
\text{by Cauchy-Schwarz inequality} \\

\begin{align*}
& = \frac{1}{(2w+1)^{2}} \sum_{1 \leq i, j \leq n} \sum_{\substack{(p, q) \in N_{w}(i, j), \\ 1 \leq p, q \leq n}} \left(a_{pq}(x) - a_{pq}(y)\right)^{2} \\
& \leq \frac{(2w+1)^{2}}{(2w+1)^{2}} \sum_{1 \leq p, q \leq n} \left(a_{pq}(x) - a_{pq}(y)\right)^{2} \\
& \qquad 
\end{align*}

because each $(p, q)$ appears in at most $(2w+1)^{2}$ possibles $N_{w}(i, j)$'s

\begin{align*}
& = \norm{\ba(x) - \ba(y)}_{2}^{2}.
\end{align*}
\end{proof}


\section{Additional results for Section \ref{sec:lens}} \label{app:more-lens-results}

\subsection{Additional results for Section \ref{subsec:lens-relevance-robustness}} \label{app:vary-k-sp-ken}
\begin{figure}[!ht]
\centering
\begin{subfigure}[b]{1.0\textwidth}
    \begin{subfigure}[b]{0.12\textwidth}
    \includegraphics[width=1.0\textwidth]{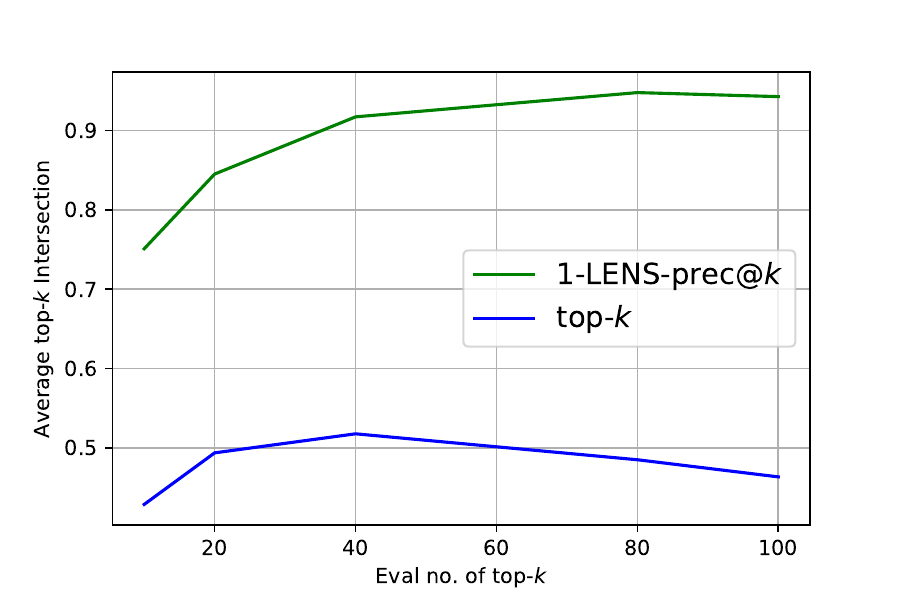}
    \caption{MNIST: top-$k$}
    \label{mnist-rand-fig:a}
    \end{subfigure}
    \begin{subfigure}[b]{0.12\textwidth}
    \includegraphics[width=1.0\textwidth]{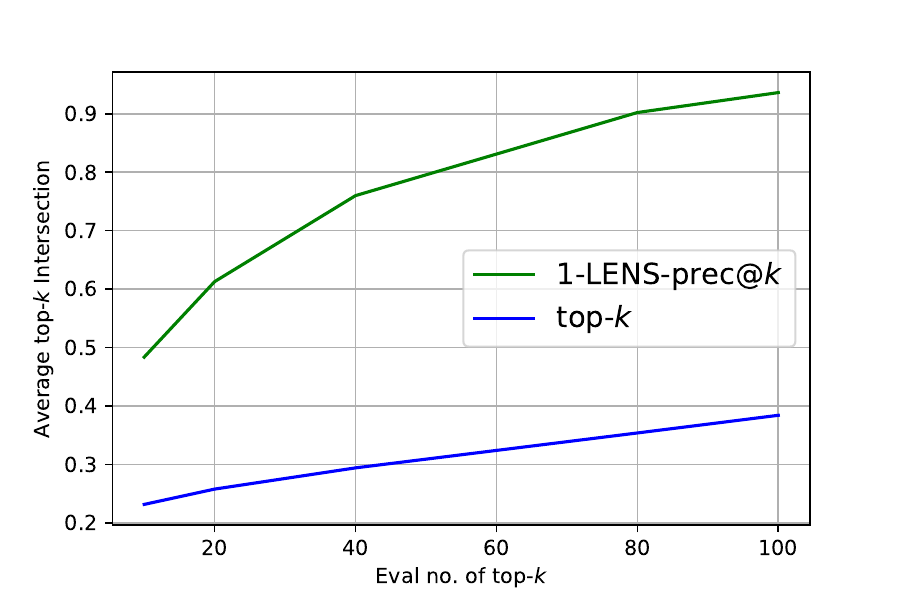}
    \caption{F-MNIST: top-$t$}
    \label{mnist-rand-fig:b}
    \end{subfigure}
    \begin{subfigure}[b]{0.12\textwidth}
    \includegraphics[width=1.0\textwidth]{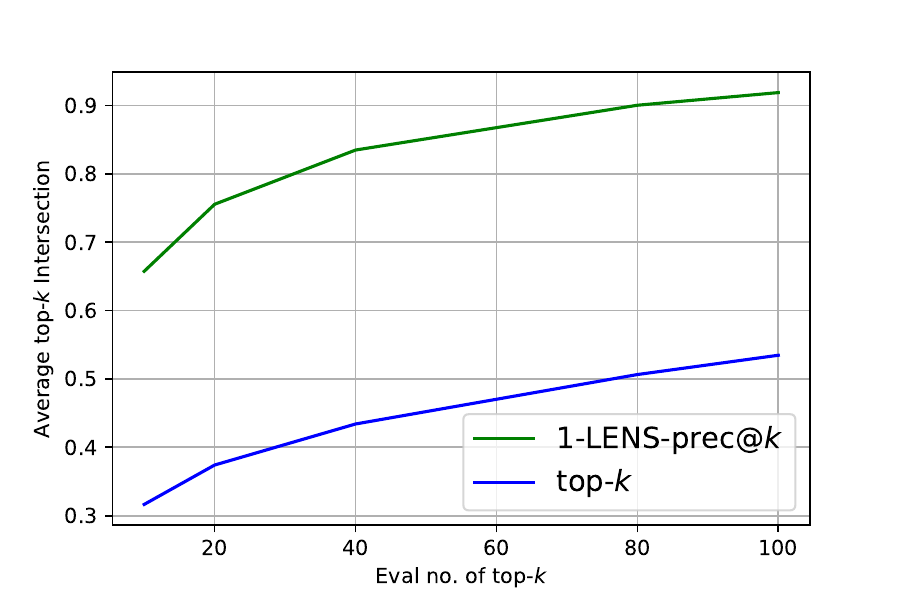}
    \caption{GTSRB: top-$k$}
    \label{mnist-rand-fig:c}
    \end{subfigure}
    \begin{subfigure}[b]{0.12\textwidth}
    \includegraphics[width=1.0\textwidth]{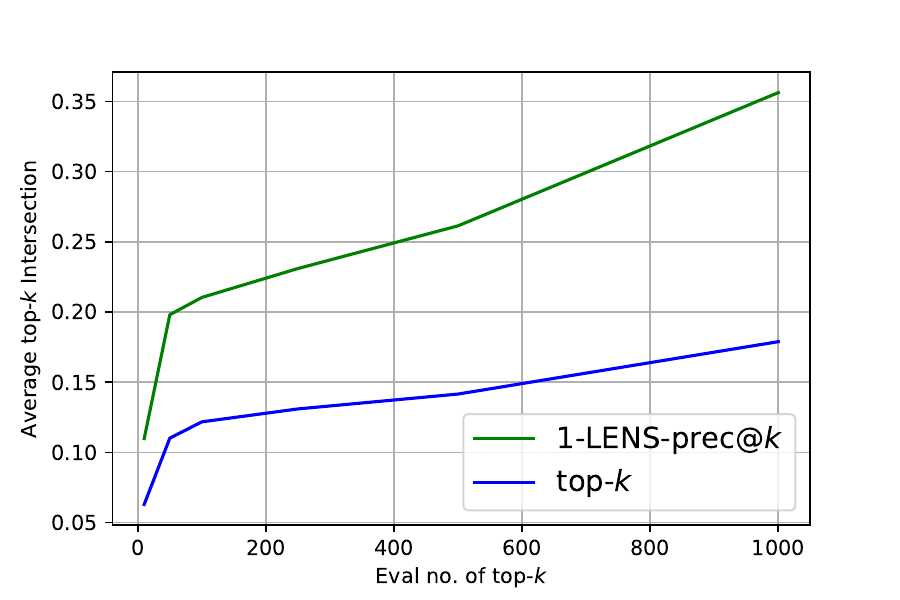}
    \caption{Flower: top-$k$}
    \label{mnist-rand-fig:c}
    \end{subfigure}
\end{subfigure}
\begin{subfigure}[b]{1.0\textwidth}
    \begin{subfigure}[b]{0.12\textwidth}
    \includegraphics[width=1.0\textwidth]{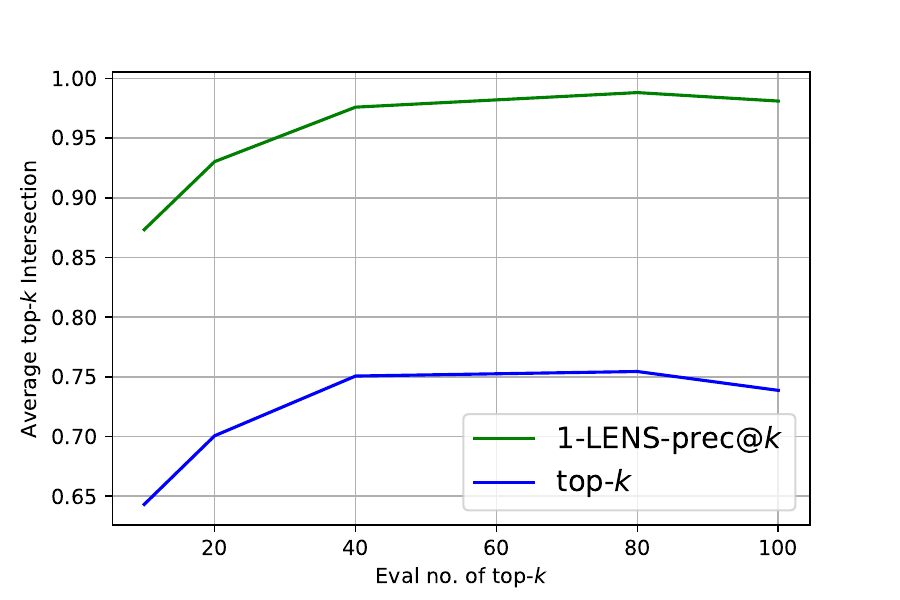}
    \caption{MNIST: random}
    \label{mnist-rand-fig:a}
    \end{subfigure}
    \begin{subfigure}[b]{0.12\textwidth}
    \includegraphics[width=1.0\textwidth]{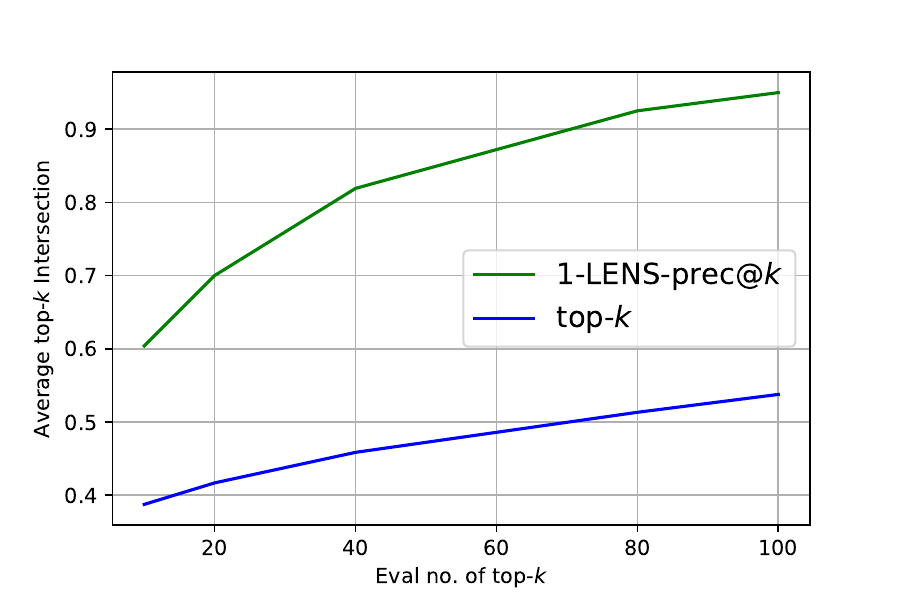}
    \caption{F-MNIST: random}
    \label{mnist-rand-fig:b}
    \end{subfigure}
    \begin{subfigure}[b]{0.12\textwidth}
    \includegraphics[width=1.0\textwidth]{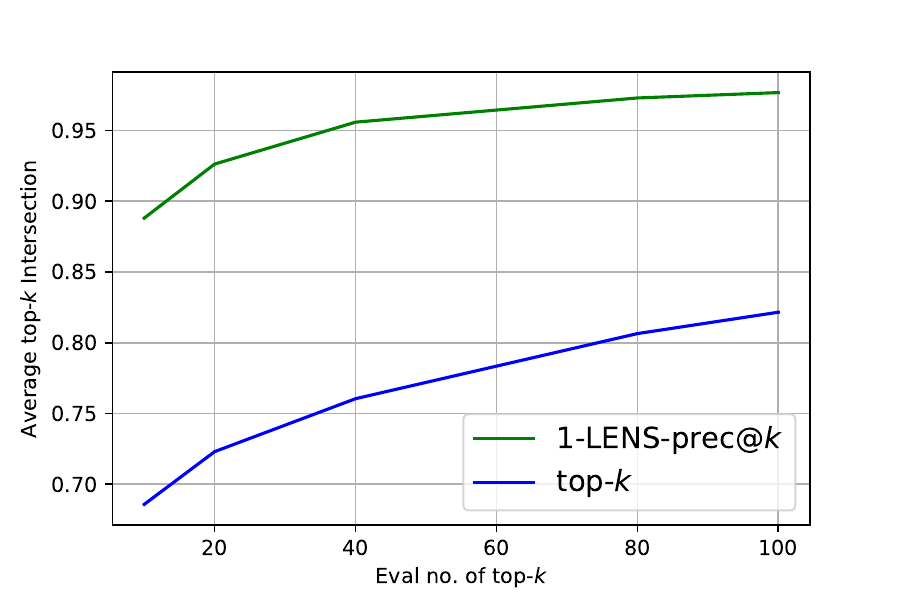}
    \caption{GTSRB: random}
    \label{mnist-rand-fig:c}
    \end{subfigure}
    \begin{subfigure}[b]{0.12\textwidth}
    \includegraphics[width=1.0\textwidth]{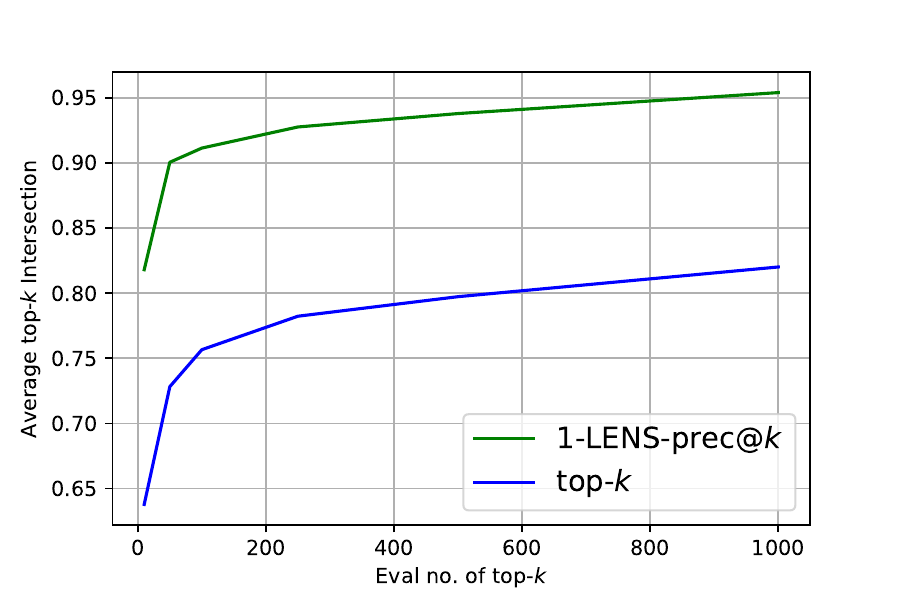}
    \caption{Flower: random}
    \label{mnist-rand-fig:c}
    \end{subfigure}
\end{subfigure}
\caption{Attributional robustness of IG on naturally trained models measured as average top-$k$ intersection and $1$-LENS-prec@$k$ between IG(original image) and IG(perturbed image). Perturbations are obtained by the {\bf top-$k$ attack} \citep{ghorbani2018interpretation} and random perturbation. The plots show how the above measures change with varying $k$ across different datasets.}
\label{dataset-topk-rand-diff-k-new-metric-nat}     
\end{figure}

\begin{figure}[!ht]
\centering
\begin{subfigure}[b]{1.0\textwidth}
    \begin{subfigure}[b]{0.12\textwidth}
    \includegraphics[width=1.0\textwidth]{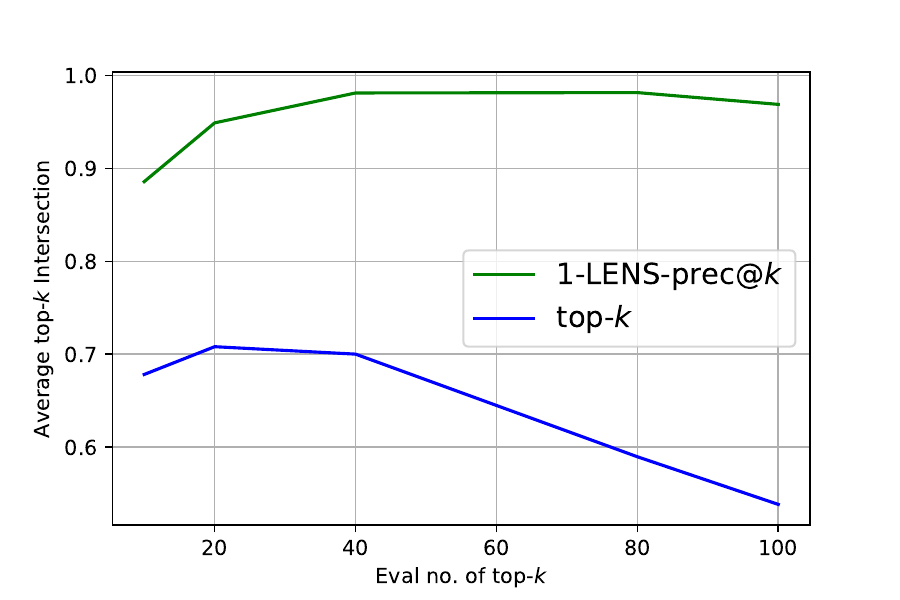}
    \caption{MNIST IG : top-$k$}
    \label{mnist-rand-fig:a}
    \end{subfigure}
    \begin{subfigure}[b]{0.12\textwidth}
    \includegraphics[width=1.0\textwidth]{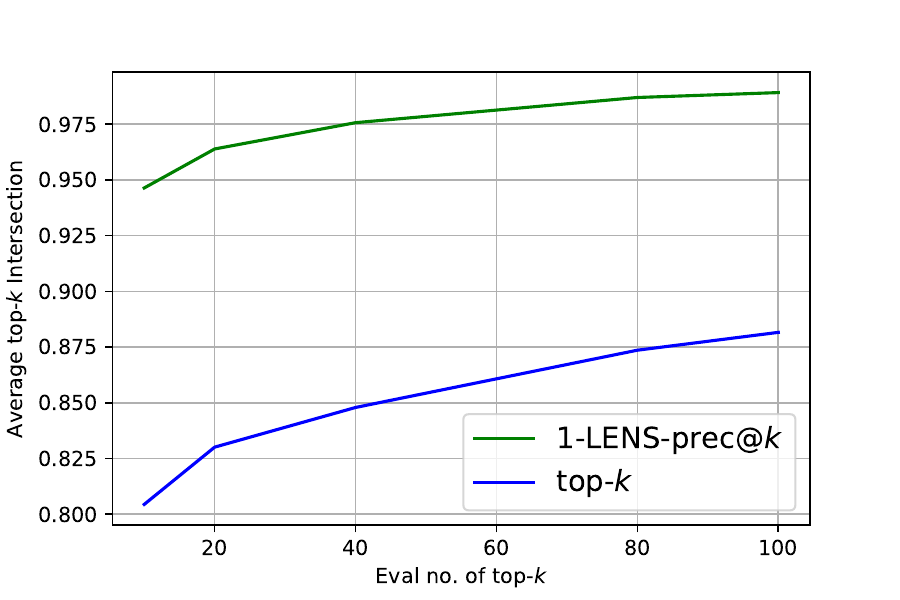}
    \caption{F-MNIST IG : top-$k$}
    \label{mnist-rand-fig:b}
    \end{subfigure}
    \begin{subfigure}[b]{0.12\textwidth}
    \includegraphics[width=1.0\textwidth]{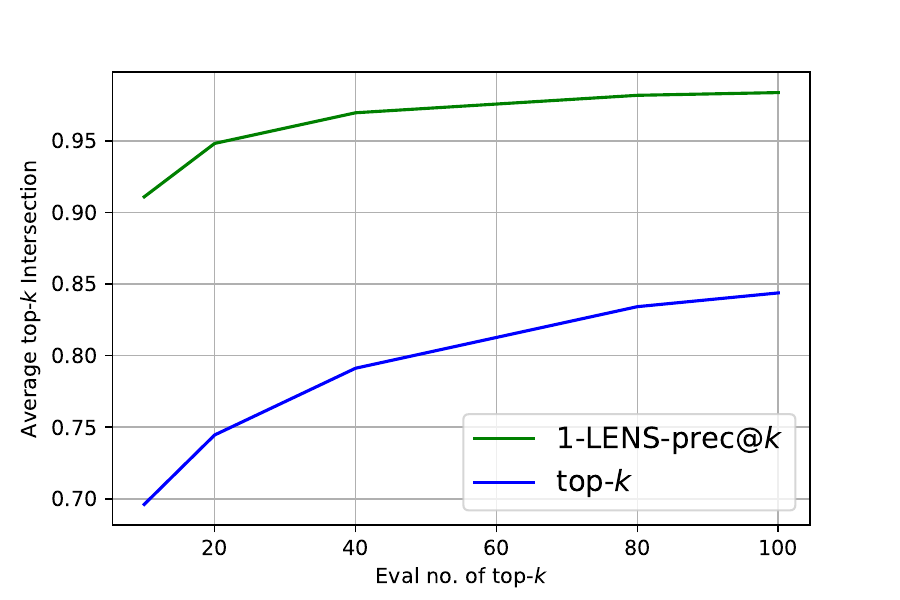}
    \caption{GTSRB IG : top-$k$}
    \label{mnist-rand-fig:b}
    \end{subfigure}
    \begin{subfigure}[b]{0.12\textwidth}
    \includegraphics[width=1.0\textwidth]{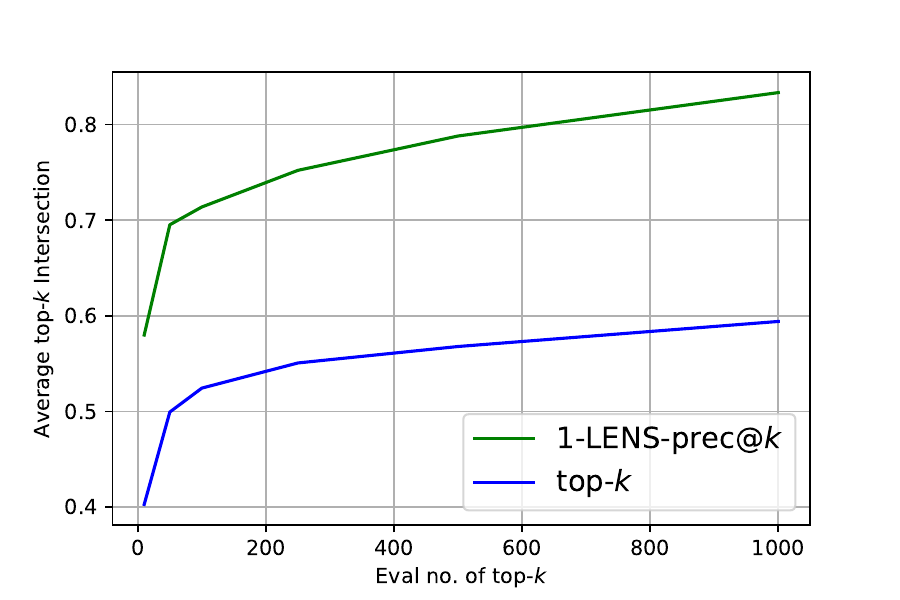}
    \caption{Flower IG : top-$k$}
    \label{mnist-rand-fig:c}
    \end{subfigure}
\end{subfigure}
\begin{subfigure}[b]{1.0\textwidth}
    \begin{subfigure}[b]{0.12\textwidth}
    \includegraphics[width=1.0\textwidth]{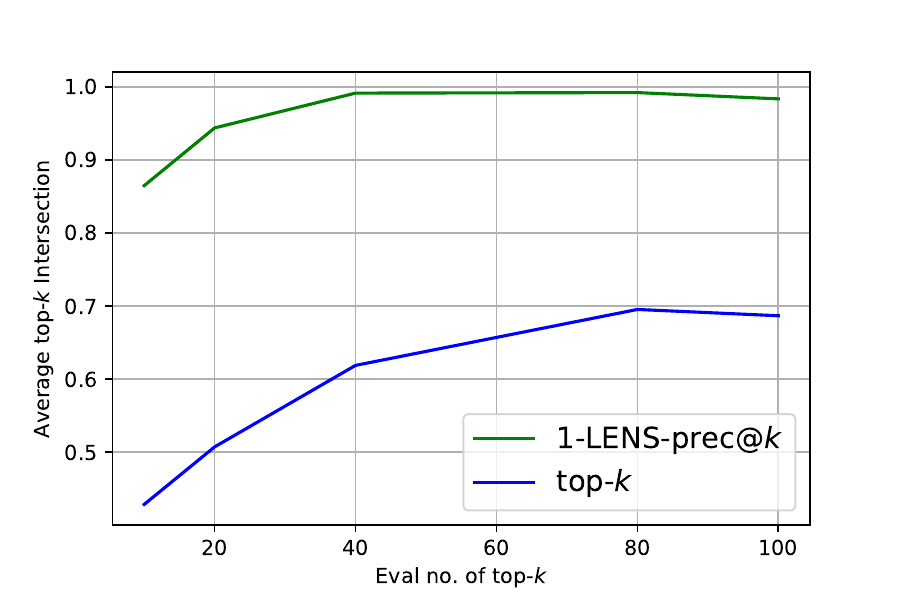}
    \caption{MNIST IG : random}
    \label{mnist-rand-fig:a}
    \end{subfigure}
    \begin{subfigure}[b]{0.12\textwidth}
    \includegraphics[width=1.0\textwidth]{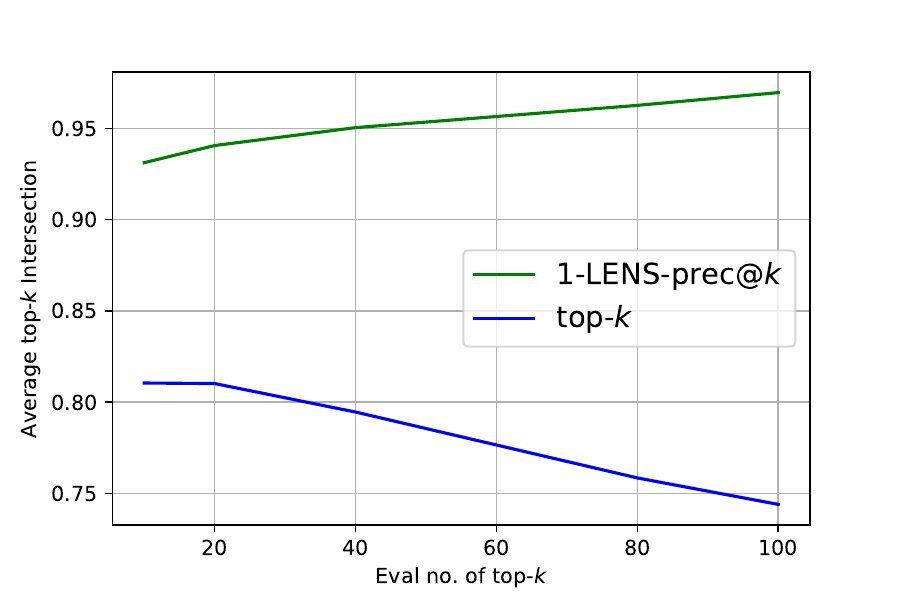}
    \caption{F-MNIST IG : random}
    \label{mnist-rand-fig:b}
    \end{subfigure}
    \begin{subfigure}[b]{0.12\textwidth}
    \includegraphics[width=1.0\textwidth]{./plots/gtsrb_topkvsrand_rand_pgd.pdf}
    \caption{GTSRB IG : random}
    \label{mnist-rand-fig:b}
    \end{subfigure}
    \begin{subfigure}[b]{0.12\textwidth}
    \includegraphics[width=1.0\textwidth]{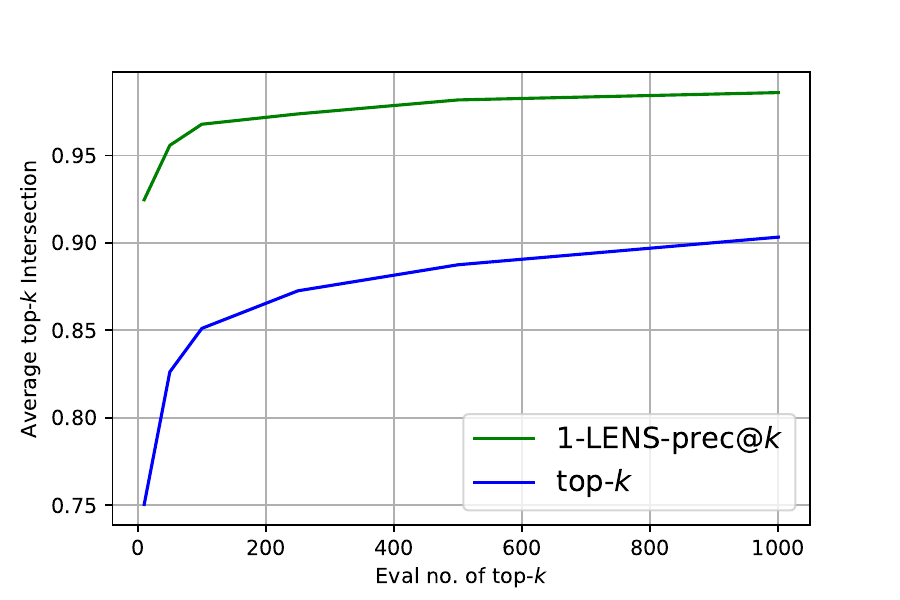}
    \caption{Flower IG : random}
    \label{mnist-rand-fig:c}
    \end{subfigure}
\end{subfigure}
\caption{Attributional robustness of IG on adversarially(PGD) trained models measured as average top-$k$ intersection and $1$-LENS-prec@$k$ between IG(original image) and IG(perturbed image). Perturbations are obtained by the {\bf top-$k$ attack} \citep{ghorbani2018interpretation} and random perturbation. The plots show how the above measures change with varying $k$ across different datasets.}
\label{dataset-topk-rand-diff-k-new-metric-pgd}  
\end{figure}

\begin{figure}[!ht]
\centering
\begin{subfigure}[b]{1.0\textwidth}
    \begin{subfigure}[b]{0.12\textwidth}
    \includegraphics[width=1.0\textwidth]{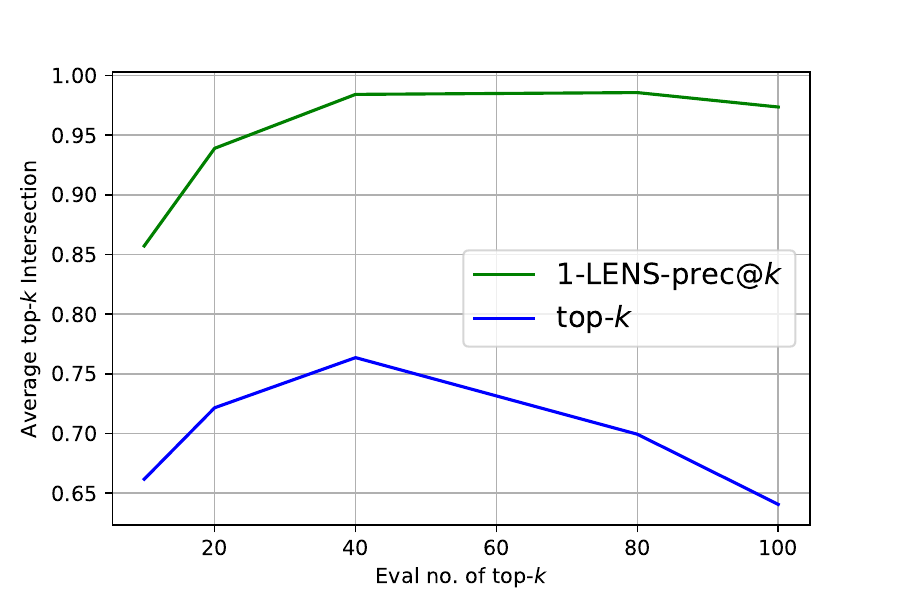}
    \caption{MNIST IG : top-$k$}
    \label{mnist-rand-fig:a}
    \end{subfigure}
    \begin{subfigure}[b]{0.12\textwidth}
    \includegraphics[width=1.0\textwidth]{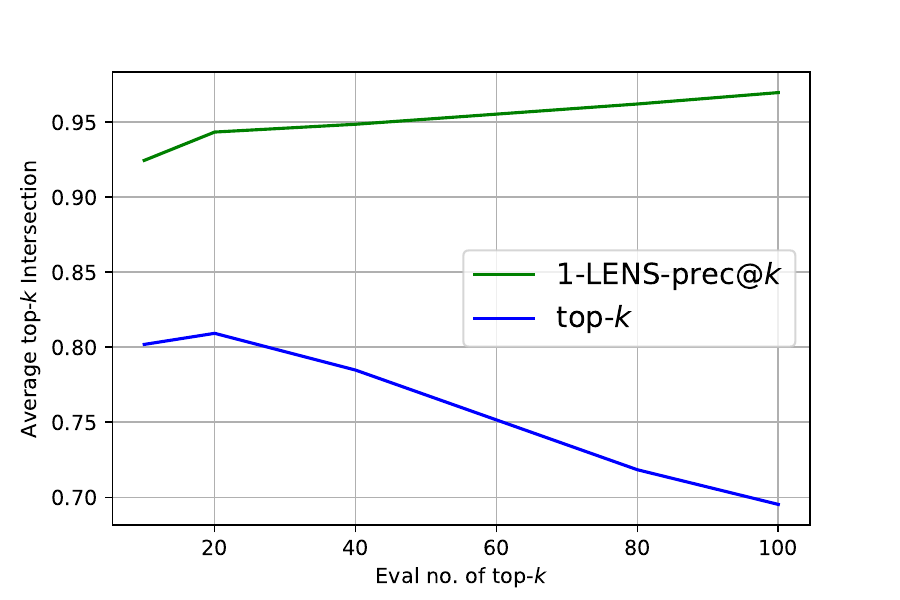}
    \caption{F-MNIST IG : top-$k$}
    \label{mnist-rand-fig:b}
    \end{subfigure}
    \begin{subfigure}[b]{0.12\textwidth}
    \includegraphics[width=1.0\textwidth]{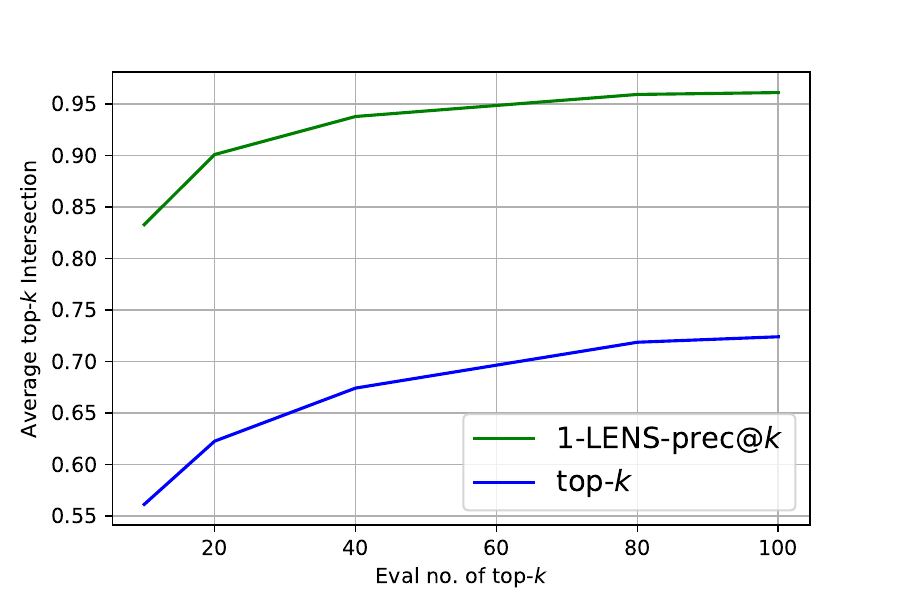}
    \caption{GTSRB IG : top-$k$}
    \label{mnist-rand-fig:b}
    \end{subfigure}
    \begin{subfigure}[b]{0.12\textwidth}
    \includegraphics[width=1.0\textwidth]{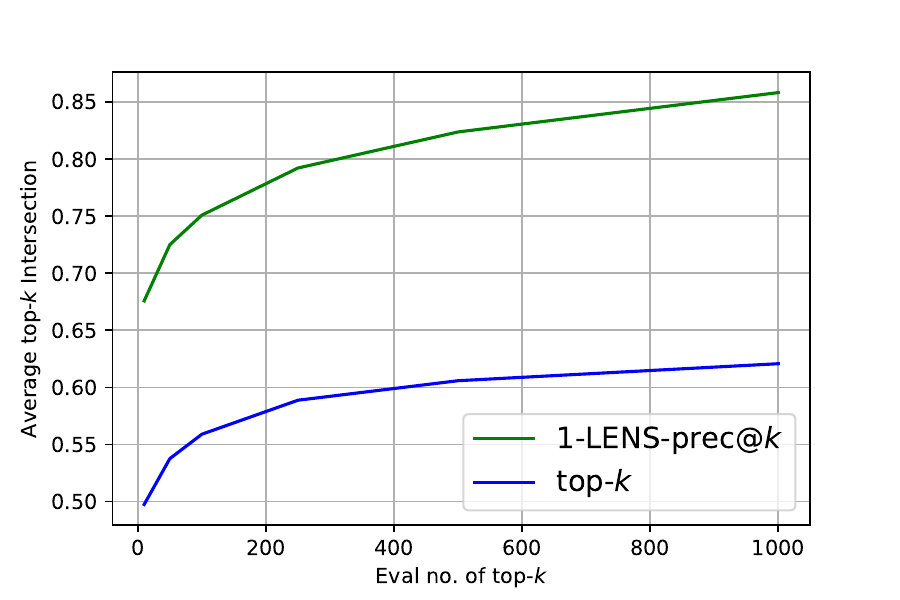}
    \caption{Flower IG : top-$k$}
    \label{mnist-rand-fig:c}
    \end{subfigure}
\end{subfigure}
\begin{subfigure}[b]{1.0\textwidth}
    \begin{subfigure}[b]{0.12\textwidth}
    \includegraphics[width=1.0\textwidth]{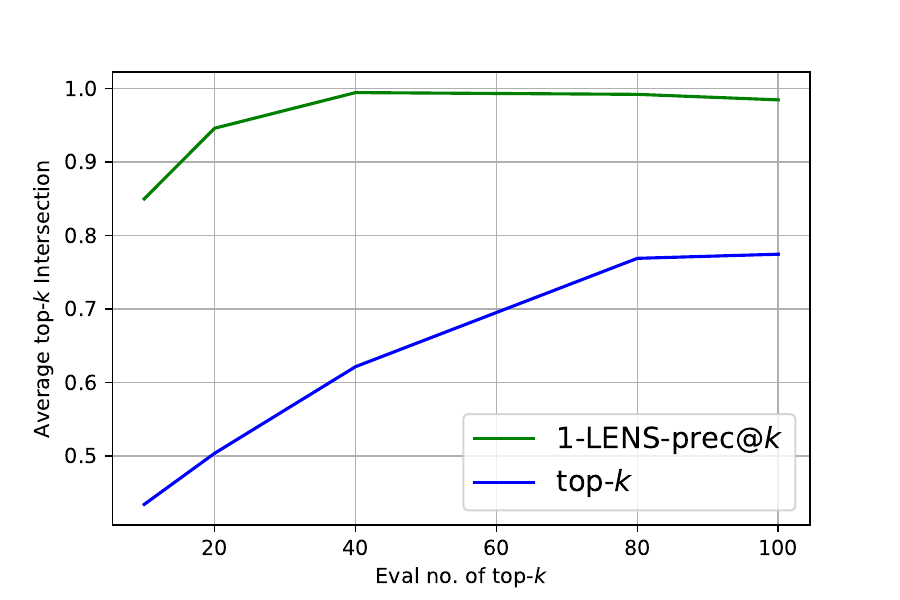}
    \caption{MNIST IG : random}
    \label{mnist-rand-fig:a}
    \end{subfigure}
    \begin{subfigure}[b]{0.12\textwidth}
    \includegraphics[width=1.0\textwidth]{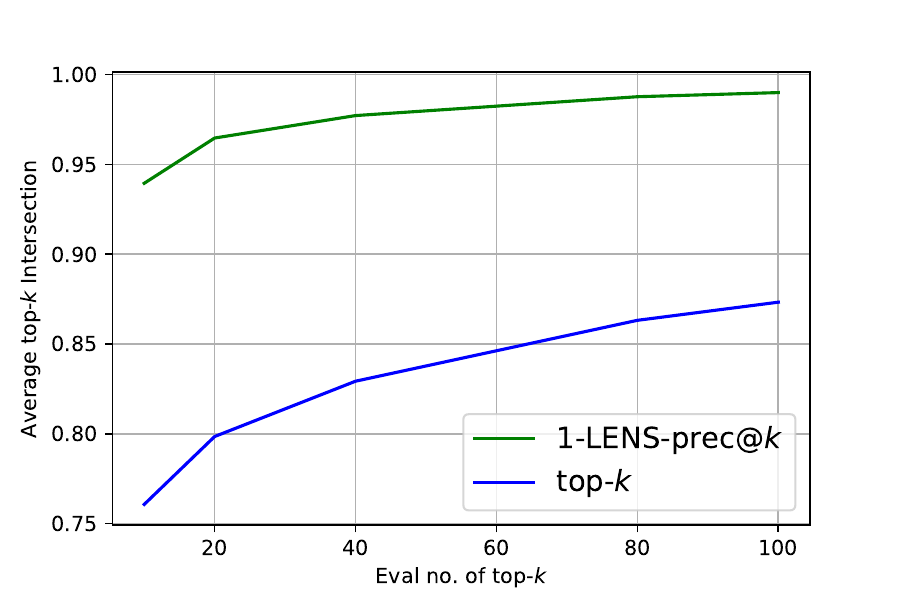}
    \caption{F-MNIST IG : random}
    \label{mnist-rand-fig:b}
    \end{subfigure}
    \begin{subfigure}[b]{0.12\textwidth}
    \includegraphics[width=1.0\textwidth]{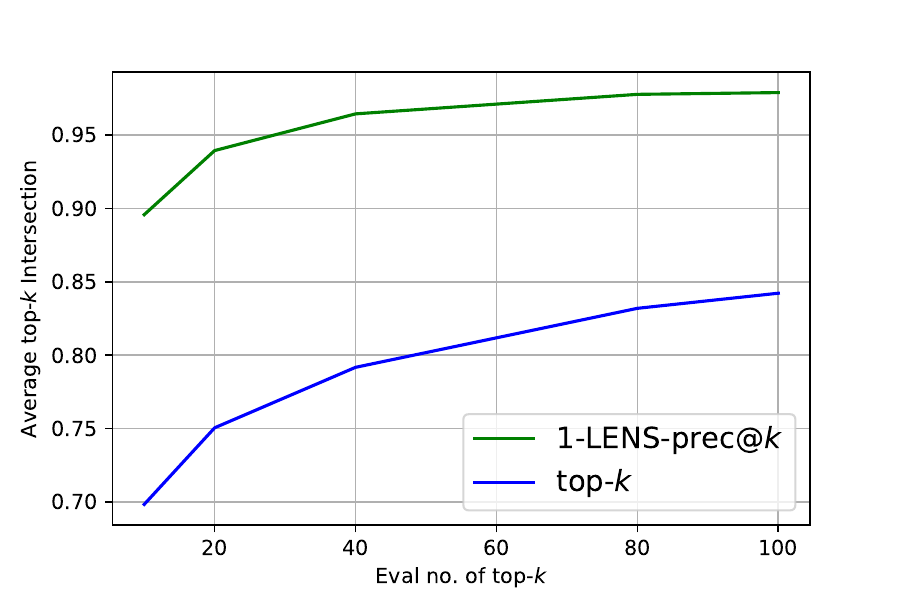}
    \caption{GTSRB IG : top-$k$}
    \label{mnist-rand-fig:b}
    \end{subfigure}
    \begin{subfigure}[b]{0.12\textwidth}
    \includegraphics[width=1.0\textwidth]{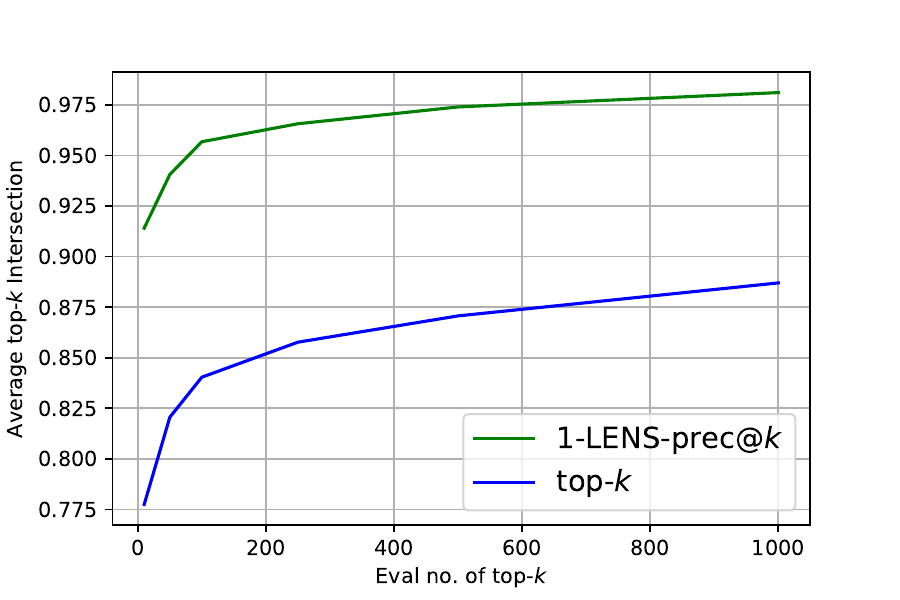}
    \caption{Flower IG : random}
    \label{mnist-rand-fig:c}
    \end{subfigure}
\end{subfigure}
\caption{Attributional robustness of IG on IG-SUM-NORM trained models measured as average top-$k$ intersection and $1$-LENS-prec@$k$ between IG(original image) and IG(perturbed image). Perturbations are obtained by the {\bf top-$k$ attack} \citep{ghorbani2018interpretation} and random perturbation. The plots show how the above measures change with varying $k$ across different datasets.}
\label{dataset-topk-rand-diff-k-new-metric-rar}  
\end{figure}

\paragraph{The effect of varying $k$.} Figure \ref{dataset-topk-rand-diff-k-new-metric-nat} shows a large disparity between top-$k$ intersection and $1$-LENS-prec@$k$ even when $k$ is large. Figure \ref{dataset-topk-rand-diff-k-new-metric-nat} shows that top-$k$ intersection can be very low even when the IG of the original and the IG of the perturbed images are locally very similar, as indicated by high $1$-LENS-prec@$k$. Our observation holds for the perturbations obtained by the top-$k$ attack \citep{ghorbani2018interpretation} as well as a random perturbation across all datasets in our experiments. Figure \ref{dataset-topk-rand-diff-k-new-metric-pgd} and \ref{dataset-topk-rand-diff-k-new-metric-rar} show the results on PGD and IG-SUM-NORM trained networks. 

\begin{figure}[!ht]
\centering
\includegraphics[width=0.38\textwidth]{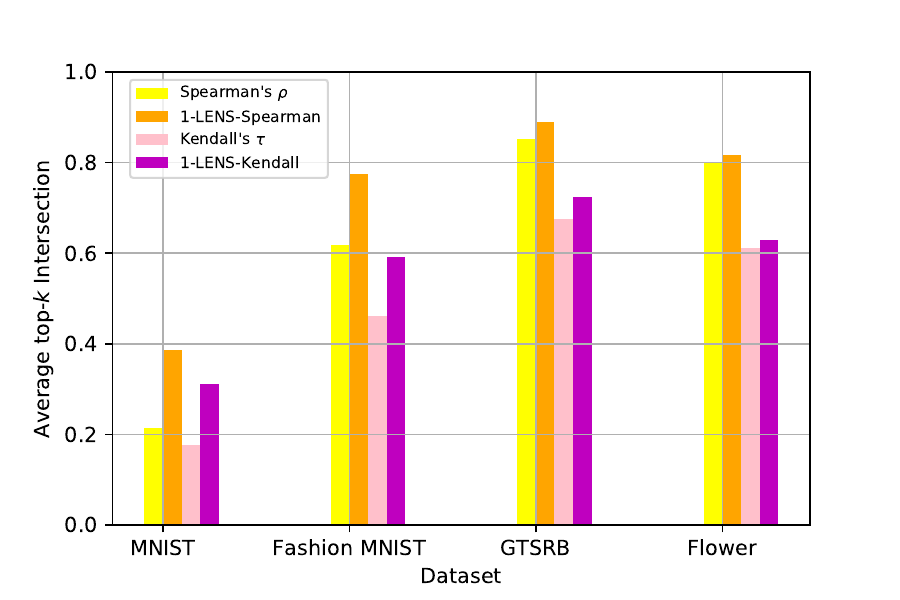}
\caption{Attributional robustness of IG on naturally trained models measured as average Spearman's $\rho$, $1$-LENS-Spearman, Kendall $\tau$ and $1$-LENS-Kendall between IG(original image) and IG(perturbed image). The perturbations are obtained by the top-$t$ attack of \citet{ghorbani2018interpretation}.} 
\label{dataset-smooth-sp-ken-nat}
\end{figure}

\begin{figure}[!ht]
\centering
\begin{subfigure}[b]{\textwidth}
    \begin{subfigure}[b]{0.24\textwidth}
    \includegraphics[width=\textwidth]{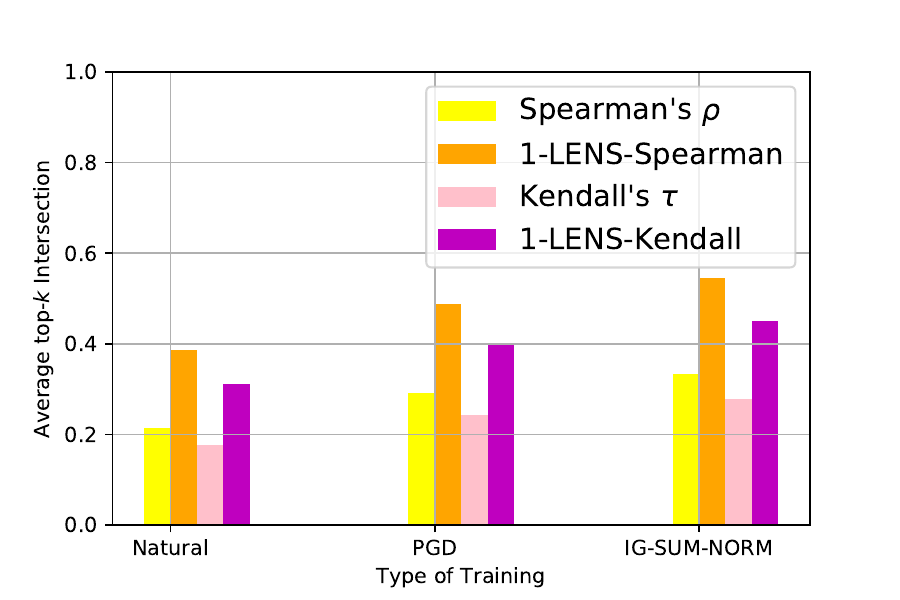}
    \caption{MNIST}
    \label{mnist-rand-fig:a}
    \end{subfigure}
    \begin{subfigure}[b]{0.24\textwidth}
    \includegraphics[width=\textwidth]{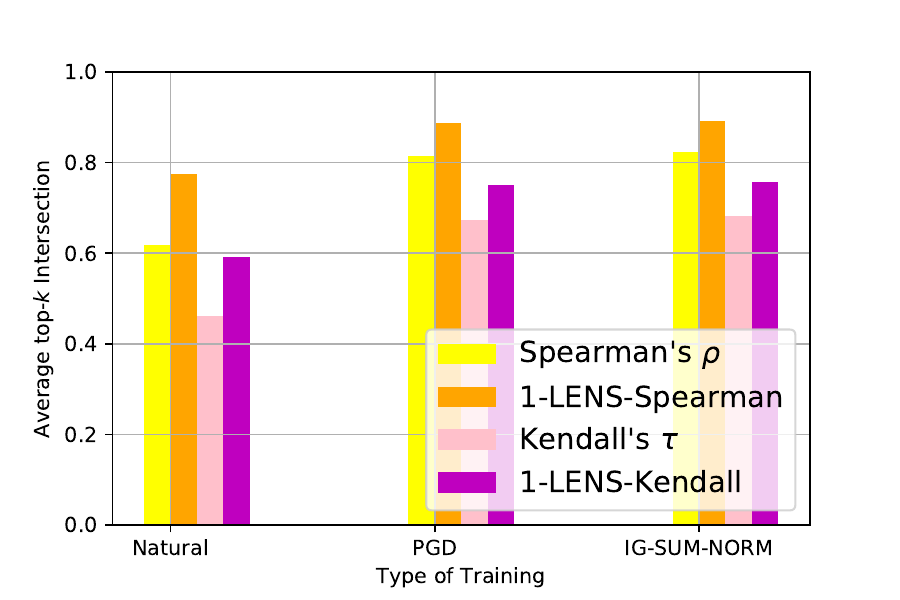}
    \caption{Fashion MNIST}
    \label{fmnist-rand-fig:b}
    \end{subfigure}\\
    \begin{subfigure}[b]{0.24\textwidth}
    \includegraphics[width=\textwidth]{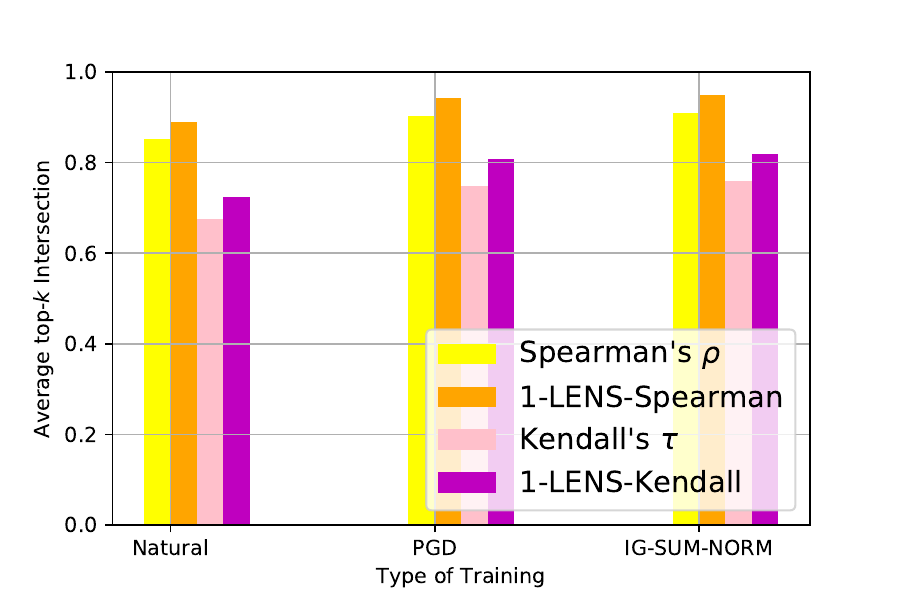}
    \caption{GTSRB}
    \label{gtsrb-rand-fig:c}
    \end{subfigure}
    \begin{subfigure}[b]{0.24\textwidth}
    \includegraphics[width=\textwidth]{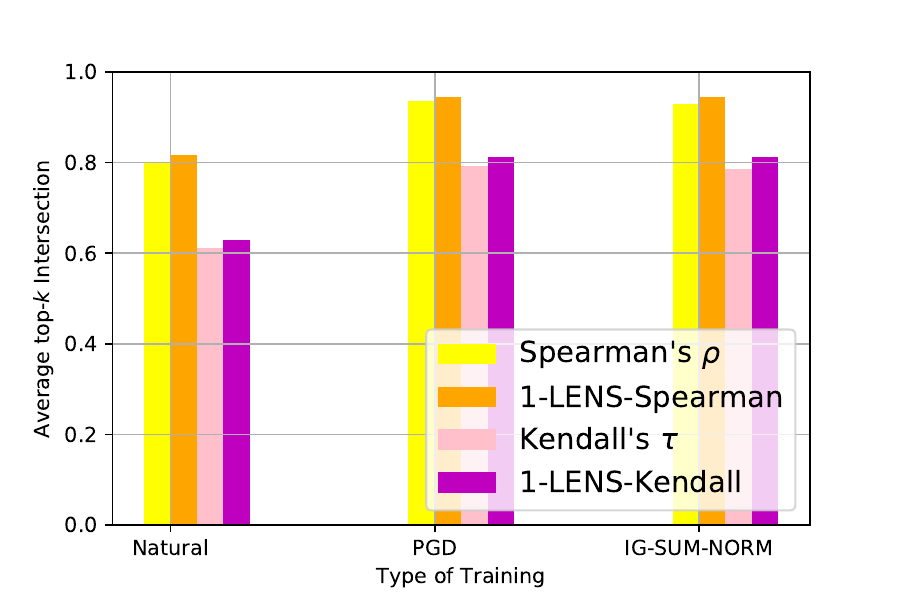}
    \caption{Flower}
    \label{flower-rand-fig:d}
    \end{subfigure}
\end{subfigure}
\caption{Average Kendall's $\tau$, Spearman's $\rho$, $1$-LENS-Kendall and $1$-LENS-Spearman used to measure the attributional robustness of IG on natrually trained, PGD-trained and IG-SUM-NORM trained models. The perturbation used is the {\bf top-$k$ attack} of \citet{ghorbani2018interpretation}. Shown for (a) MNIST, (b) Fashion MNIST, (c) GTSRB and (d) Flower datasets.}
\label{dataset-smooth-sp-ken-nat-pgd-rar}   
\end{figure}

\paragraph{Comparison of Spearman's $\rho$ and Kendall's $\tau$ with $1$-LENS-Spearman and $1$-LENS-Kendall.} 
Figure \ref{dataset-smooth-sp-ken-nat} compares Spearman's $\rho$ and Kendall's $\tau$ with $1$-LENS-Spearman and $1$-LENS-Kendall measures for attributional robustness. We observe that $1$-smoothing of attribution maps increases the corresponding Kendall's $\tau$ and Spearman's $\rho$ measures of attributional robustness, and this observation holds across all datasets in our experiments. As a result, we believe that $1$-LENS-Spearman and $1$-LENS-Kendall result in better or tighter attributional robustness measures than Spearman's $\rho$ and Kendall's $\tau$. Additional results in Figure \ref{dataset-smooth-sp-ken-nat-pgd-rar}. 

\paragraph{Modifying the attack of \citet{ghorbani2018interpretation} for $1$-LENS-prec@$k$ objective}

\begin{figure}[!ht]
\centering
    \includegraphics[width=0.24\textwidth]{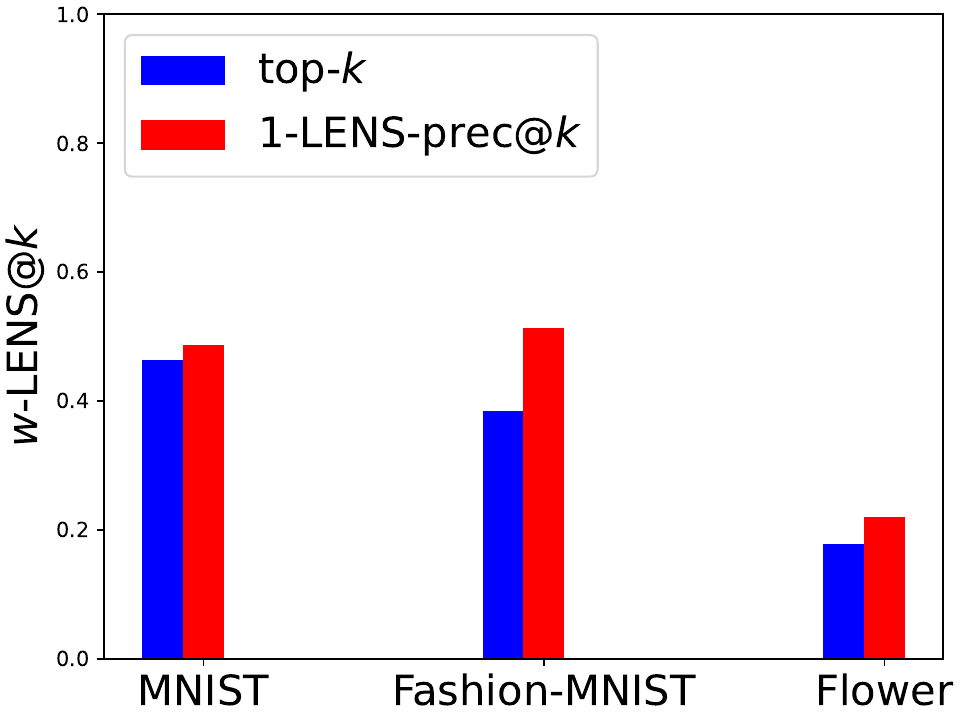}
\caption{Average top-$k$ intersection between IG(original image) and IG(perturbed image) on naturally trained models where the perturbation is obtained by incorporating $1$-LENS-prec@$k$ objective in the \citet{ghorbani2018interpretation} attack.
} 
\label{app:modify-ghorbani-with-lens}
\end{figure}

A natural question is whether the original top-$k$ attack of \citet{ghorbani2018interpretation} seem weaker under locality-senstitive robustness measures only because the attack was specifically constructed for a corresponding top-$k$ intersection objective. Since the construction of the attack in \citet{ghorbani2018interpretation} is modifiable for any similarity objective, we use $1$-LENS-prec@$k$ to construct a new attributional attack for $1$-LENS-prec@$k$ objective based on the $k \times k$ neighborhood of pixels. Surprisingly, we notice that it leads to a \emph{worse} attributional attack, if we measure its effectiveness using the top-$k$ intersection; see Figure \ref{app:modify-ghorbani-with-lens}. In other words, attributional attacks against locality-sensitive measures of attributional robustness are non-trivial and may require fundamentally different ideas.

\subsection{Experiments with Integrated Gradients} \label{app:all-ig-exp}
Below we present additional experimental results for Integrated Gradients (IG).

\begin{figure}[!ht]
\centering
\begin{subfigure}[b]{\textwidth}
    \begin{subfigure}[b]{0.24\textwidth}
    \includegraphics[width=\textwidth]{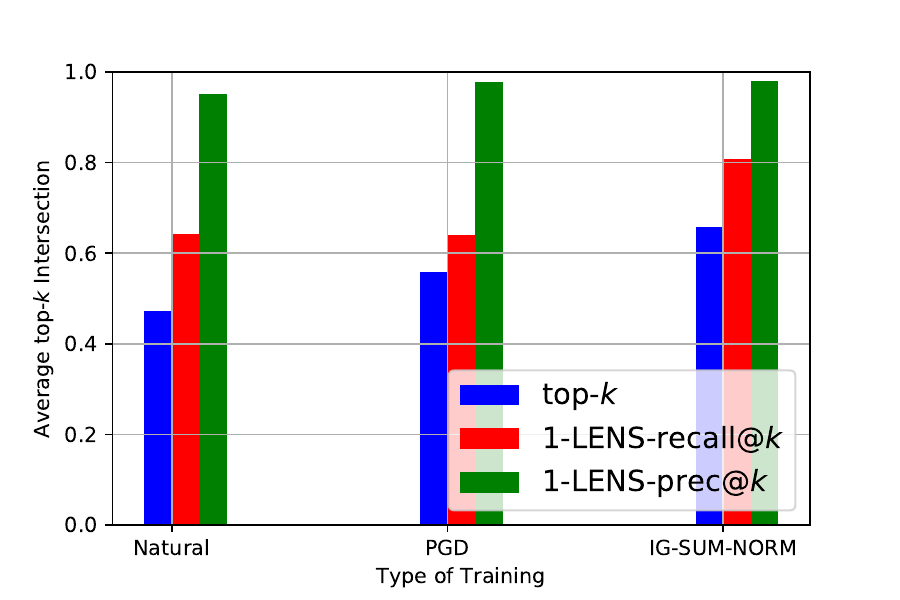}
    \caption{MNIST}
    \label{mnist-rand-fig:a}
    \end{subfigure}
    \begin{subfigure}[b]{0.24\textwidth}
    \includegraphics[width=\textwidth]{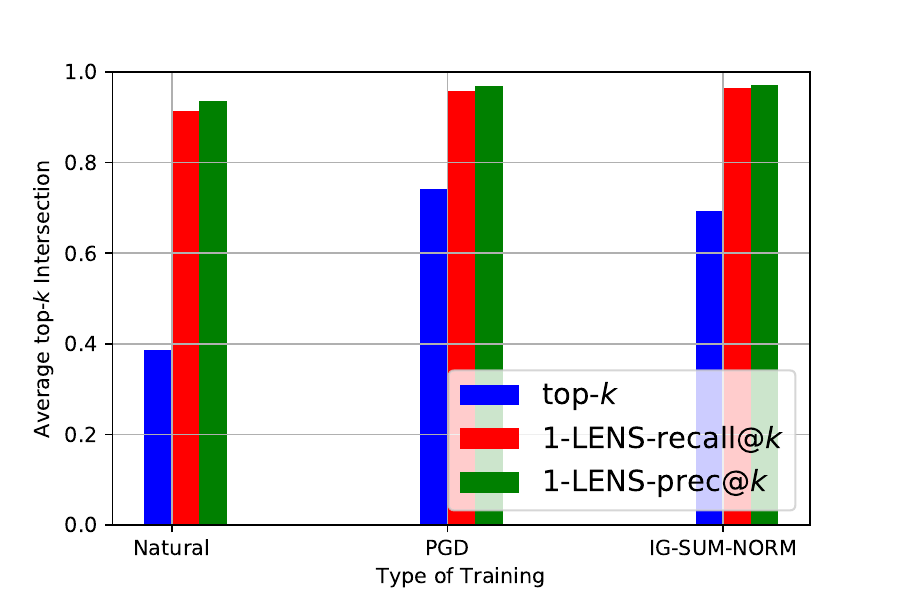}
    \caption{Fashion MNIST}
    \label{fmnist-rand-fig:b}
    \end{subfigure}\\
    \begin{subfigure}[b]{0.24\textwidth}
    \includegraphics[width=\textwidth]{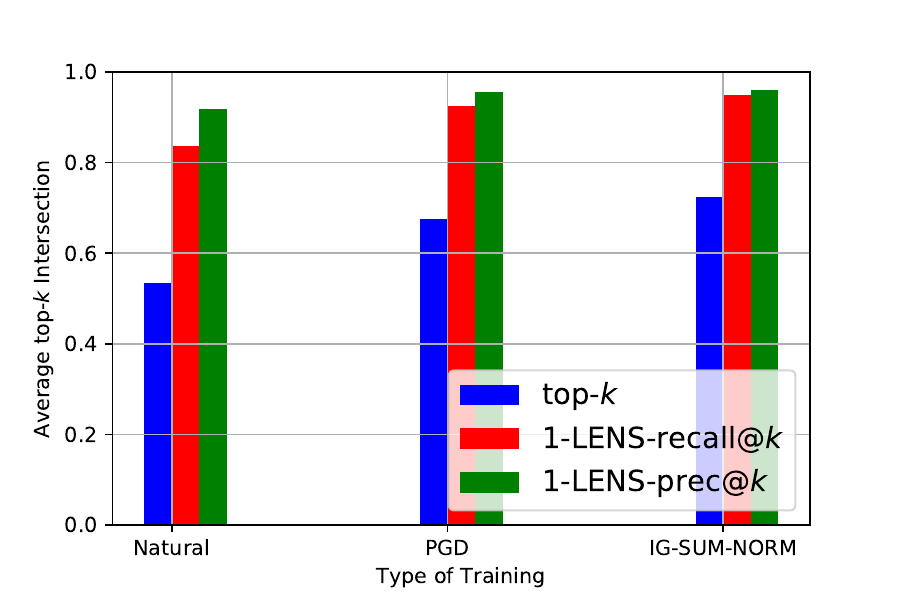}
    \caption{GTSRB}
    \label{gtsrb-rand-fig:c}
    \end{subfigure}
    \begin{subfigure}[b]{0.24\textwidth}
    \includegraphics[width=\textwidth]{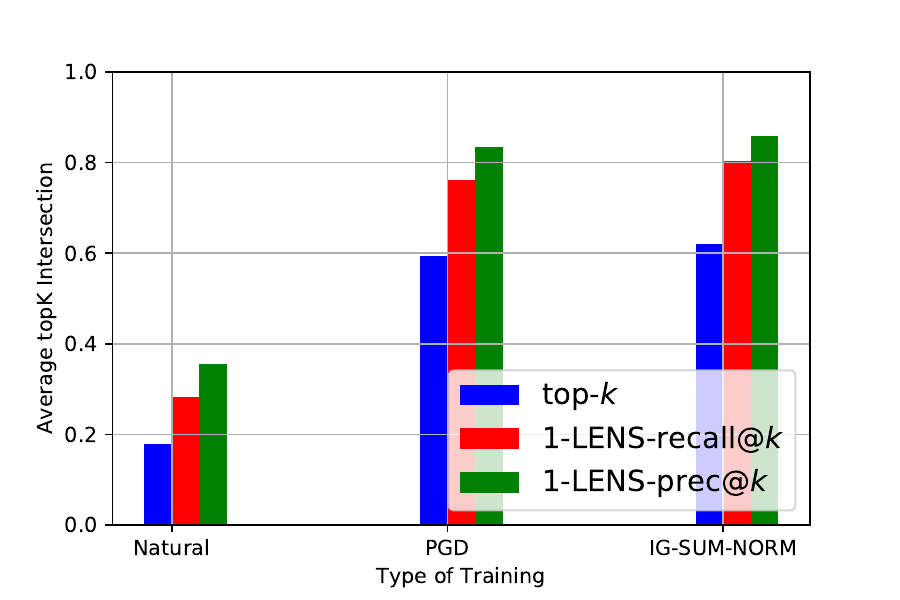}
    \caption{Flower}
    \label{flower-rand-fig:d}
    \end{subfigure}
\end{subfigure}
\caption{Average top-$k$ intersection, $1$-LENS-prec@$k$, $1$-LENS-recall@$k$ measured between IG(original image) and IG(perturbed image) for models that are naturally trained, PGD-trained and IG-SUM-NORM trained. The perturbation used is the {\bf top-$k$} attack of \cite{ghorbani2018interpretation}. Shown for (a) MNIST, (b) Fashion MNIST, (c) GTSRB and (d) Flower datasets.}
\label{dataset-topk-new-metric-st-nat-pgd-rar}     
\end{figure}

\begin{figure}[!ht]
\centering
\begin{subfigure}[b]{1.0\textwidth}
    \begin{subfigure}[b]{0.24\textwidth}
    \includegraphics[width=1.0\textwidth]{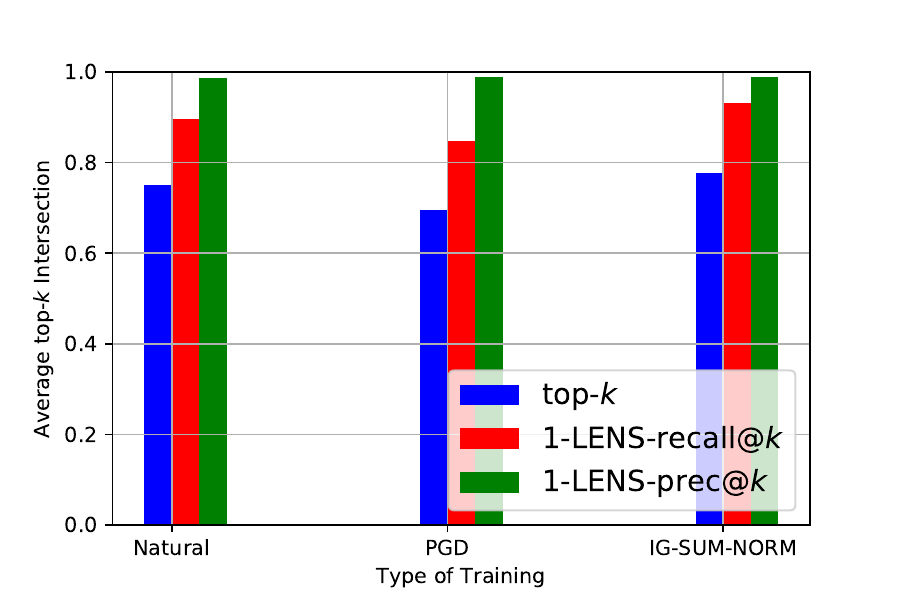}
    \caption{MNIST}
    \label{mnist-rand-fig:a}
    \end{subfigure}
    \begin{subfigure}[b]{0.24\textwidth}
    \includegraphics[width=1.0\textwidth]{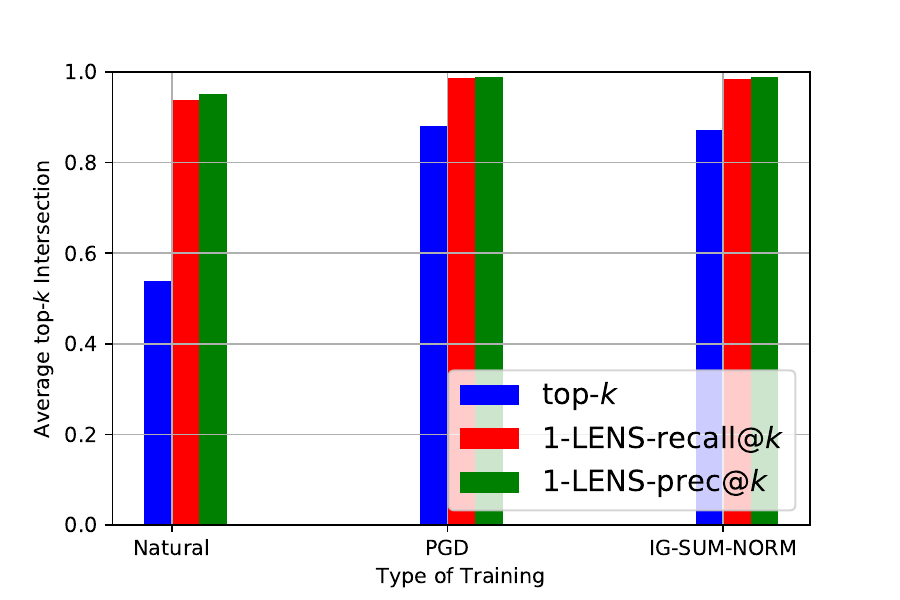}
    \caption{Fashion MNIST}
    \label{fmnist-rand-fig:b}
    \end{subfigure}\\
    \begin{subfigure}[b]{0.24\textwidth}
    \includegraphics[width=1.0\textwidth]{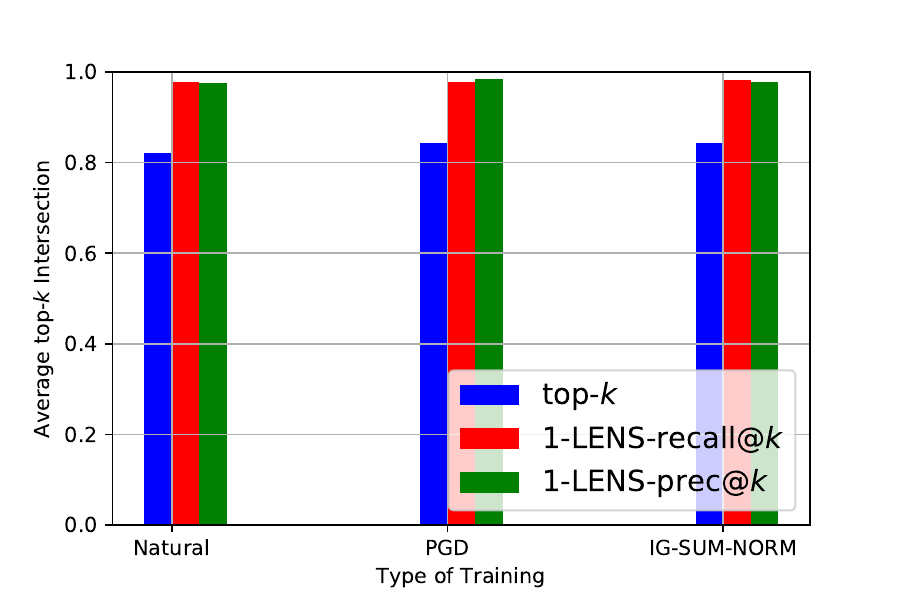}
    \caption{GTSRB}
    \label{gtsrb-rand-fig:c}
    \end{subfigure}
    \begin{subfigure}[b]{0.24\textwidth}
    \includegraphics[width=1.0\textwidth]{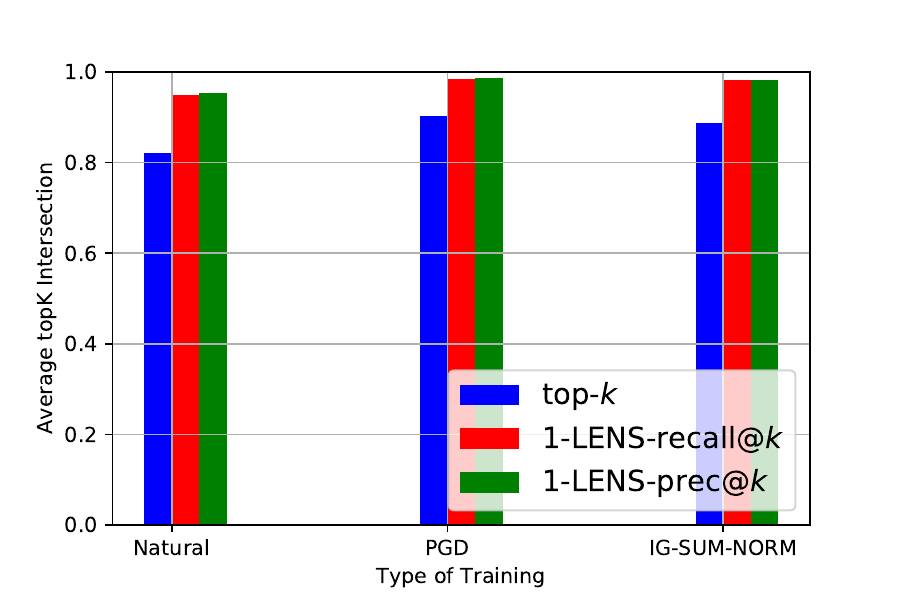}
    \caption{Flower}
    \label{flower-rand-fig:d}
    \end{subfigure}
\end{subfigure}
\caption{Attributional robustness of IG on naturally, PGD and IG-SUM-NORM trained models measured as top-$k$ intersection, $1$-LENS-prec@$k$ and $1$-LENS-recall@$k$ between the IG of the original images and the IG of their perturbations obtained by the {\bf random sign} attack \citep{ghorbani2018interpretation} across different datasets.} 
\label{rand-new-metric-training-method}     
\end{figure}

\begin{figure}[!ht]
\centering
\begin{subfigure}[b]{1.0\textwidth}
    \begin{subfigure}[b]{0.24\textwidth}
    \includegraphics[width=1.0\textwidth]{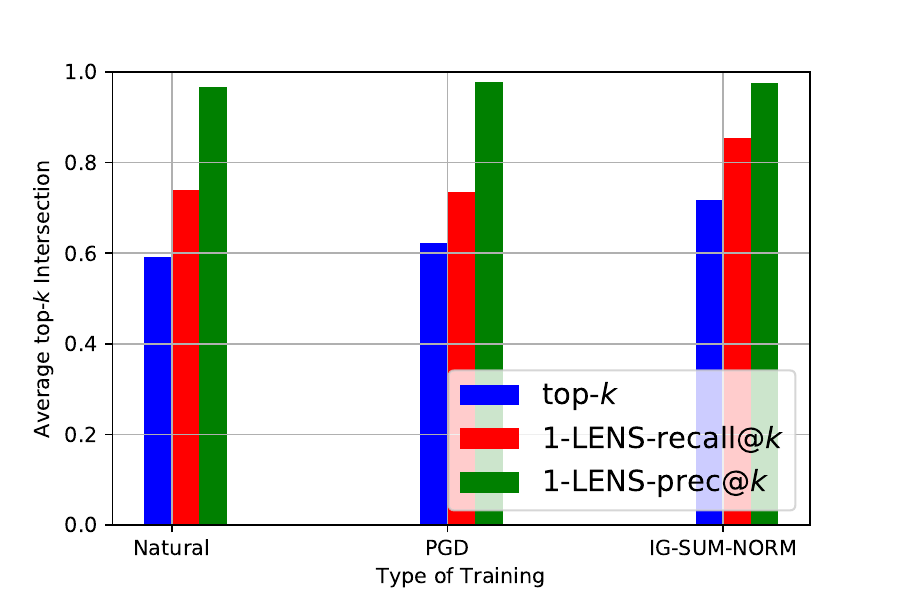}
    \caption{MNIST}
    \label{mnist-rand-fig:a}
    \end{subfigure}
    \begin{subfigure}[b]{0.24\textwidth}
    \includegraphics[width=1.0\textwidth]{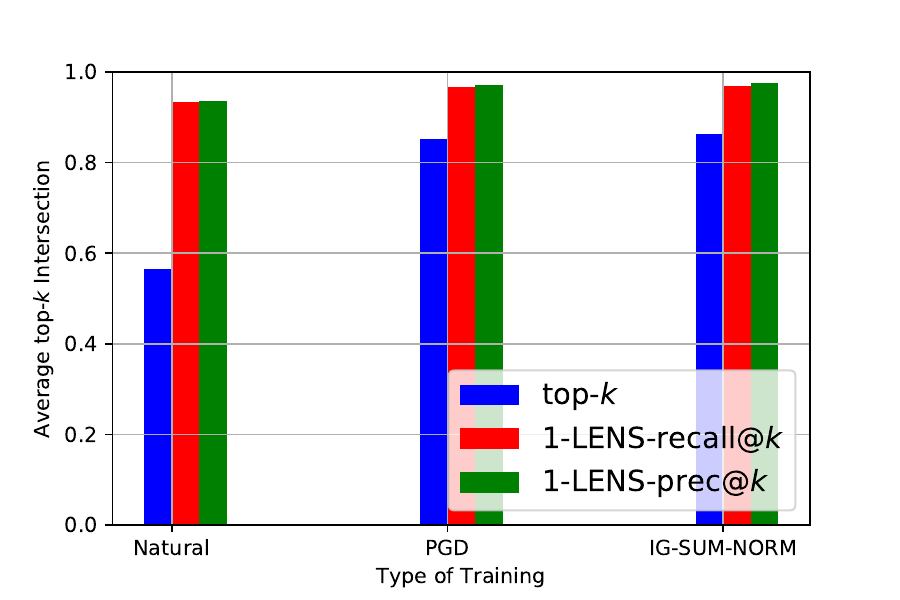}
    \caption{Fashion MNIST}
    \label{fmnist-rand-fig:b}
    \end{subfigure}\\
    \begin{subfigure}[b]{0.24\textwidth}
    \includegraphics[width=1.0\textwidth]{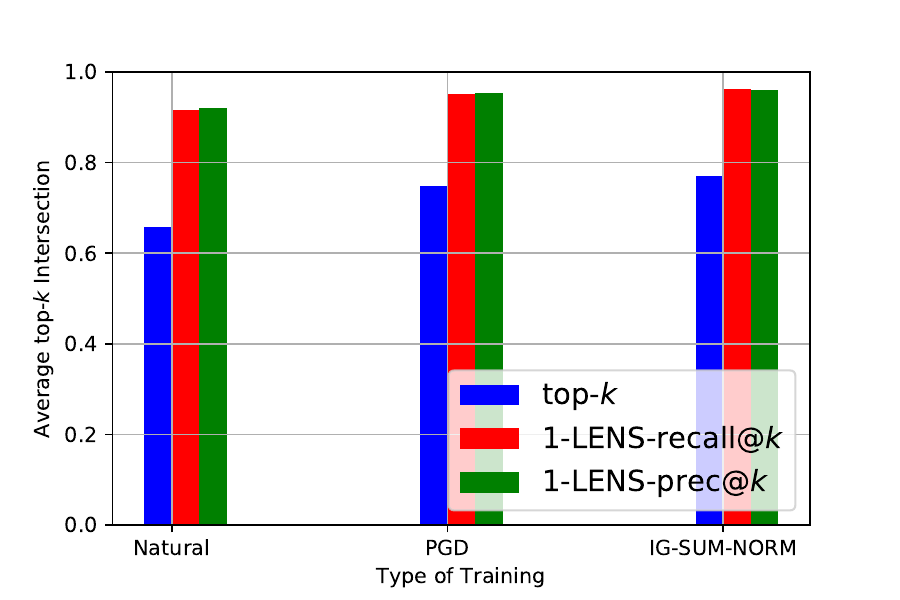}
    \caption{GTSRB}
    \label{gtsrb-rand-fig:c}
    \end{subfigure}
    \begin{subfigure}[b]{0.24\textwidth}
    \includegraphics[width=1.0\textwidth]{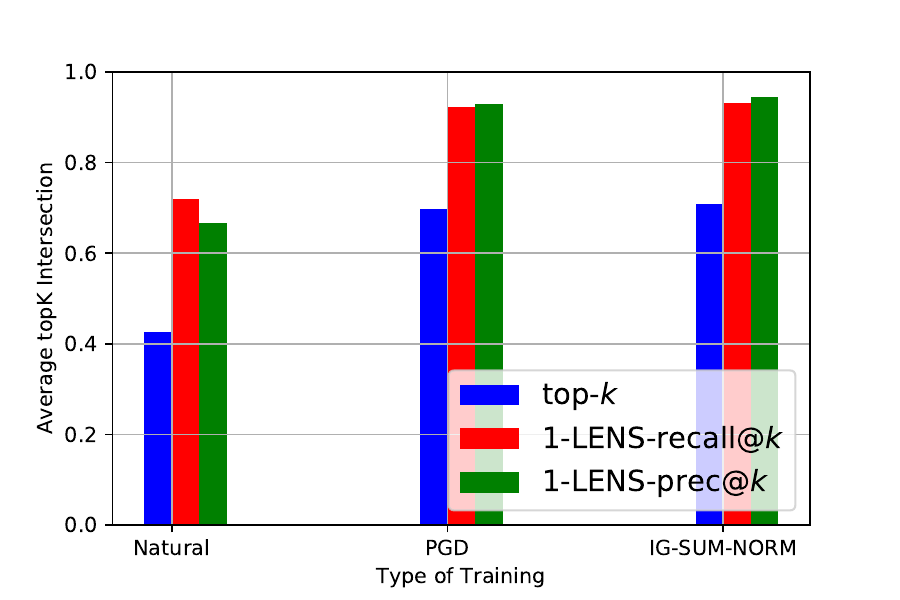}
    \caption{Flower}
    \label{flower-rand-fig:d}
    \end{subfigure}
\end{subfigure}
\caption{Attributional robustness of IG on naturally, PGD and IG-SUM-NORM trained models measured as top-$k$ intersection, $1$-LENS-prec@$k$ and $1$-LENS-recall@$k$ between the IG of the original images and the IG of their perturbations obtained by the {\bf mass center} attack \citep{ghorbani2018interpretation} across different datasets.} 
\label{mc-new-metric-training-method}     
\end{figure}


\begin{figure}[!ht]
\centering
\begin{subfigure}[b]{1.0\textwidth}
    \begin{subfigure}[b]{0.24\textwidth}
    \includegraphics[width=1.0\textwidth]{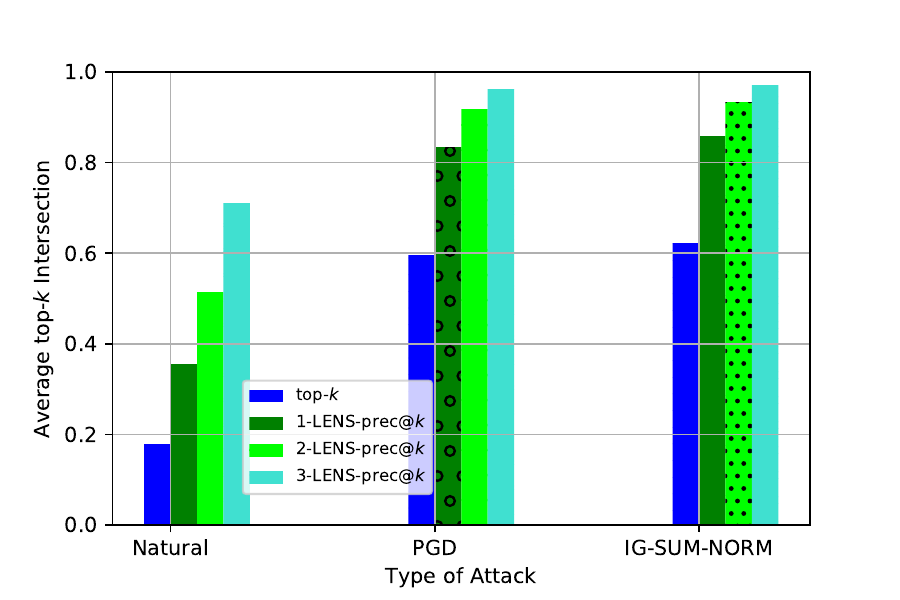}
    \caption{IG : top-$k$}
    \label{mnist-rand-fig:a}
    \end{subfigure}
    \begin{subfigure}[b]{0.24\textwidth}
    \includegraphics[width=1.0\textwidth]{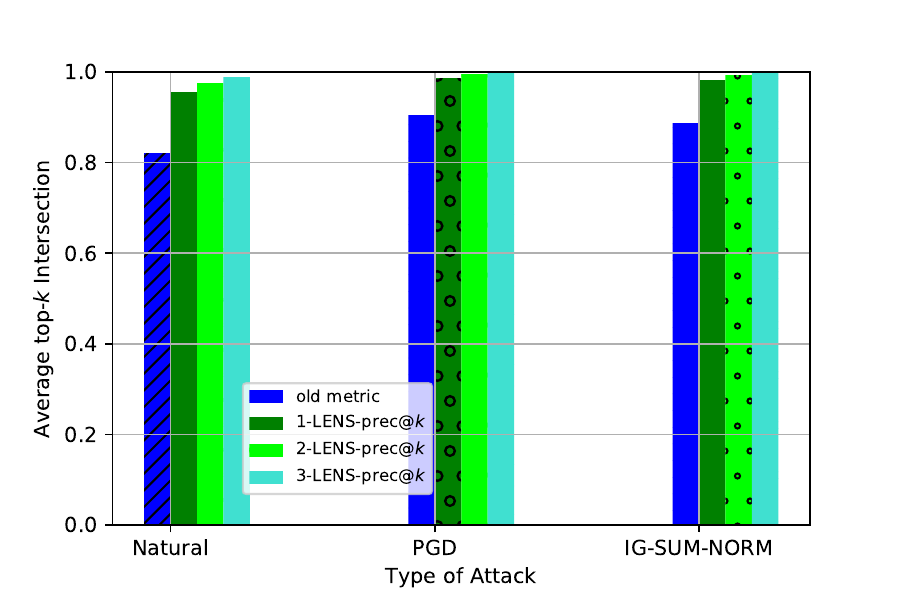}
    \caption{IG : random}
    \label{mnist-rand-fig:b}
    \end{subfigure}\\
    \begin{subfigure}[b]{0.24\textwidth}
    \includegraphics[width=1.0\textwidth]{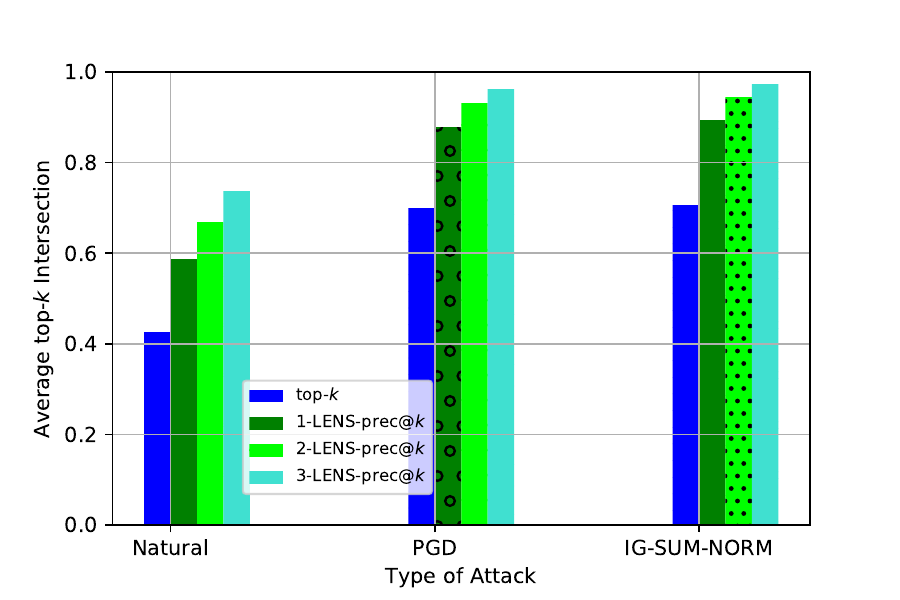}
    \caption{IG : center of mass}
    \label{mnist-rand-fig:c}
    \end{subfigure}
\end{subfigure}
\caption{Attributional robustness of IG on naturally, PGD and IG-SUM-NORM trained models measured as top-$k$ intersection and $w$-LENS-prec@$k$ between the IG of the original images and the IG of their perturbations. Perturbations are obtained by the top-$k$ attack and the mass center attack \citep{ghorbani2018interpretation} as well as a random perturbation. The plots show the effect of varying $w$ on Flower dataset.}
\label{flower-topk-new-metric-training-method}     
\end{figure}

\clearpage

\subsection{Experiments with Simple Gradients} \label{app:all-sg-exp}
We observe that our conclusions about Integrated Gradients (IG) continue to hold qualitatively, even if we replace IG with Simple Gradients as our attribution method.

\begin{figure}[!ht]
\centering
\begin{subfigure}[b]{1.0\textwidth}
    \begin{subfigure}[b]{0.24\textwidth}
    \includegraphics[width=1.0\textwidth]{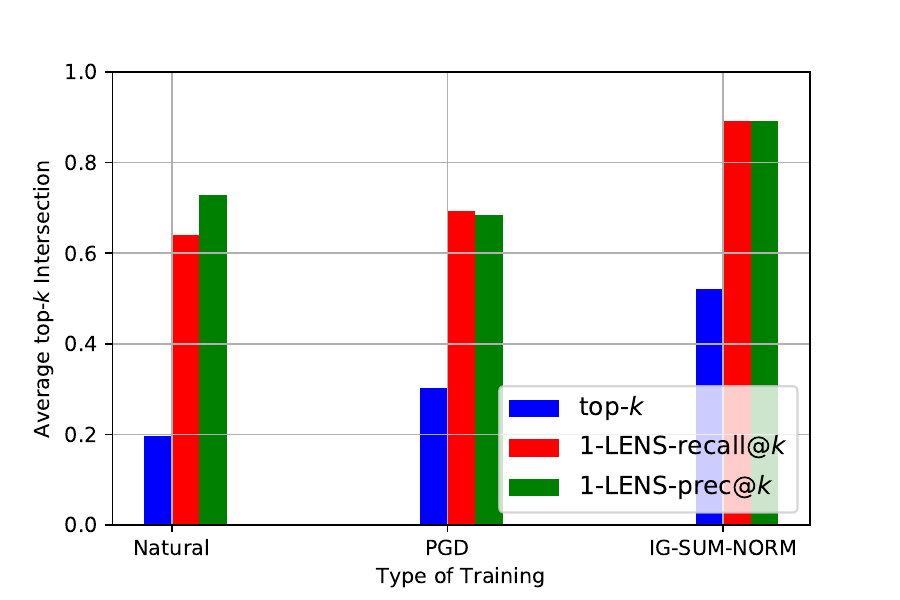}
    \caption{MNIST}
    \label{mnist-rand-fig:a}
    \end{subfigure}
    \begin{subfigure}[b]{0.24\textwidth}
    \includegraphics[width=1.0\textwidth]{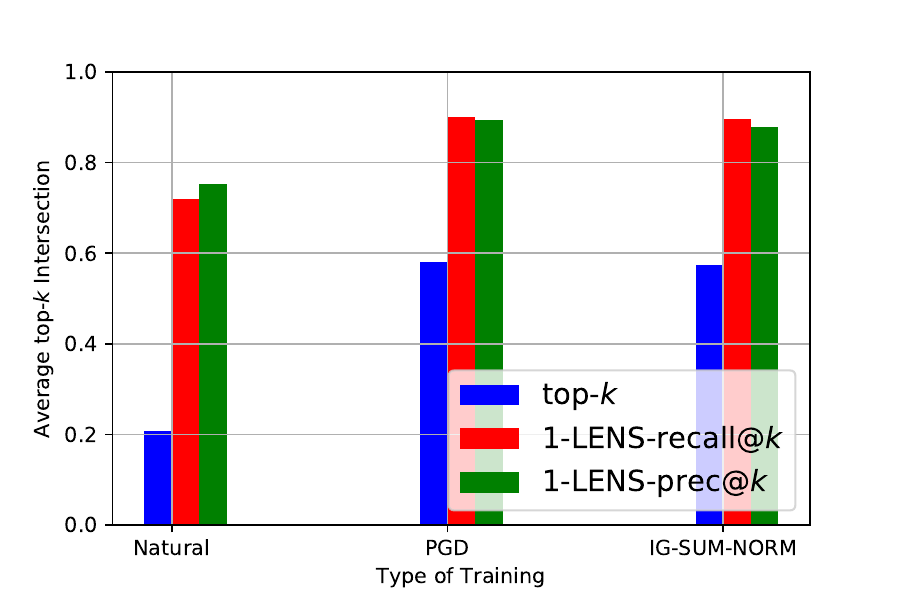}
    \caption{Fashion MNIST}
    \label{fmnist-rand-fig:b}
    \end{subfigure}\\
    \begin{subfigure}[b]{0.24\textwidth}
    \includegraphics[width=1.0\textwidth]{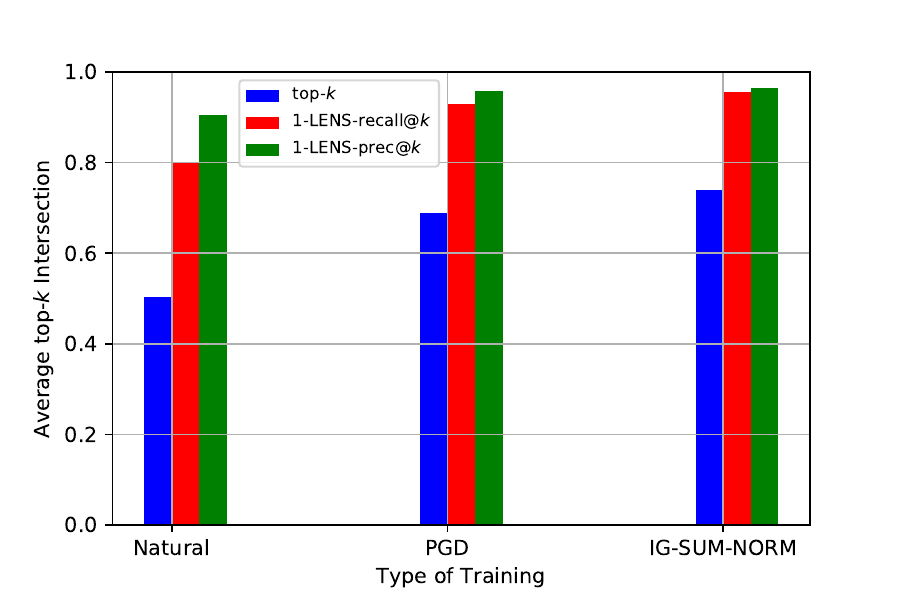}
    \caption{GTSRB}
    \label{gtsrb-rand-fig:c}
    \end{subfigure}
    \begin{subfigure}[b]{0.24\textwidth}
    \includegraphics[width=1.0\textwidth]{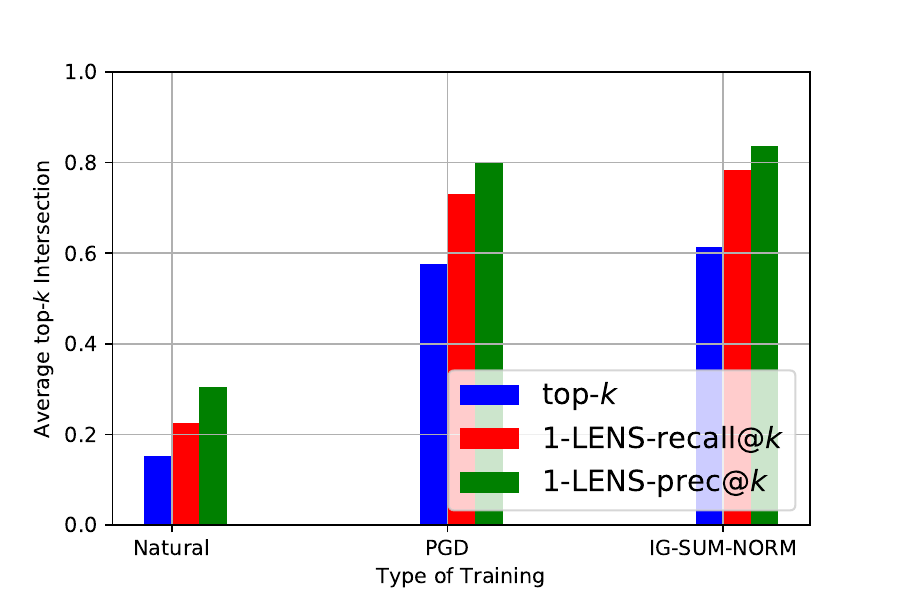}
    \caption{Flower}
    \label{flower-rand-fig:d}
    \end{subfigure}
\end{subfigure}
\caption{Attributional robustness of Simple Gradients on naturally, PGD and IG-SUM-NORM trained models measured as top-$k$ intersection, $1$-LENS-prec@$k$ and $1$-LENS-recall@$k$ between the Simple Gradient of the original images and the Simple Gradient of their perturbations obtained by the {\bf top-$k$} attack \citep{ghorbani2018interpretation} across different datasets.} 
\label{topk-new-metric-training-method-sg}     
\end{figure}

\begin{figure}[!ht]
\centering
\begin{subfigure}[b]{1.0\textwidth}
    \begin{subfigure}[b]{0.24\textwidth}
    \includegraphics[width=1.0\textwidth]{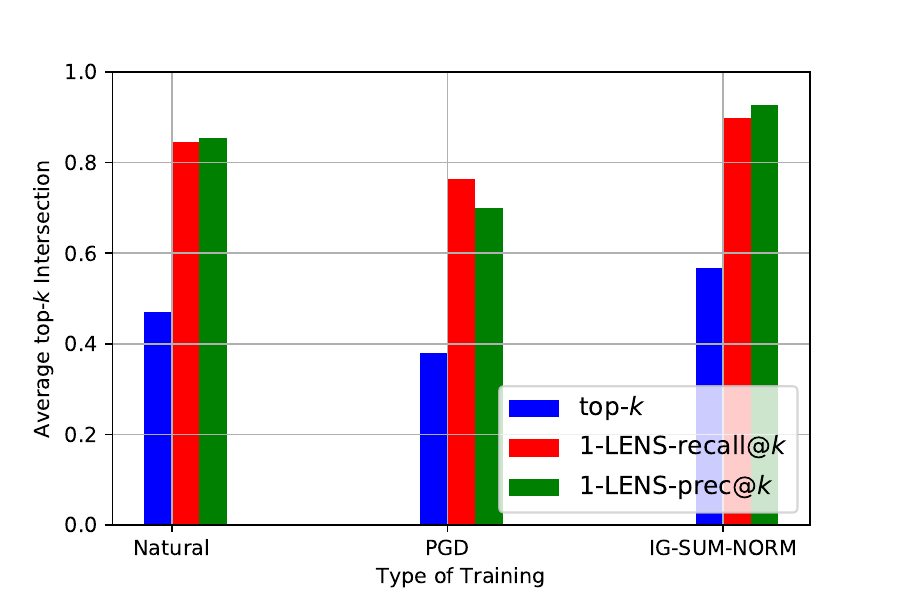}
    \caption{MNIST}
    \label{mnist-rand-fig:a}
    \end{subfigure}
    \begin{subfigure}[b]{0.24\textwidth}
    \includegraphics[width=1.0\textwidth]{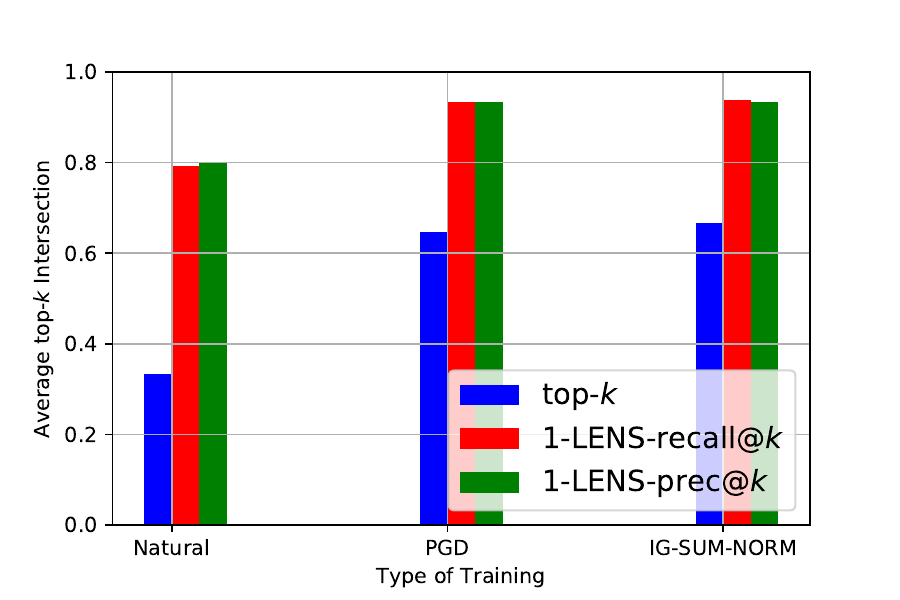}
    \caption{Fashion MNIST}
    \label{fmnist-rand-fig:b}
    \end{subfigure}\\
    \begin{subfigure}[b]{0.24\textwidth}
    \includegraphics[width=1.0\textwidth]{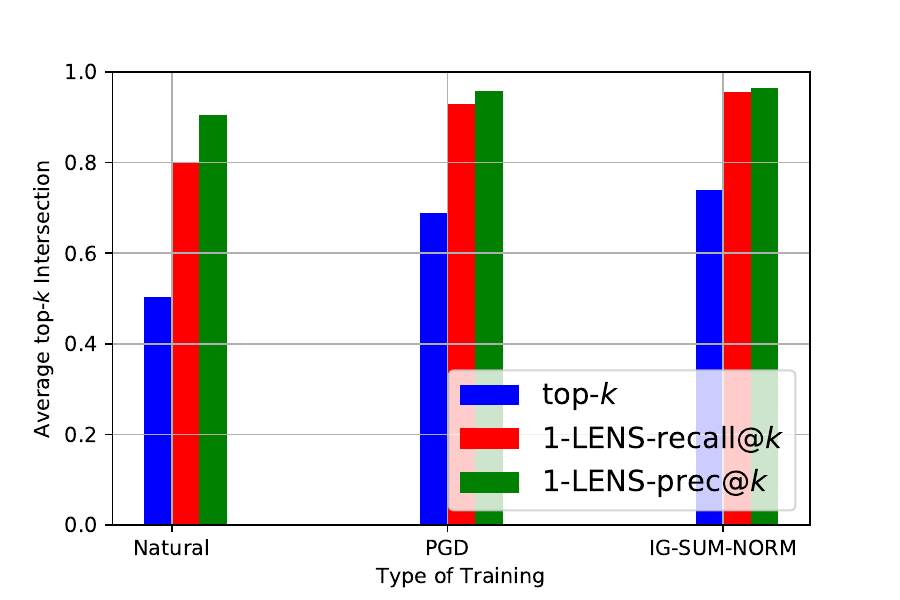}
    \caption{GTSRB}
    \label{gtsrb-rand-fig:c}
    \end{subfigure}
    \begin{subfigure}[b]{0.24\textwidth}
    \includegraphics[width=1.0\textwidth]{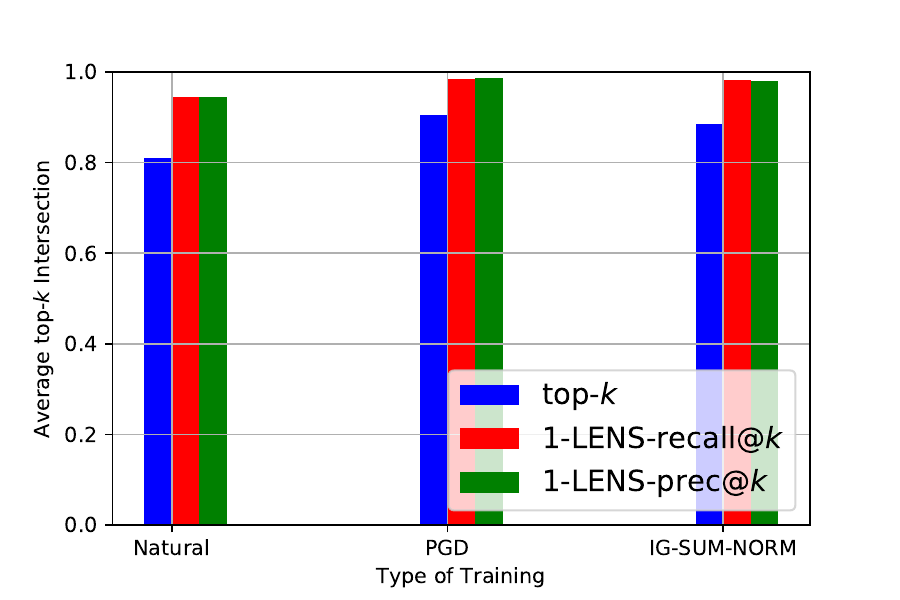}
    \caption{Flower}
    \label{flower-rand-fig:d}
    \end{subfigure}
\end{subfigure}
\caption{Attributional robustness of Simple Gradients on naturally, PGD and IG-SUM-NORM trained models measured as top-$k$ intersection, $1$-LENS-prec@$k$ and $1$-LENS-recall@$k$ between the Simple Gradient of the original images and the Simple Gradient of their perturbations obtained by the {\bf random sign} attack \citep{ghorbani2018interpretation} across different datasets.} 
\label{rand-new-metric-training-method-sg}     
\end{figure}

\begin{figure}[!ht]
\centering
\begin{subfigure}[b]{1.0\textwidth}
    \begin{subfigure}[b]{0.24\textwidth}
    \includegraphics[width=1.0\textwidth]{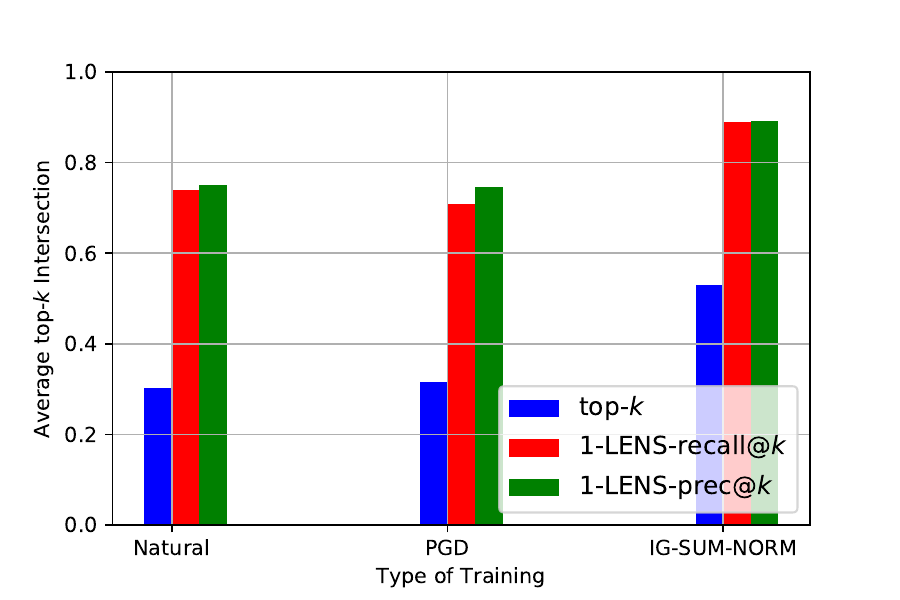}
    \caption{MNIST}
    \label{mnist-rand-fig:a}
    \end{subfigure}
    \begin{subfigure}[b]{0.24\textwidth}
    \includegraphics[width=1.0\textwidth]{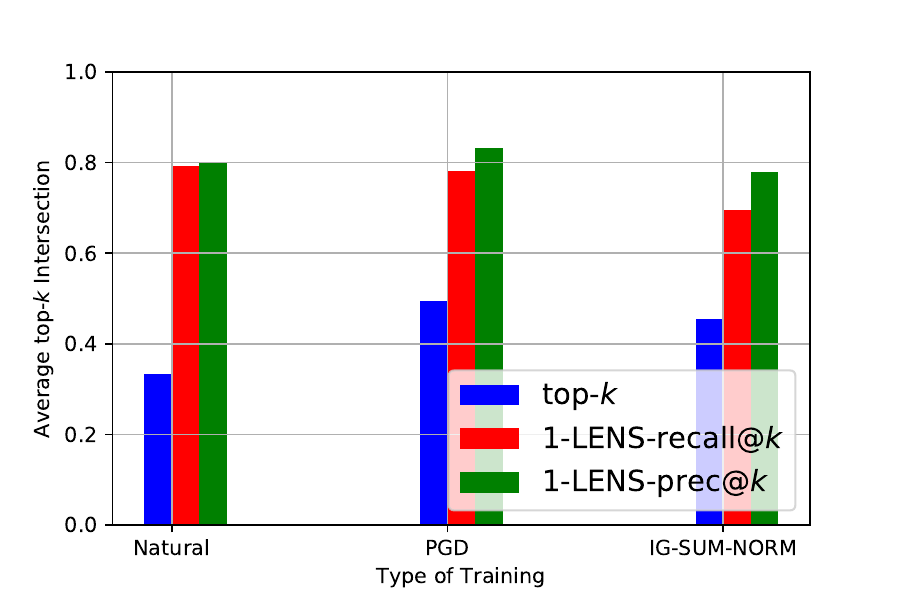}
    \caption{Fashion MNIST}
    \label{fmnist-rand-fig:b}
    \end{subfigure}\\
    \begin{subfigure}[b]{0.24\textwidth}
    \includegraphics[width=1.0\textwidth]{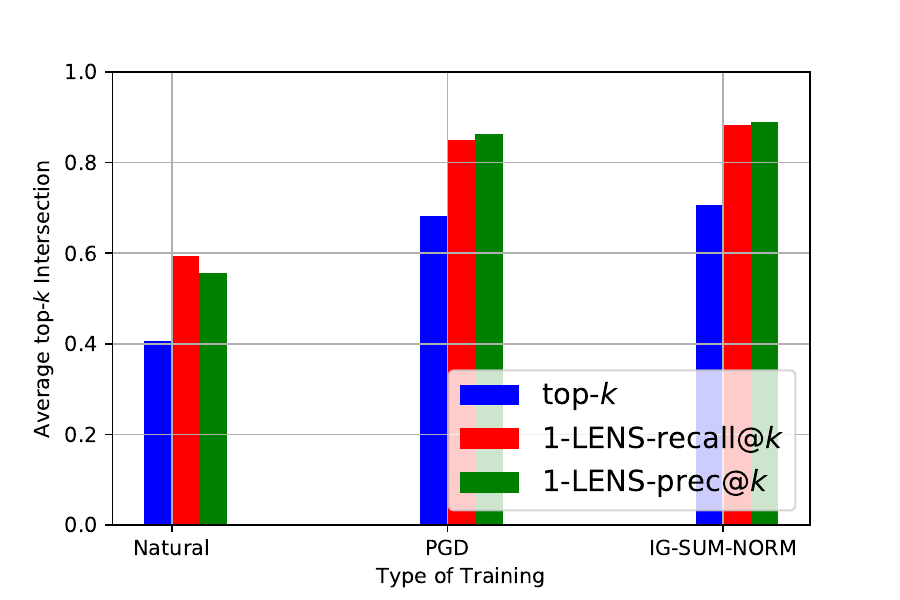}
    \caption{Flower}
    \label{flower-rand-fig:d}
    \end{subfigure}
\end{subfigure}
\caption{Attributional robustness of Simple Gradients on naturally, PGD and IG-SUM-NORM trained models measured as top-$k$ intersection, $1$-LENS-prec@$k$ and $1$-LENS-recall@$k$ between the Simple Gradient of the original images and the Simple Gradient of their perturbations obtained by the {\bf mass center} attack \citep{ghorbani2018interpretation} across different datasets.} 
\label{mc-new-metric-training-method-sg}     
\end{figure}

\begin{figure}[!ht]
\centering
\begin{subfigure}[b]{1.0\textwidth}
    \begin{subfigure}[b]{0.24\textwidth}
    \includegraphics[width=1.0\textwidth]{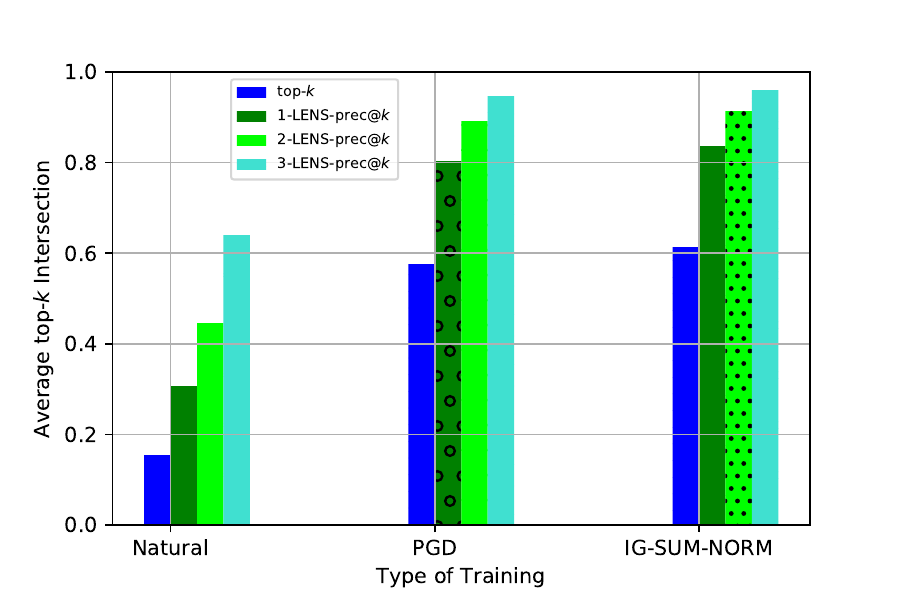}
    \caption{SG : top-$k$}
    \label{mnist-rand-fig:a}
    \end{subfigure}
    \begin{subfigure}[b]{0.24\textwidth}
    \includegraphics[width=1.0\textwidth]{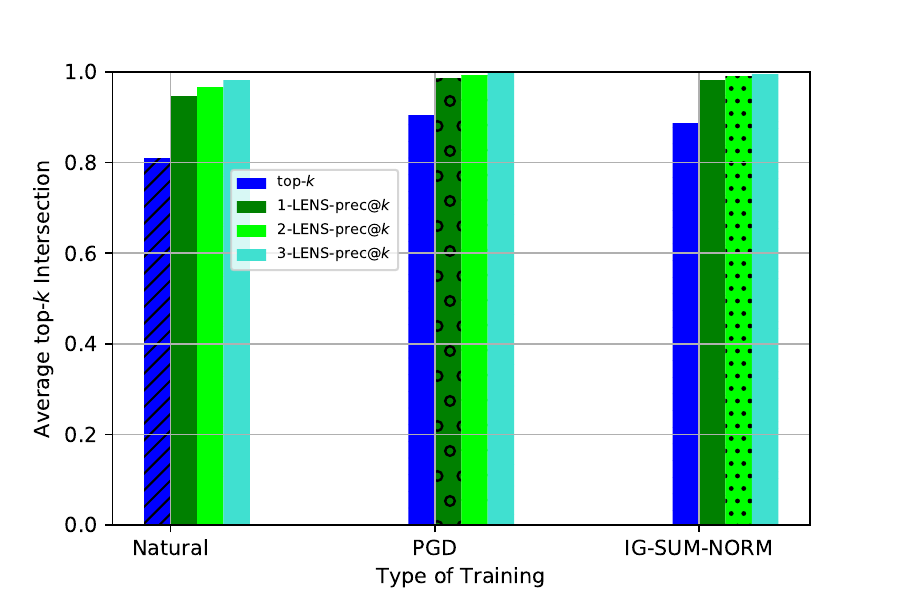}
    \caption{SG : random}
    \label{mnist-rand-fig:b}
    \end{subfigure}\\
    \begin{subfigure}[b]{0.24\textwidth}
    \includegraphics[width=1.0\textwidth]{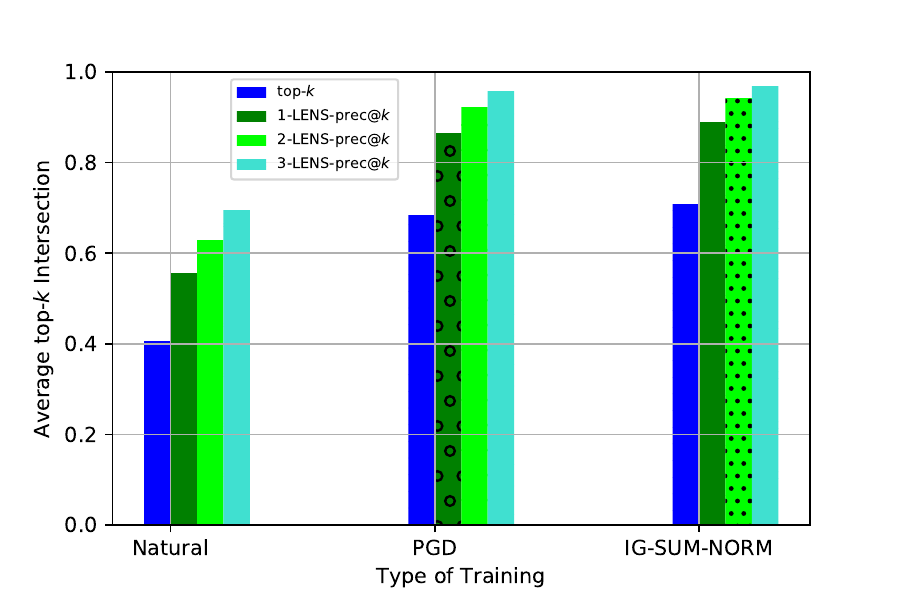}
    \caption{SG : center of mass}
    \label{mnist-rand-fig:c}
    \end{subfigure}
\end{subfigure}
\caption{Attributional robustness of Simple Gradients on naturally, PGD and IG-SUM-NORM trained models measured as top-$k$ intersection and $w$-LENS-prec@$k$ between the IG of the original images and the IG of their perturbations. Perturbations are obtained by the top-$k$ attack and the mass center attack \citep{ghorbani2018interpretation} as well as a random perturbation. The plots show the effect of varying $w$ on Flower dataset.}
\label{flower-topk-new-metric-training-method-sg}     
\end{figure}

\clearpage

\subsection{Experiments with Other Explanation Methods} \label{app:all-other-exp}

In this section we present all the results with other explanation methods other than IG and SG shown in previous sections with naturally and PGD trained models using mainly random sign perturbation.

\begin{figure*}[t]
 \centering
 \includegraphics[width=0.80\textwidth]{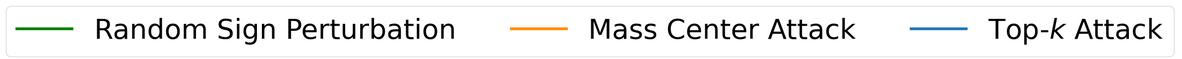}\\
     \includegraphics[width=0.23\textwidth]{./plots/imagenet_gh_lens_sg_orig_pert_all.pdf}
     \includegraphics[width=0.23\textwidth]{./plots/imagenet_gh_lens_ig_orig_pert_all.pdf}
     \includegraphics[width=0.23\textwidth]{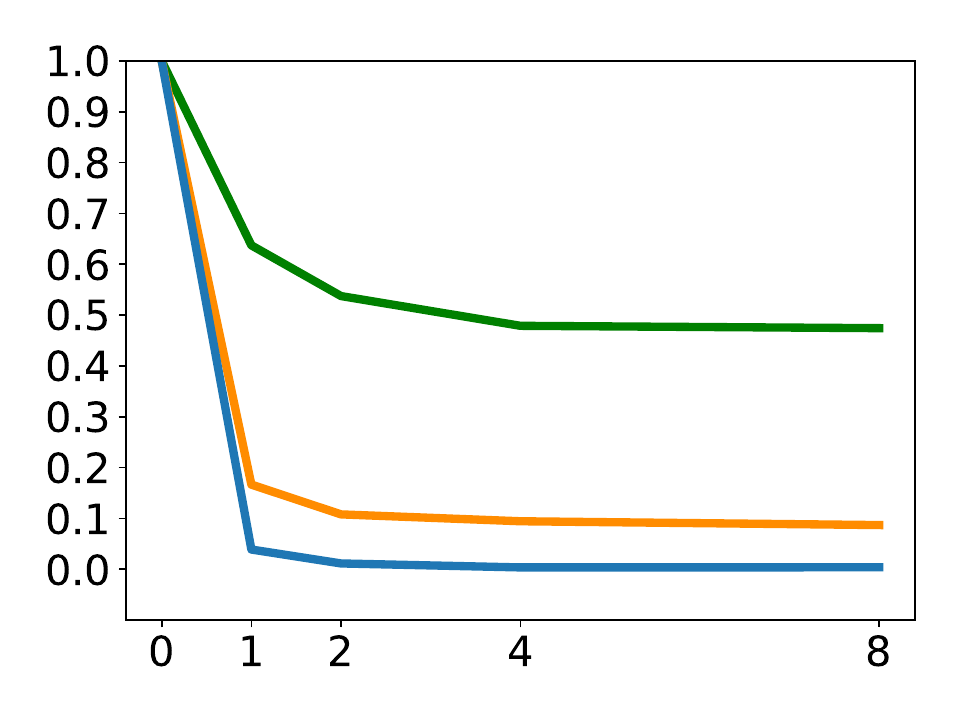}
     \includegraphics[width=0.23\textwidth]{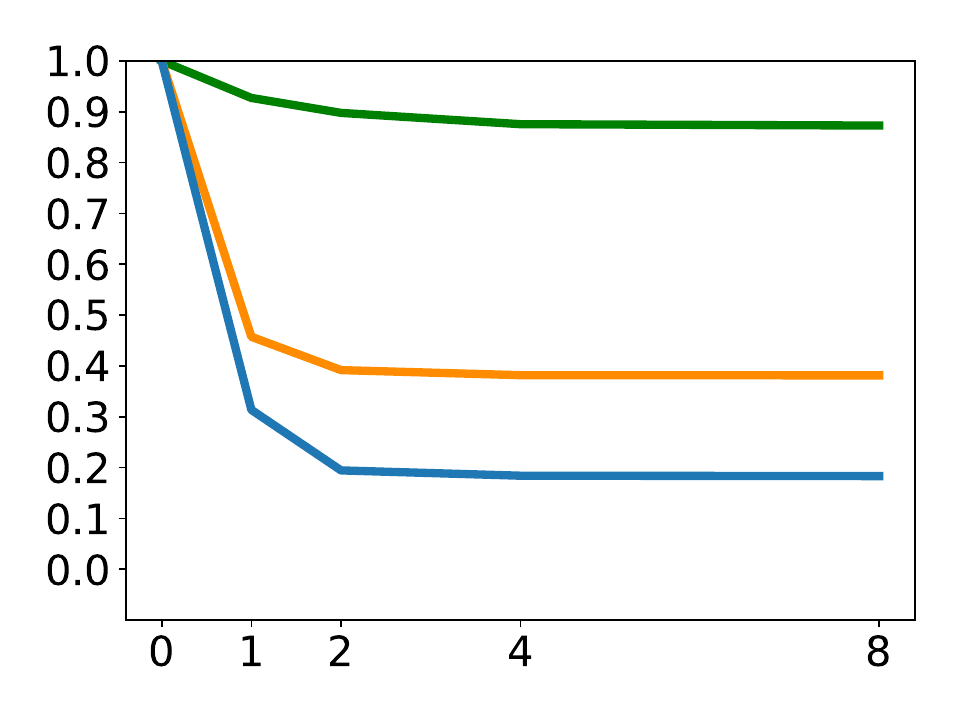}\\
     
     \includegraphics[width=0.23\textwidth]{./plots/imagenet_gh_lens_sg_orig_pert_all_lens.pdf}
     \includegraphics[width=0.23\textwidth]{./plots/imagenet_gh_lens_ig_orig_pert_all_lens.pdf}         
     \includegraphics[width=0.23\textwidth]{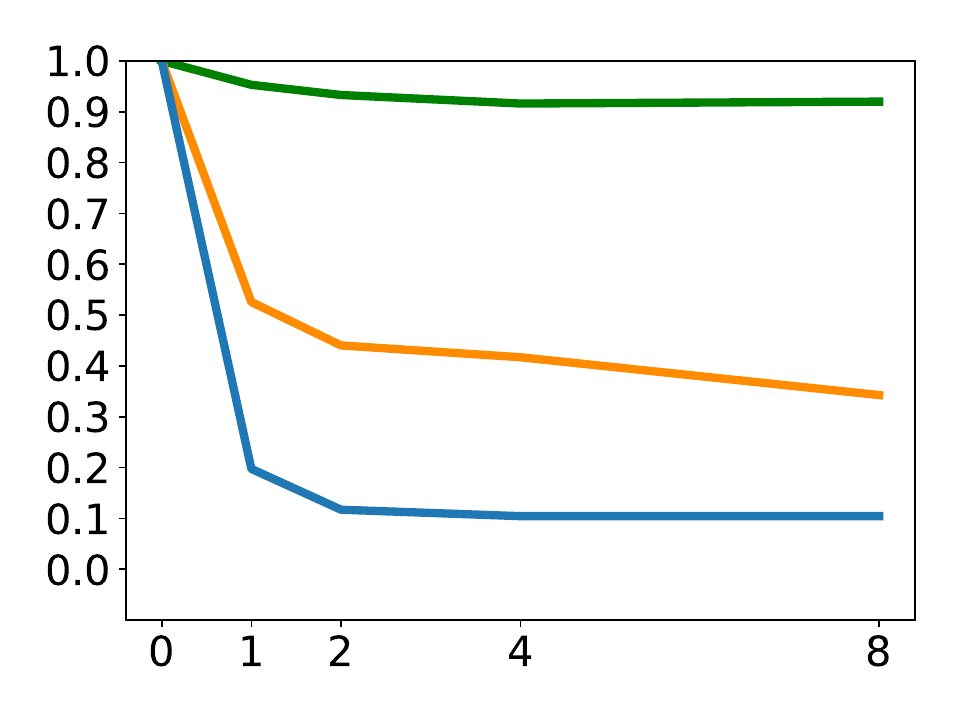}
     \includegraphics[width=0.23\textwidth]{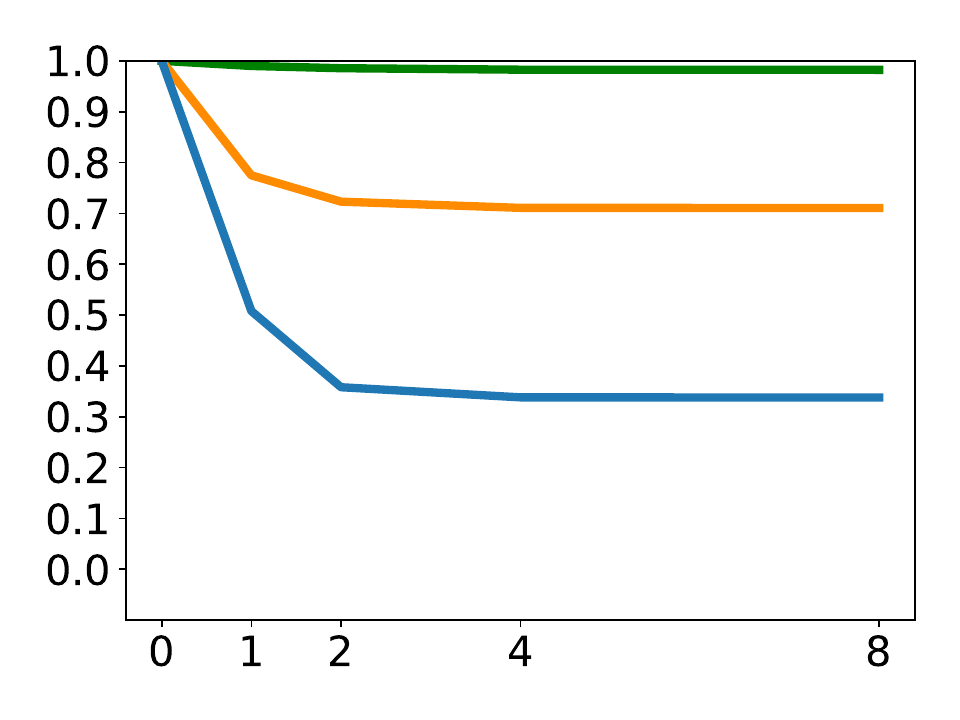}\\     
     \includegraphics[width=0.23\textwidth]{./plots/imagenet_gh_lens_sg_orig_pert_all_div.pdf}     
     \includegraphics[width=0.23\textwidth]{./plots/imagenet_gh_lens_ig_orig_pert_all_div.pdf}     
     \includegraphics[width=0.23\textwidth]{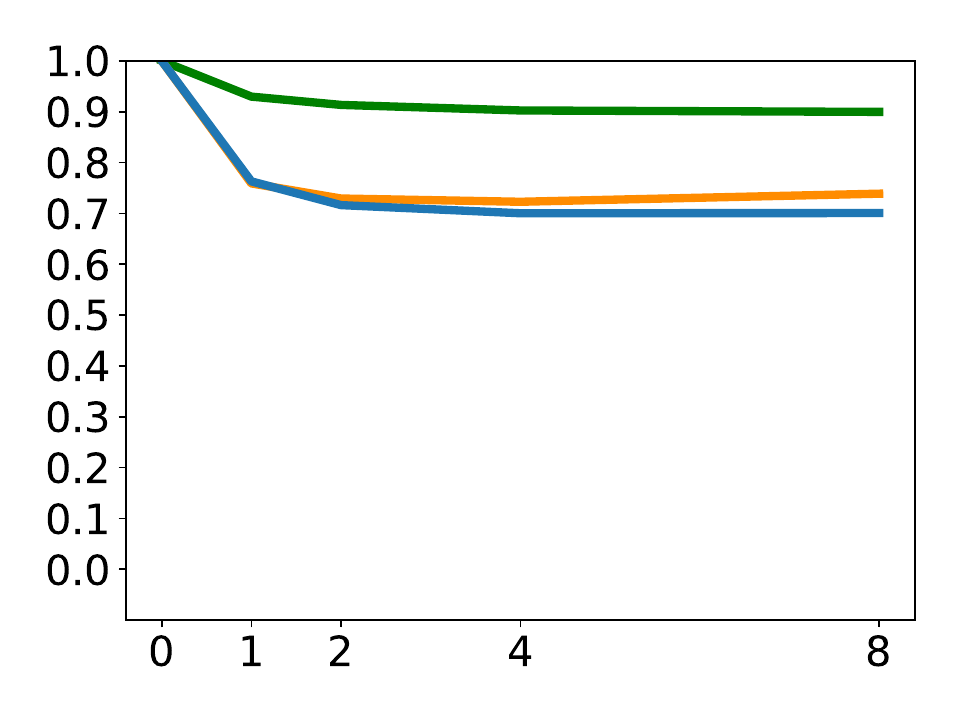}    
     \includegraphics[width=0.23\textwidth]{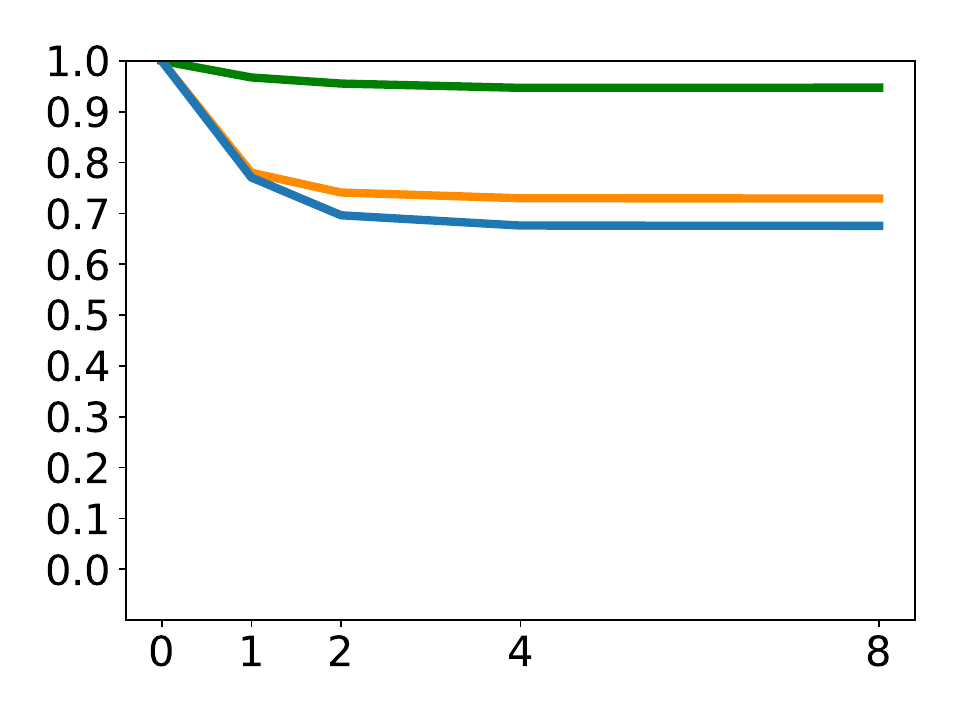}\\
     \includegraphics[width=0.20\textwidth]{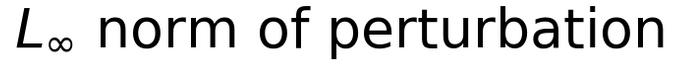}
    \caption{\footnotesize Similar to Figure \ref{imagenet-ghorbani-with-lens-div}, From top to bottom, we plot average top-$k$ intersection (currently used metric, \textbf{\textit{top}}), $3$-LENS-recall@$k$ and $3$-LENS-recall@$k$-div (proposed metrics, \textbf{\textit{middle}} and \textbf{\textit{bottom}} respectively) against attributional attack perturbations for four attribution methods of a SqueezeNet model (as used by \citet{ghorbani2018interpretation}) on Imagenet: \textit{(left)} Simple Gradients (SG), \textit{(center left)} Integrated Gradients (IG), \textit{(center right)} DeepLift, \textit{(right)} Deep Taylor Decomposition. We use $k=1000$ with an $\ell_{\infty}$-norm attack and three attack variants proposed by \citet{ghorbani2018interpretation}. Evidently, the proposed metrics show more robustness under the same attacks.}
    \vspace{-8pt}
\label{imagenet-ghorbani-with-lens-div-with-dl}     
\end{figure*}

\begin{figure}[!ht]
\centering
\begin{subfigure}[b]{1.0\textwidth}
    \begin{subfigure}[b]{0.15\textwidth}
    \includegraphics[width=1.0\textwidth]{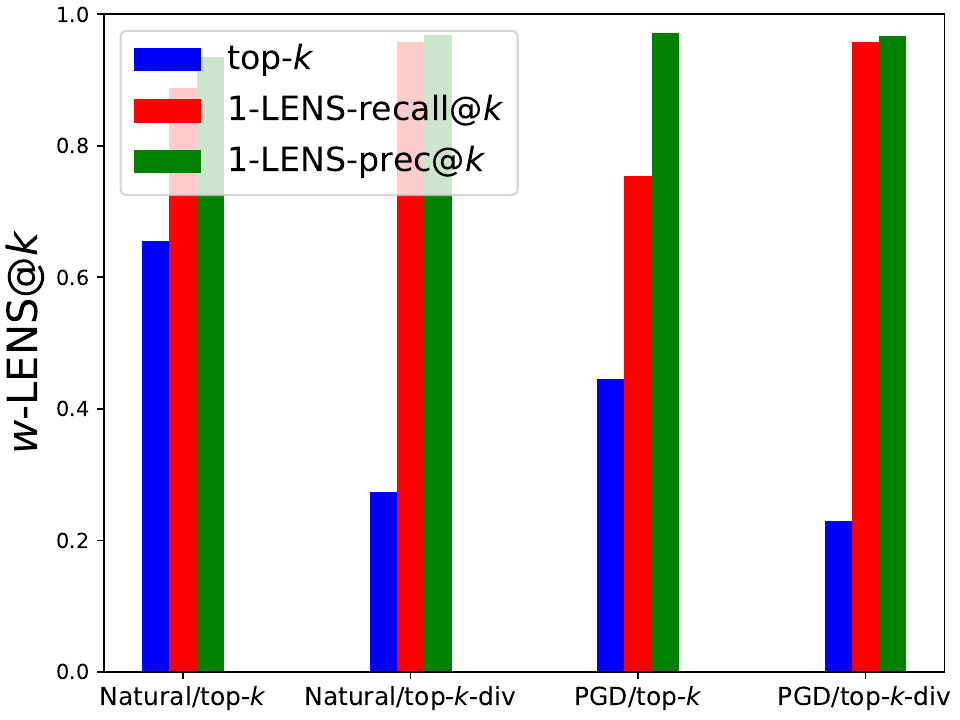}
    \caption{MNIST}
    \label{mnist-rand-fig:a}
    \end{subfigure}
    \begin{subfigure}[b]{0.15\textwidth}
    \includegraphics[width=1.0\textwidth]{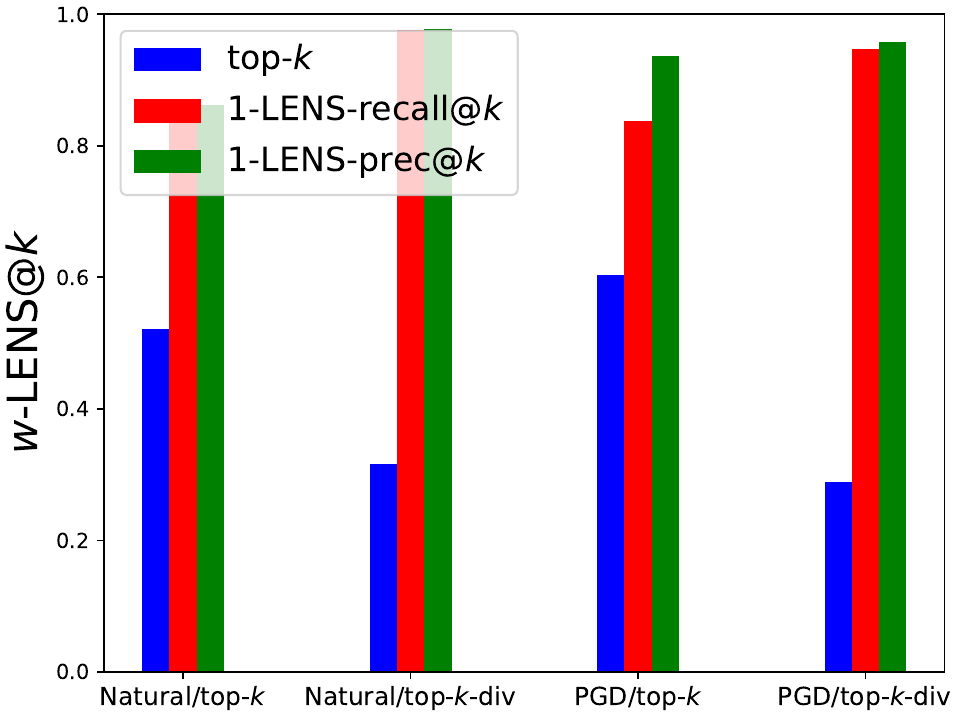}
    \caption{Fashion MNIST}
    \label{fmnist-rand-fig:b}
    \end{subfigure}
    \begin{subfigure}[b]{0.15\textwidth}
    \includegraphics[width=1.0\textwidth]{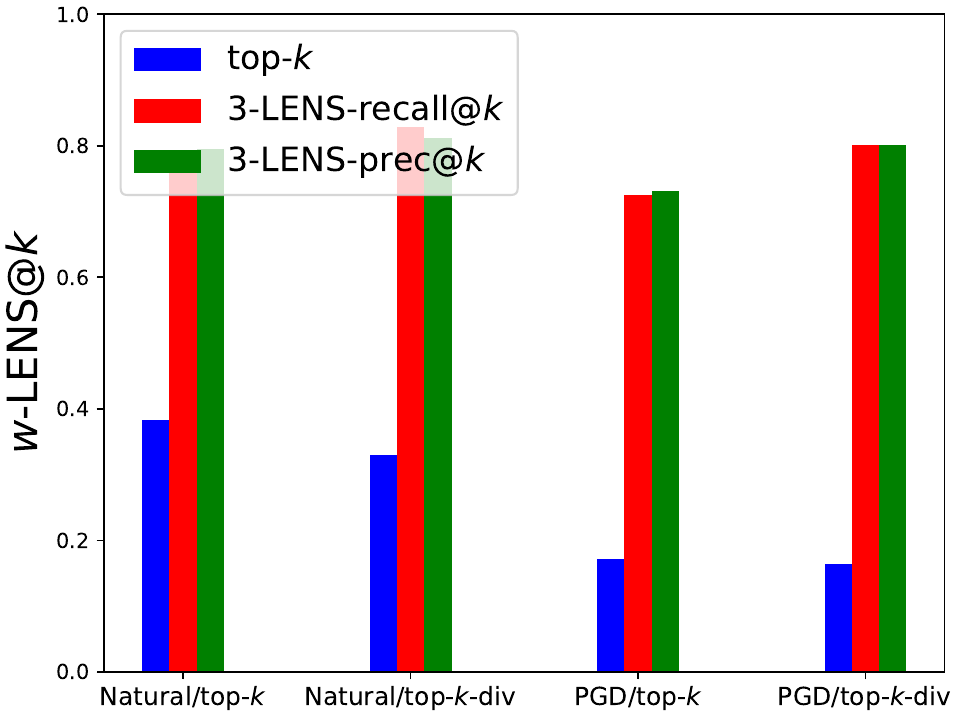}
    \caption{ImageNet}
    \label{flower-rand-fig:d}
    \end{subfigure}
\end{subfigure}
\caption{Attributional robustness of Simple Gradients(SG) on naturally and PGD trained models measured as top-$k$ intersection, $w$-LENS-prec@$k$ and $w$-LENS-recall@$k$ between the {\bf Simple Gradients(SG)} of the original images and of their perturbations obtained by the random sign perturbation across different datasets.} 
\label{all-datasets-topk-div-both-lens-sg}
\end{figure}

\begin{figure}[!ht]
\centering
\begin{subfigure}[b]{1.0\textwidth}
    \begin{subfigure}[b]{0.15\textwidth}
    \includegraphics[width=1.0\textwidth]{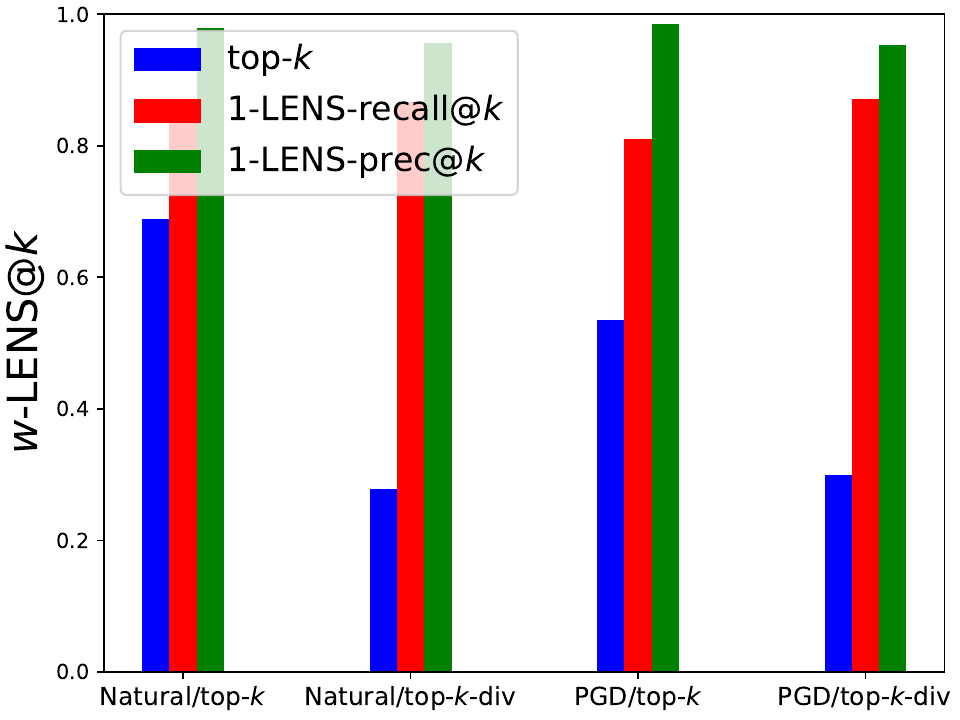}
    \caption{MNIST}
    \label{mnist-rand-fig:a}
    \end{subfigure}
    \begin{subfigure}[b]{0.15\textwidth}
    \includegraphics[width=1.0\textwidth]{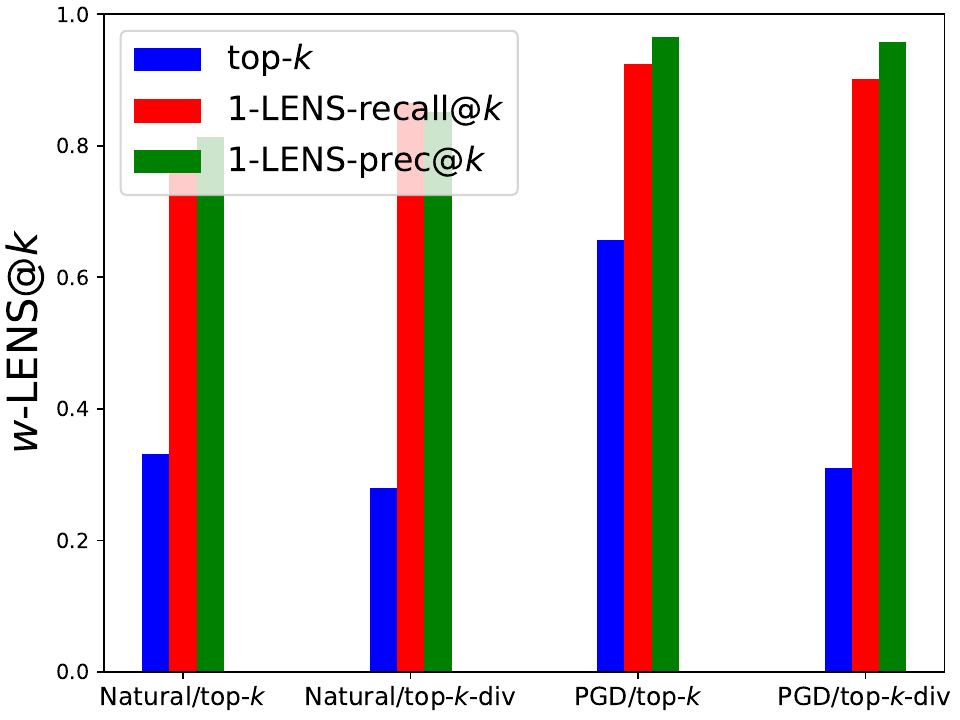}
    \caption{Fashion MNIST}
    \label{fmnist-rand-fig:b}
    \end{subfigure}
    \begin{subfigure}[b]{0.15\textwidth}
    \includegraphics[width=1.0\textwidth]{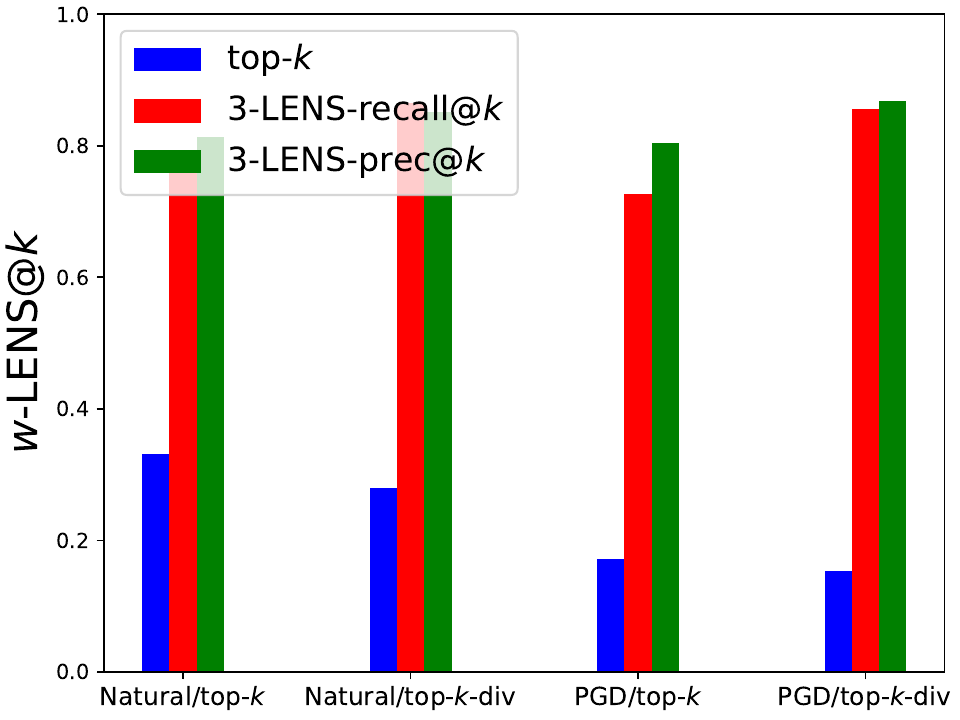}
    \caption{ImageNet}
    \label{flower-rand-fig:d}
    \end{subfigure}
\end{subfigure}
\caption{Attributional robustness of Image $\times$ Gradients on naturally and PGD trained models measured as top-$k$ intersection, $w$-LENS-prec@$k$ and $w$-LENS-recall@$k$ between the {\bf Image $\times$ Gradients} of the original images and of their perturbations obtained by the random sign perturbation across different datasets.} 
\label{all-datasets-topk-div-both-lens-ixg}
\end{figure}

\begin{figure}[!ht]
\centering
\begin{subfigure}[b]{1.0\textwidth}
    \begin{subfigure}[b]{0.15\textwidth}
    \includegraphics[width=1.0\textwidth]{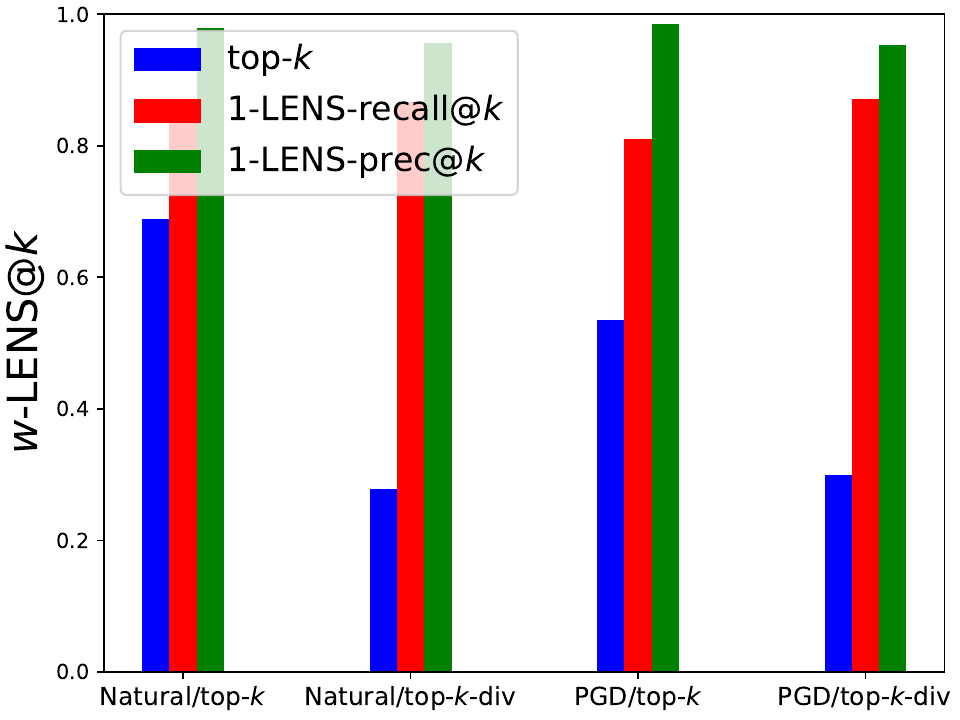}
    \caption{MNIST}
    \label{mnist-rand-fig:a}
    \end{subfigure}
    \begin{subfigure}[b]{0.15\textwidth}
    \includegraphics[width=1.0\textwidth]{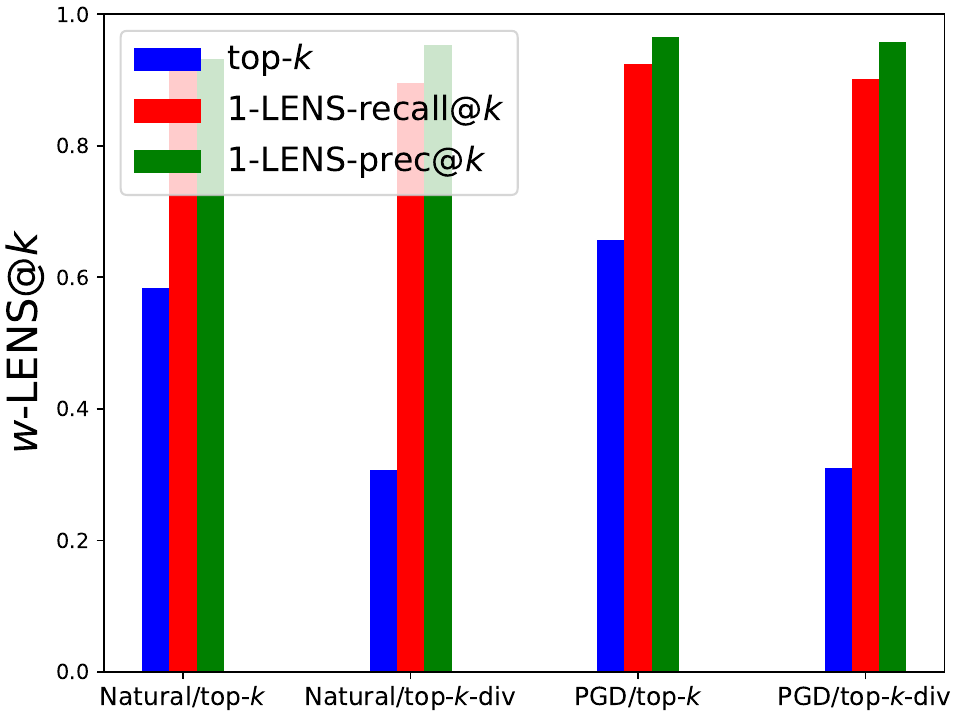}
    \caption{Fashion MNIST}
    \label{fmnist-rand-fig:b}
    \end{subfigure}
    \begin{subfigure}[b]{0.15\textwidth}
    \includegraphics[width=1.0\textwidth]{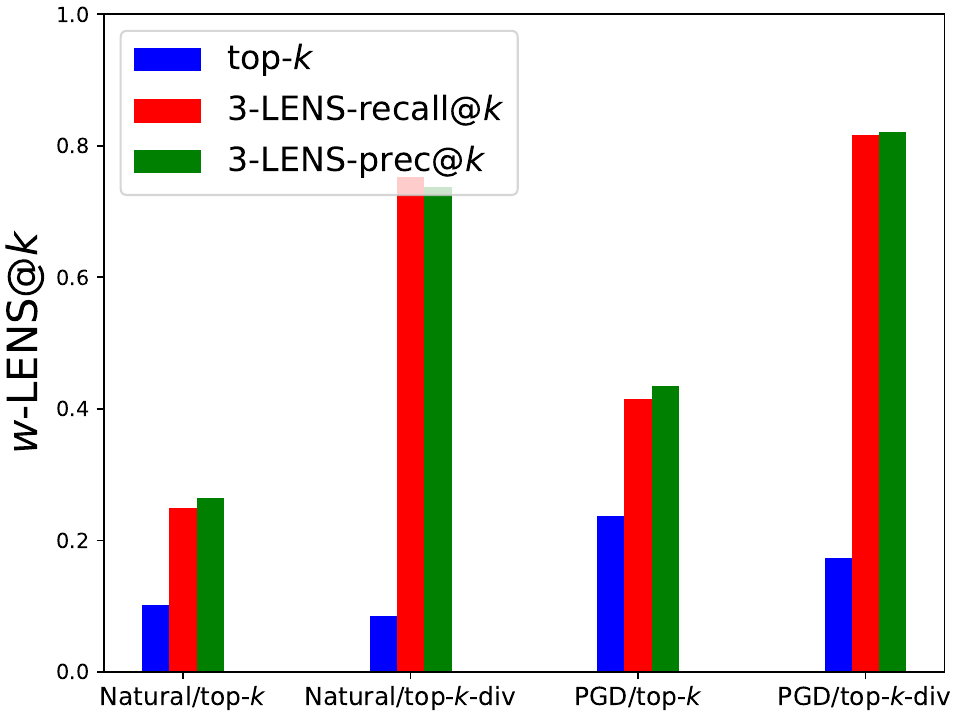}
    \caption{ImageNet}
    \label{flower-rand-fig:d}
    \end{subfigure}
\end{subfigure}
\caption{Attributional robustness of LRP \citep{bach2015lrp} on naturally and PGD trained models measured as top-$k$ intersection, $w$-LENS-prec@$k$ and $w$-LENS-recall@$k$ between the {\bf LRP} of the original images and of their perturbations obtained by the random sign perturbation across different datasets.} 
\label{all-datasets-topk-div-both-lens-lrp}
\end{figure}

\begin{figure}[!ht]
\centering
\begin{subfigure}[b]{1.0\textwidth}
    \begin{subfigure}[b]{0.15\textwidth}
    \includegraphics[width=1.0\textwidth]{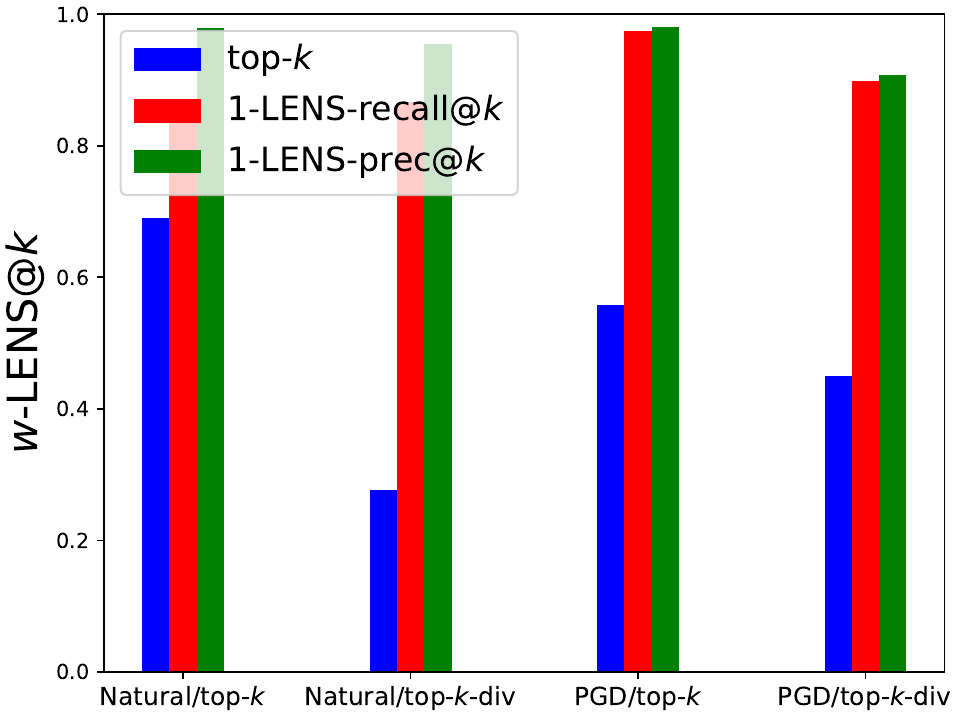}
    \caption{MNIST}
    \label{mnist-rand-fig:a}
    \end{subfigure}
    \begin{subfigure}[b]{0.15\textwidth}
    \includegraphics[width=1.0\textwidth]{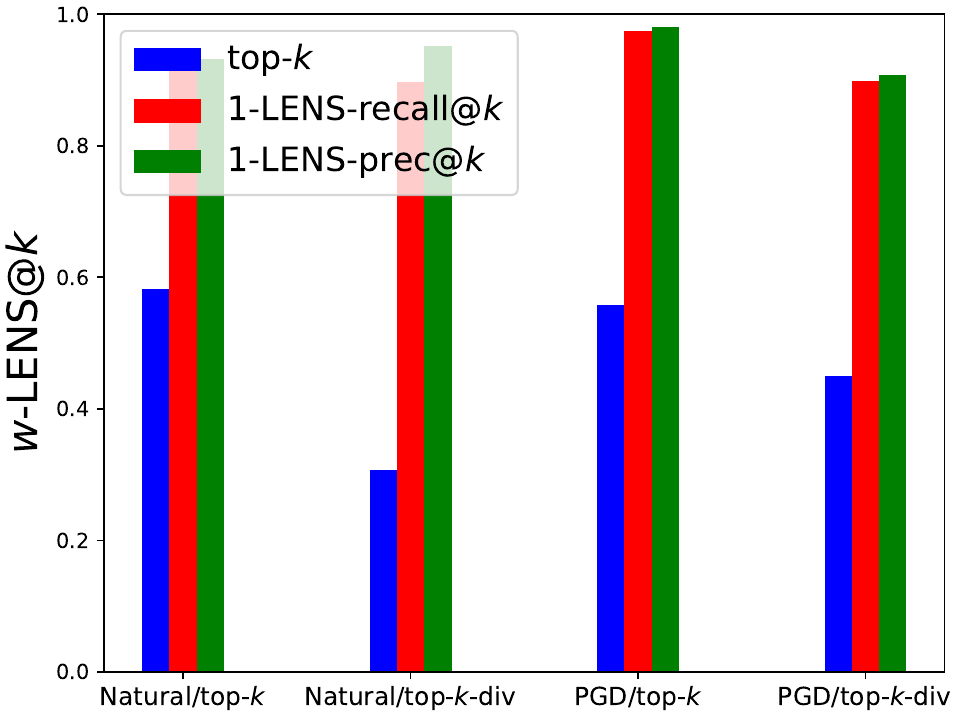}
    \caption{Fashion MNIST}
    \label{fmnist-rand-fig:b}
    \end{subfigure}
    \begin{subfigure}[b]{0.15\textwidth}
    \includegraphics[width=1.0\textwidth]{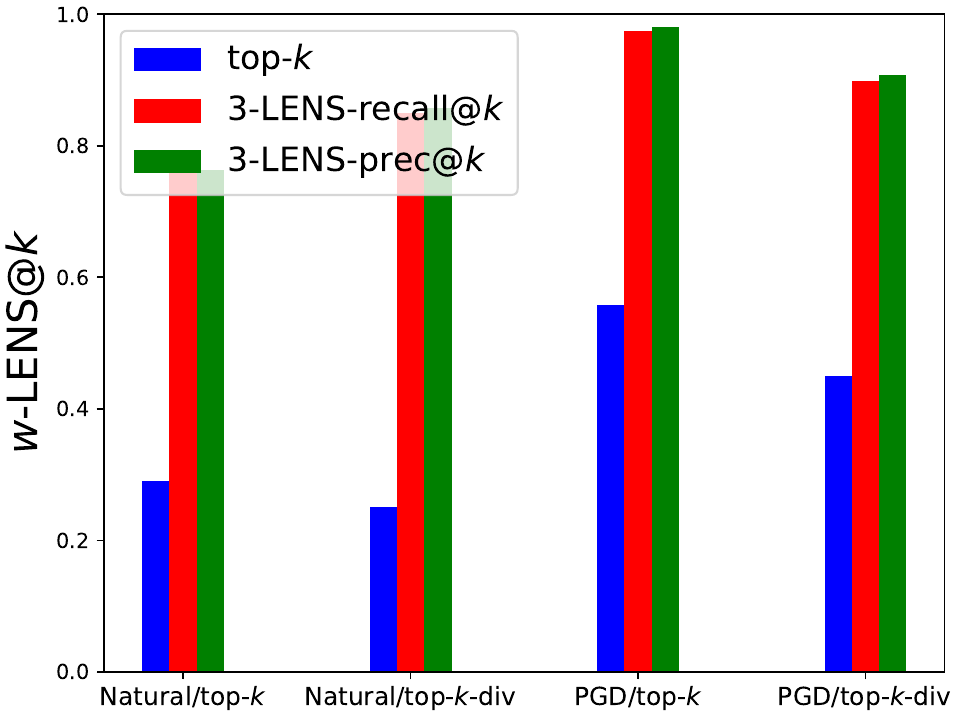}
    \caption{ImageNet}
    \label{flower-rand-fig:d}
    \end{subfigure}
\end{subfigure}
\caption{Attributional robustness of DeepLIFT \citep{shrikumar2019learning} on naturally and PGD trained models measured as top-$w$ intersection, $w$-LENS-prec@$k$ and $w$-LENS-recall@$k$ between the {\bf DeepLIFT} of the original images and of their perturbations obtained by the random sign perturbation across different datasets.} 
\label{all-datasets-topk-div-both-lens-dl}
\end{figure}

\begin{figure}[!ht]
\centering
\begin{subfigure}[b]{1.0\textwidth}
    \begin{subfigure}[b]{0.15\textwidth}
    \includegraphics[width=1.0\textwidth]{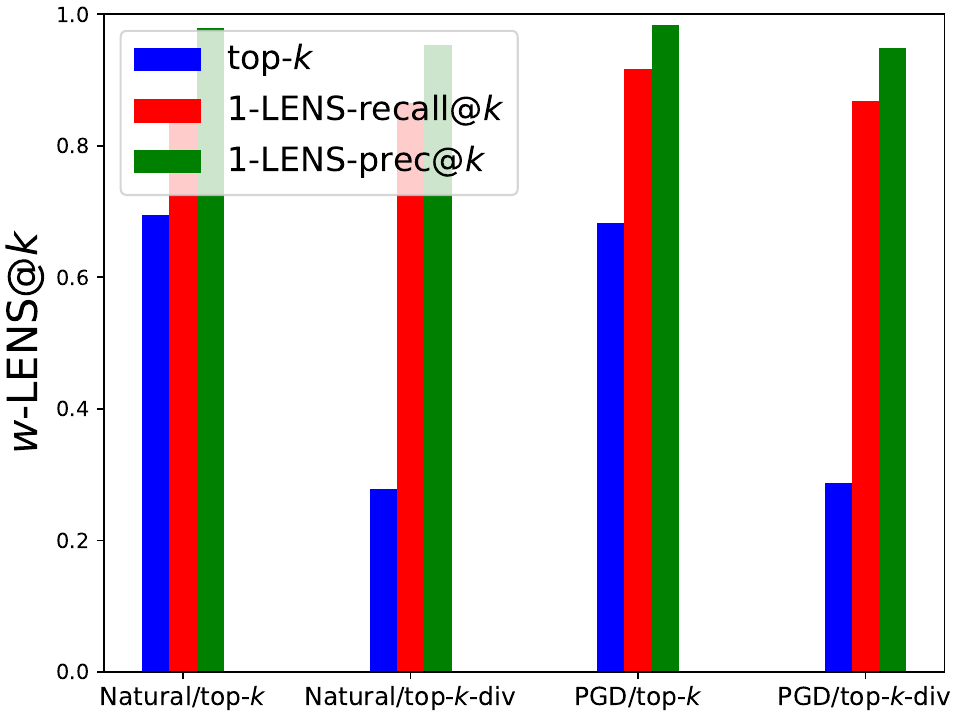}
    \caption{MNIST}
    \label{mnist-rand-fig:a}
    \end{subfigure}
    \begin{subfigure}[b]{0.15\textwidth}
    \includegraphics[width=1.0\textwidth]{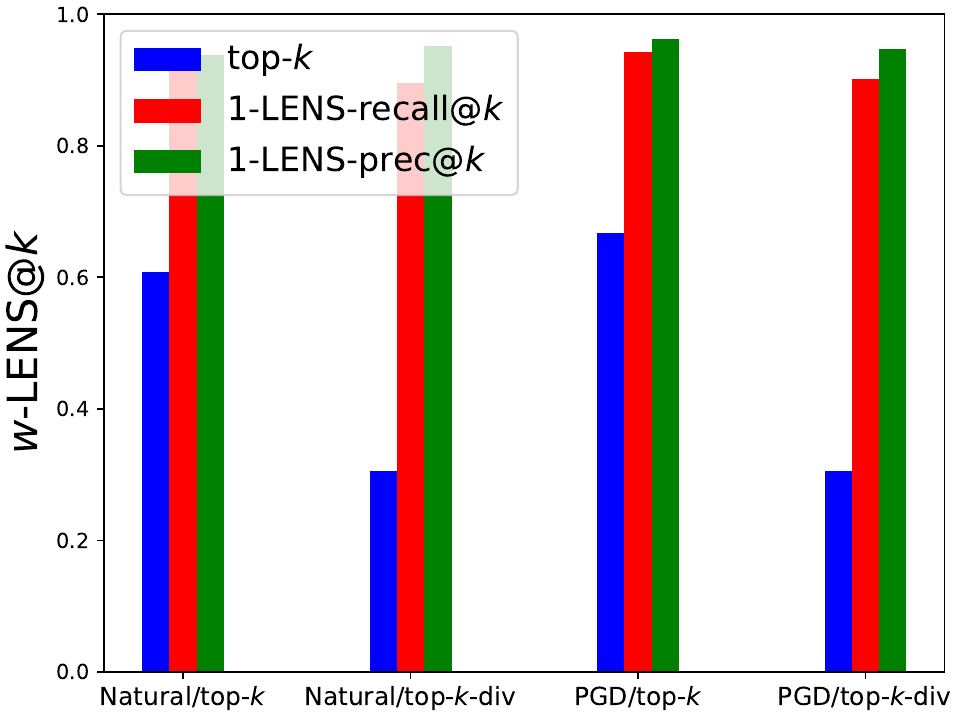}
    \caption{Fashion MNIST}
    \label{fmnist-rand-fig:b}
    \end{subfigure}
    \begin{subfigure}[b]{0.15\textwidth}
    \includegraphics[width=1.0\textwidth]{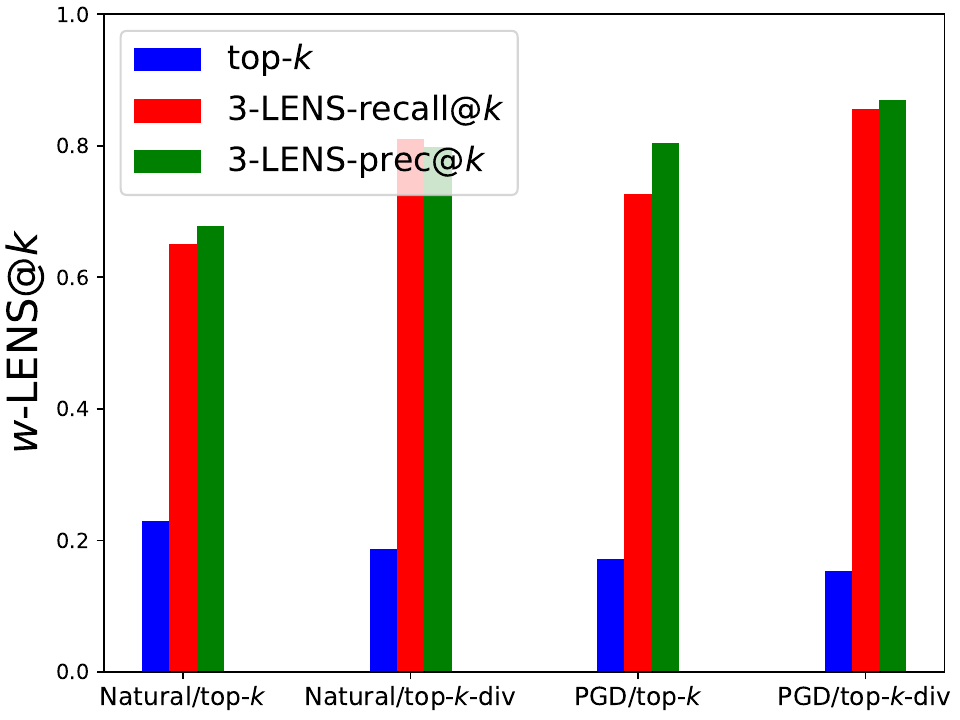}
    \caption{ImageNet}
    \label{flower-rand-fig:d}
    \end{subfigure}
\end{subfigure}
\caption{Attributional robustness of GradSHAP \citep{lundberg2017unified} on naturally and PGD trained models measured as top-$k$ intersection, $w$-LENS-prec@$k$ and $w$-LENS-recall@$k$ between the {\bf GradSHAP} of the original images and of their perturbations obtained by the random sign perturbation across different datasets.} 
\label{all-datasets-topk-div-both-lens-gs}
\end{figure}

\begin{figure}[!ht]
\centering
\begin{subfigure}[b]{1.0\textwidth}
    \begin{subfigure}[b]{0.15\textwidth}
    \includegraphics[width=1.0\textwidth]{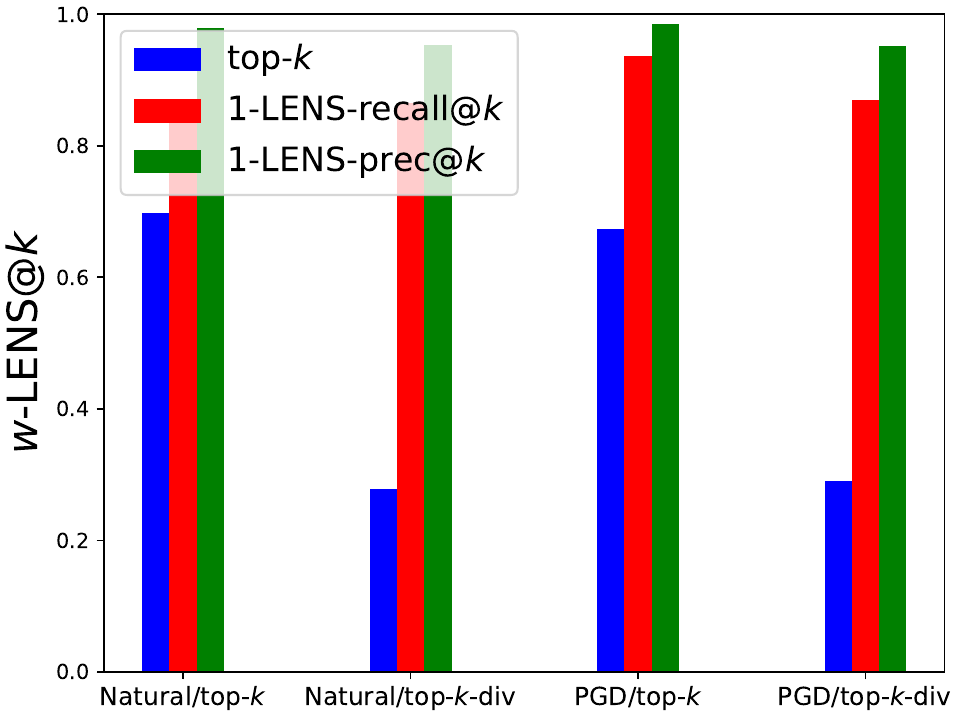}
    \caption{MNIST}
    \label{mnist-rand-fig:a}
    \end{subfigure}
    \begin{subfigure}[b]{0.15\textwidth}
    \includegraphics[width=1.0\textwidth]{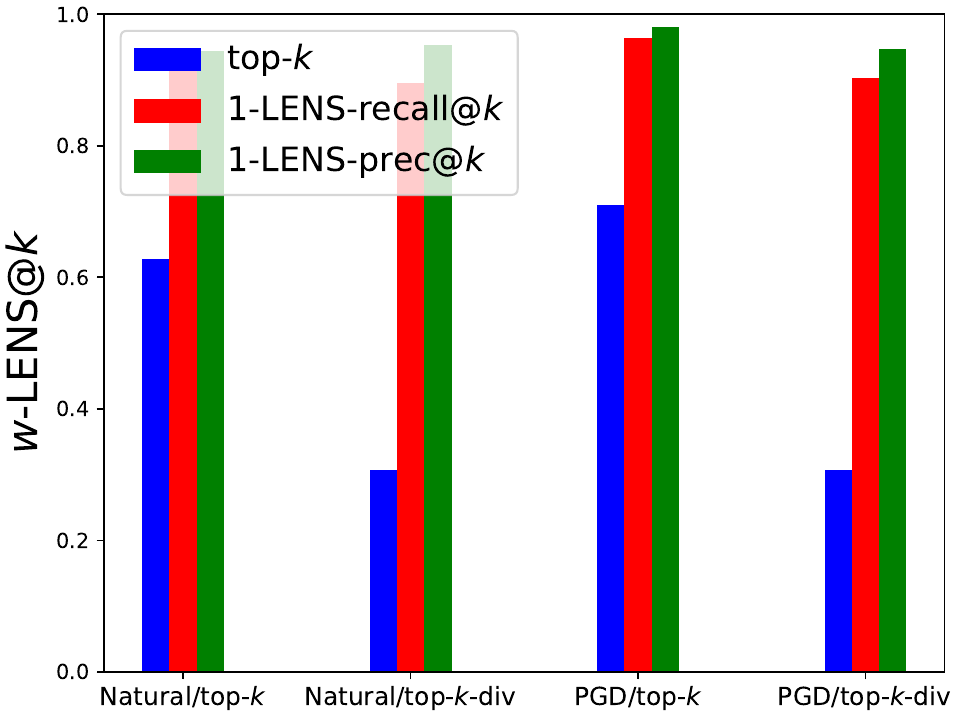}
    \caption{Fashion MNIST}
    \label{fmnist-rand-fig:b}
    \end{subfigure}
    \begin{subfigure}[b]{0.15\textwidth}
    \includegraphics[width=1.0\textwidth]{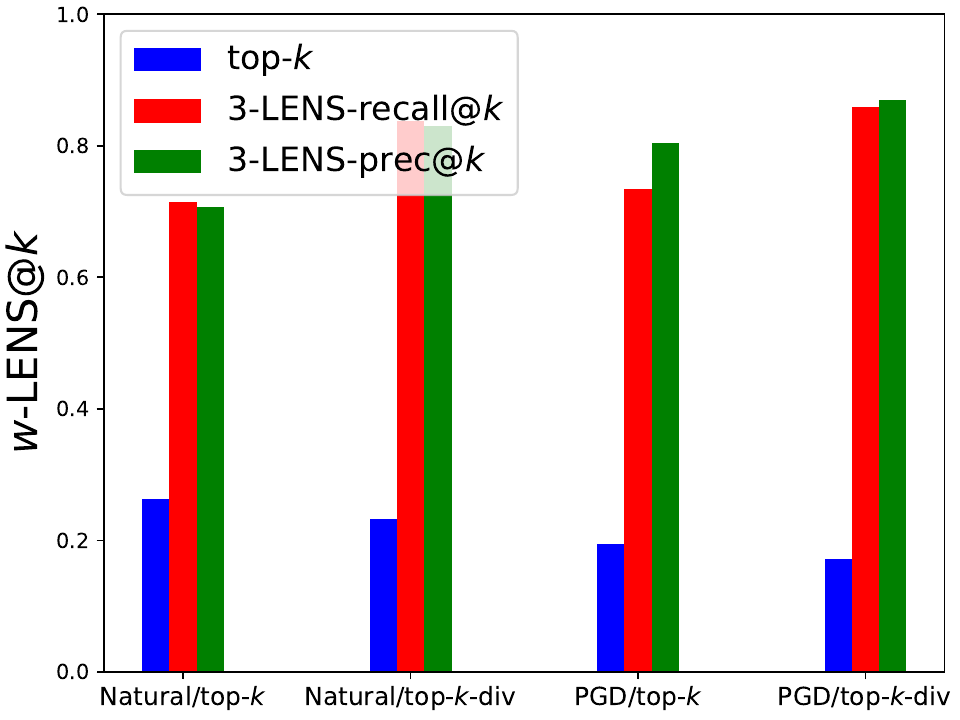}
    \caption{ImageNet}
    \label{flower-rand-fig:d}
    \end{subfigure}
\end{subfigure}
\caption{Attributional robustness of Integrated Gradients \citep{SundararajanTY17} on naturally and PGD trained models measured as top-$k$ intersection, $w$-LENS-prec@$k$ and $w$-LENS-recall@$k$ between the {\bf Integrated Gradients(IG)} of the original images and of their perturbations obtained by the random sign perturbation across different datasets.} 
\label{all-datasets-topk-div-both-lens-ig}
\end{figure}

\begin{figure}[!ht]
\centering
\begin{subfigure}[b]{1.0\textwidth}
    \begin{subfigure}[b]{0.24\textwidth}
    \includegraphics[width=1.0\textwidth]{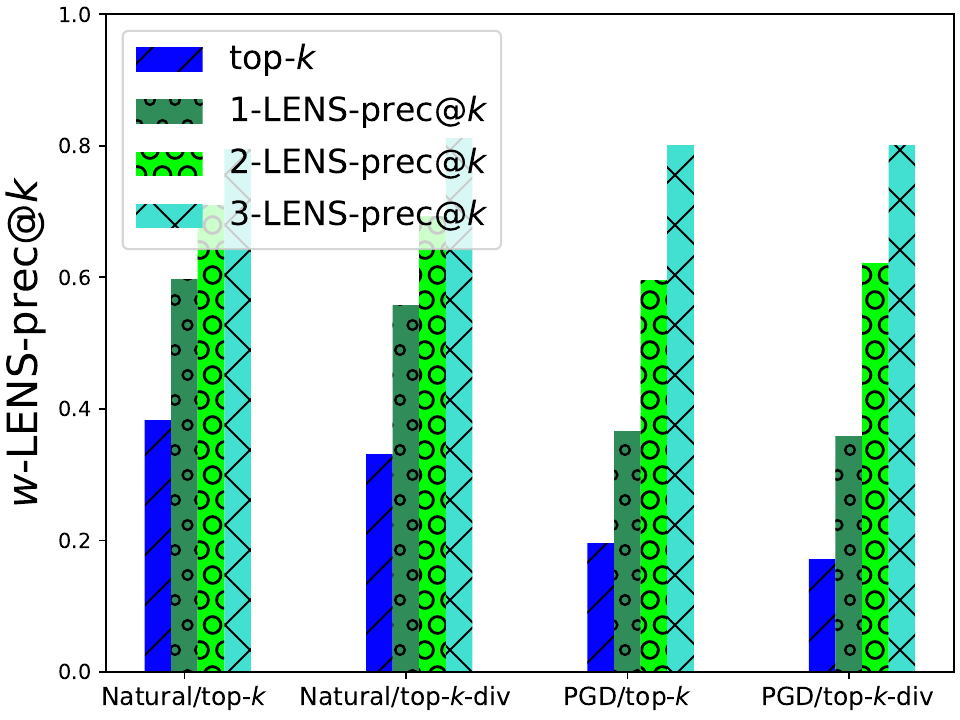}
    \caption{Simple Gradient}
    \label{mnist-rand-fig:a}
    \end{subfigure}
    \begin{subfigure}[b]{0.24\textwidth}
    \includegraphics[width=1.0\textwidth]{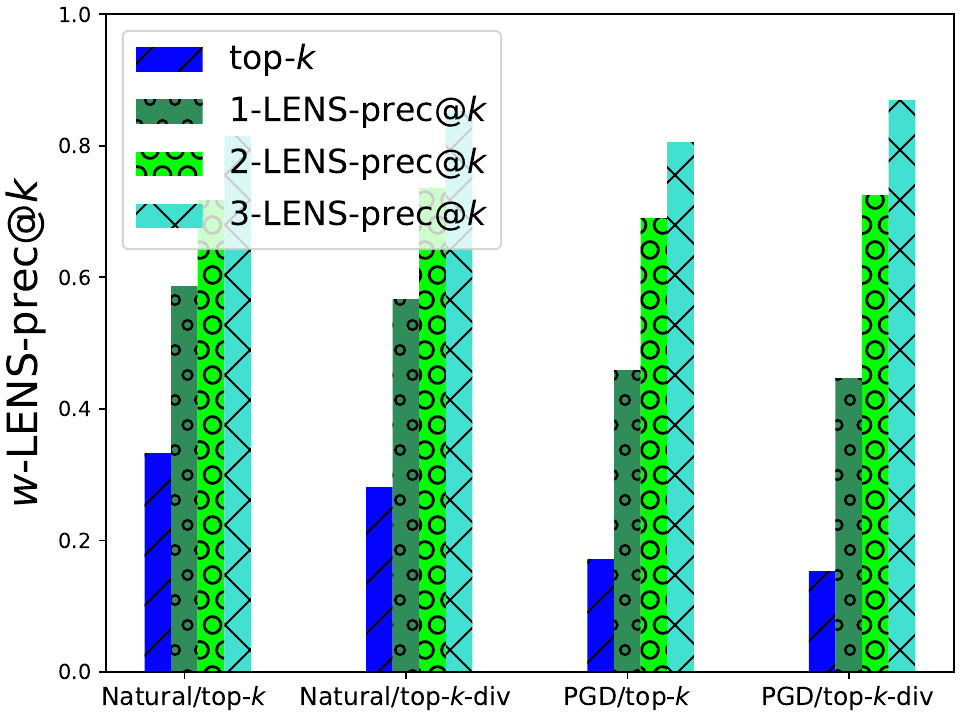}
    \caption{Image $\times$ Gradient}
    \label{mnist-rand-fig:b}
    \end{subfigure}\\
    \begin{subfigure}[b]{0.24\textwidth}
    \includegraphics[width=1.0\textwidth]{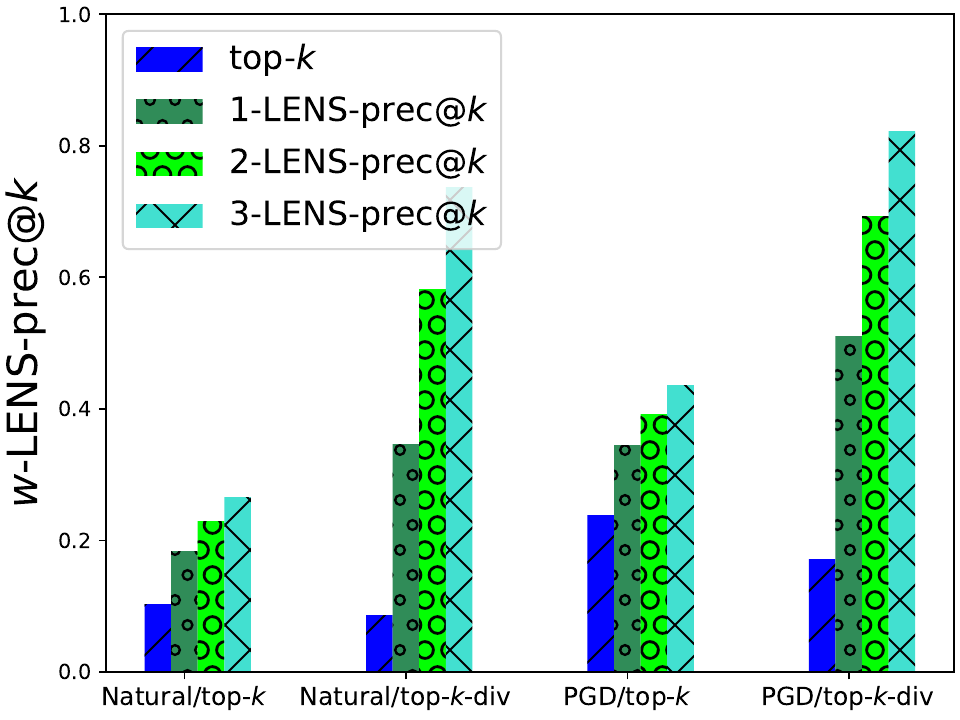}
    \caption{LRP [Bach 2015]}
    \label{mnist-rand-fig:c}
    \end{subfigure}
    \begin{subfigure}[b]{0.24\textwidth}
    \includegraphics[width=1.0\textwidth]{./plots/imagenet_nat_pgd_rand_sg_lens_div.pdf}
    \caption{DeepLIFT [Shrikumar 2017]}
    \label{mnist-rand-fig:a}
    \end{subfigure}\\    
    \begin{subfigure}[b]{0.24\textwidth}
    \includegraphics[width=1.0\textwidth]{./plots/imagenet_nat_pgd_rand_ixg_lens_div.pdf}
    \caption{GradSHAP [Lundberg 2017]}
    \label{mnist-rand-fig:b}
    \end{subfigure}
    \begin{subfigure}[b]{0.24\textwidth}
    \includegraphics[width=1.0\textwidth]{./plots/imagenet_nat_pgd_rand_lrp_lens_div.pdf}
    \caption{IG [Sundararajan 2017]}
    \label{mnist-rand-fig:c}
    \end{subfigure}
\end{subfigure}
\caption{Attributional robustness of explanation methods on naturally and PGD trained models measured as top-$k$ intersection and $w$-LENS-prec@$k$ between the explanation map of the original images and their perturbations. Perturbations are obtained by the random sign perturbation. The plots show the effect of varying $w$ on ImageNet dataset with naturally and PGD trained ResNet50 model.}
\label{imagenet-nat-pgd-w-lens-div}     
\end{figure}

\clearpage

\section{The fragility of top-$k$ intersection} \label{app:sample-fragility-examples}

\begin{figure*}[!ht]
\centering
\begin{subfigure}[b]{1.0\textwidth}
    \begin{subfigure}[b]{0.22\textwidth}
    \includegraphics[width=1.0\textwidth]{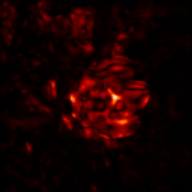}
    \caption{Original IG}
    \label{mnist-rand-fig:a}
    \end{subfigure}
    \begin{subfigure}[b]{0.22\textwidth}
    \includegraphics[width=1.0\textwidth]{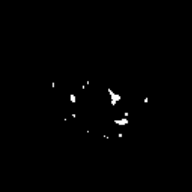}
    \caption{top-100 pixels}
    \label{fmnist-rand-fig:b}
    \end{subfigure}
    \begin{subfigure}[b]{0.22\textwidth}
    \includegraphics[width=1.0\textwidth]{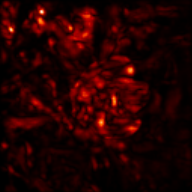}
    \caption{Perturbed IG}
    \label{gtsrb-rand-fig:c}
    \end{subfigure}
    \begin{subfigure}[b]{0.22\textwidth}
    \includegraphics[width=1.0\textwidth]{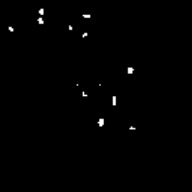}
    \caption{top-100 pixels}
    \label{flower-rand-fig:d}
    \end{subfigure}
\end{subfigure}
\begin{subfigure}[b]{1.0\textwidth}
    \begin{subfigure}[b]{0.22\textwidth}
    \includegraphics[width=1.0\textwidth]{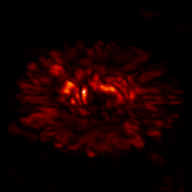}
    \caption{Original IG}
    \label{mnist-rand-fig:a}
    \end{subfigure}
    \begin{subfigure}[b]{0.22\textwidth}
    \includegraphics[width=1.0\textwidth]{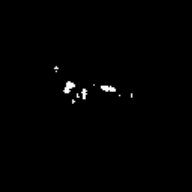}
    \caption{top-100 pixels}
    \label{fmnist-rand-fig:b}
    \end{subfigure}
    \begin{subfigure}[b]{0.22\textwidth}
    \includegraphics[width=1.0\textwidth]{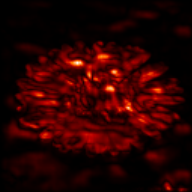}
    \caption{Perturbed IG}
    \label{gtsrb-rand-fig:c}
    \end{subfigure}
    \begin{subfigure}[b]{0.22\textwidth}
    \includegraphics[width=1.0\textwidth]{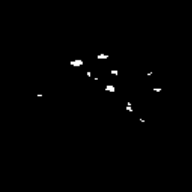}
    \caption{top-100 pixels}
    \label{flower-rand-fig:d}
    \end{subfigure}    
\end{subfigure}
\caption{Sample Integrated Gradients(IG) map using Flower dataset where top-100 is highlighted before and after image is perturbed with top-$k$ attack of \citet{ghorbani2018interpretation}.}
\label{app:flower-topk-highlight}     
\end{figure*}

\begin{figure*}[!ht]
\centering
\begin{subfigure}[b]{1.0\textwidth}
    \begin{subfigure}[b]{0.22\textwidth}
    \includegraphics[width=1.0\textwidth]{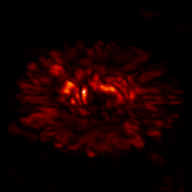}
    \caption{Original IG}
    \label{mnist-rand-fig:a}
    \end{subfigure}
    \begin{subfigure}[b]{0.22\textwidth}
    \includegraphics[width=1.0\textwidth]{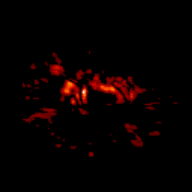}
    \caption{top-1000 pixels}
    \label{fmnist-rand-fig:b}
    \end{subfigure}
    \begin{subfigure}[b]{0.22\textwidth}
    \includegraphics[width=1.0\textwidth]{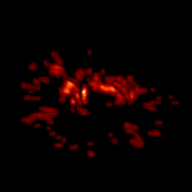}
    \caption{1-LENS-recall@$k$}
    \label{gtsrb-rand-fig:c}
    \end{subfigure}
    \begin{subfigure}[b]{0.22\textwidth}
    \includegraphics[width=1.0\textwidth]{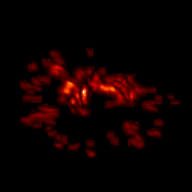}
    \caption{2-LENS-recall@$k$}
    \label{flower-rand-fig:d}
    \end{subfigure}
\end{subfigure}
\begin{subfigure}[b]{1.0\textwidth}
    \begin{subfigure}[b]{0.22\textwidth}
    \includegraphics[width=1.0\textwidth]{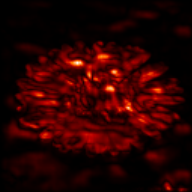}
    \caption{Perturbed IG}
    \label{mnist-rand-fig:a}
    \end{subfigure}
    \begin{subfigure}[b]{0.22\textwidth}
    \includegraphics[width=1.0\textwidth]{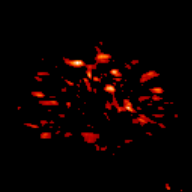}
    \caption{top-1000 pixels}
    \label{fmnist-rand-fig:b}
    \end{subfigure}
    \begin{subfigure}[b]{0.22\textwidth}
    \includegraphics[width=1.0\textwidth]{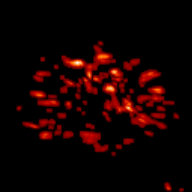}
    \caption{1-LENS-prec@$k$}
    \label{gtsrb-rand-fig:c}
    \end{subfigure}
    \begin{subfigure}[b]{0.22\textwidth}
    \includegraphics[width=1.0\textwidth]{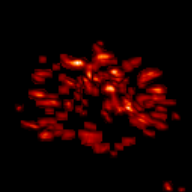}
    \caption{2-LENS-prec@$k$}
    \label{flower-rand-fig:d}
    \end{subfigure}
\end{subfigure}
\caption{Example based on locality. Sample Integrated Gradients(IG) map using Flower dataset with top-$k$ highlighted followed by $w$-LENS@$k$ maps (row 1) Original IG and (row 2) Perturbed IG with top-$k$ attack \citep{ghorbani2018interpretation}.}
\label{app:flower-topk-highlight-lens-19}     
\end{figure*}

\begin{figure*}[!ht]
\centering
\begin{subfigure}[b]{1.0\textwidth}
    \begin{subfigure}[b]{0.22\textwidth}
    \includegraphics[width=1.0\textwidth]{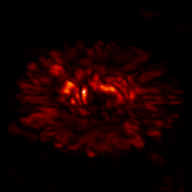}
    \caption{Original IG}
    \label{mnist-rand-fig:a}
    \end{subfigure}
    \begin{subfigure}[b]{0.22\textwidth}
    \includegraphics[width=1.0\textwidth]{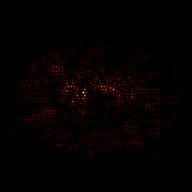}
    \caption{top-1000-div pixels}
    \label{fmnist-rand-fig:b}
    \end{subfigure}
    \begin{subfigure}[b]{0.22\textwidth}
    \includegraphics[width=1.0\textwidth]{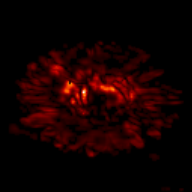}
    \caption{1-LENS-recall@$k$-div}
    \label{gtsrb-rand-fig:c}
    \end{subfigure}
    \begin{subfigure}[b]{0.22\textwidth}
    \includegraphics[width=1.0\textwidth]{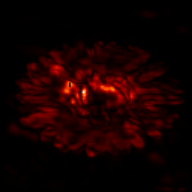}
    \caption{2-LENS-recall@$k$-div}
    \label{flower-rand-fig:d}
    \end{subfigure}
\end{subfigure}
\begin{subfigure}[b]{1.0\textwidth}
    \begin{subfigure}[b]{0.22\textwidth}
    \includegraphics[width=1.0\textwidth]{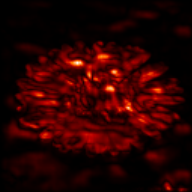}
    \caption{Perturbed IG}
    \label{mnist-rand-fig:a}
    \end{subfigure}
    \begin{subfigure}[b]{0.22\textwidth}
    \includegraphics[width=1.0\textwidth]{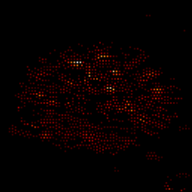}
    \caption{top-1000-div pixels}
    \label{fmnist-rand-fig:b}
    \end{subfigure}
    \begin{subfigure}[b]{0.22\textwidth}
    \includegraphics[width=1.0\textwidth]{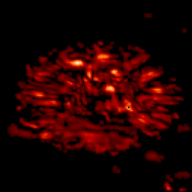}
    \caption{1-LENS-prec@$k$-div}
    \label{gtsrb-rand-fig:c}
    \end{subfigure}
    \begin{subfigure}[b]{0.22\textwidth}
    \includegraphics[width=1.0\textwidth]{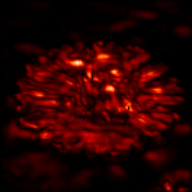}
    \caption{2-LENS-prec@$k$-div}
    \label{flower-rand-fig:d}
    \end{subfigure}
\end{subfigure}
\caption{Example based on diversity. Sample Integrated Gradients(IG) map using Flower dataset with top-$k$-div highlighted followed by $w$-LENS@$k$-div maps (row 1) Original IG and (row 2) Perturbed IG with top-$k$ attack \citep{ghorbani2018interpretation}.}
\label{app:flower-topk-highlight-div-19}     
\end{figure*}

\begin{figure*}[!ht]
\centering
\begin{subfigure}[b]{1.0\textwidth}
    \begin{subfigure}[b]{0.33\textwidth}
    \includegraphics[width=1.0\textwidth]{./images/flower_19_topk_nat_orig.pdf}
    \caption{Original IG}
    \label{mnist-rand-fig:a}
    \end{subfigure}
    \begin{subfigure}[b]{0.33\textwidth}
    \includegraphics[width=1.0\textwidth]{./images/flower_19_topk_nat_orig_orig_div.pdf}
    \caption{top-1000 diverse pixels with $3{\times}3$ window}
    \label{fmnist-rand-fig:b}
    \end{subfigure}
    \begin{subfigure}[b]{0.33\textwidth}
    \includegraphics[width=1.0\textwidth]{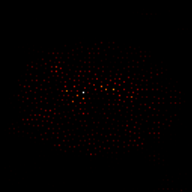}
    \caption{top-1000 diverse pixels with $5{\times}5$ window}
    \label{gtsrb-rand-fig:c}
    \end{subfigure}
\end{subfigure}
\begin{subfigure}[b]{1.0\textwidth}
    \begin{subfigure}[b]{0.33\textwidth}
    \includegraphics[width=1.0\textwidth]{./images/flower_19_topk_nat_pert.pdf}
    \caption{Perturbed IG}
    \label{mnist-rand-fig:a}
    \end{subfigure}
    \begin{subfigure}[b]{0.33\textwidth}
    \includegraphics[width=1.0\textwidth]{./images/flower_19_topk_nat_orig_pert_div.pdf}
    \caption{top-1000 diverse pixels with $3{\times}3$ window}
    \label{fmnist-rand-fig:b}
    \end{subfigure}
    \begin{subfigure}[b]{0.33\textwidth}
    \includegraphics[width=1.0\textwidth]{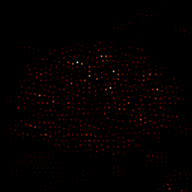}
    \caption{top-1000 diverse pixels with $5{\times}5$ window}
    \label{gtsrb-rand-fig:c}
    \end{subfigure}
\end{subfigure}
\caption{Example based on diversity with different window sizes. Sample Integrated Gradients(IG) map using Flower dataset with top-$k$-div highlighted with (column 2) 3$\times$3 window (column 3) 5$\times$5 window. (column 1) (top) map of unperturbed image (bottom) map of perturbed image with top-$k$ attack \citep{ghorbani2018interpretation}.}
\label{app:flower-topk-highlight-div-win-19}     
\end{figure*}

\begin{figure*}[!ht]
\centering
\begin{subfigure}[b]{1.0\textwidth}
    \begin{subfigure}[b]{0.24\textwidth}
    \includegraphics[width=1.0\textwidth]{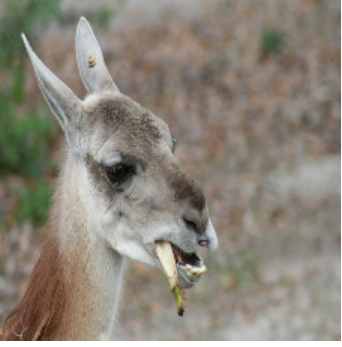}
    \caption{Original Image}
    \label{mnist-rand-fig:a}
    \end{subfigure}
    \begin{subfigure}[b]{0.24\textwidth}
    \includegraphics[width=1.0\textwidth]{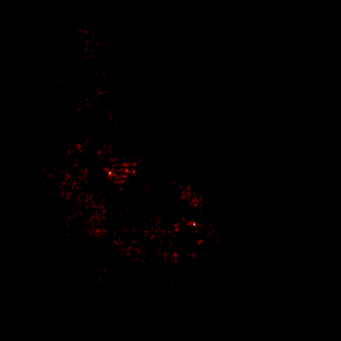}
    \caption{top-1000 of Original IG}
    \label{mnist-rand-fig:a}
    \end{subfigure}  
    \begin{subfigure}[b]{0.24\textwidth}
    \includegraphics[width=1.0\textwidth]{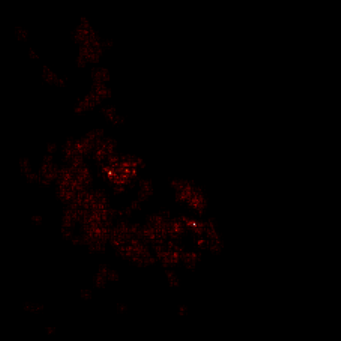}
    \caption{3-LENS-recall@1000}
    \label{fmnist-rand-fig:b}
    \end{subfigure}
    \begin{subfigure}[b]{0.24\textwidth}
    \includegraphics[width=1.0\textwidth]{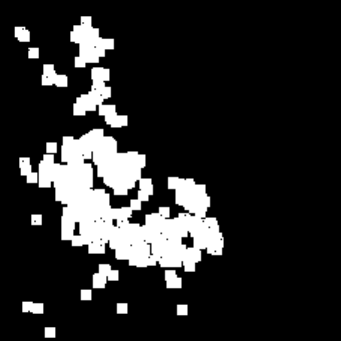}
    \caption{3-LENS-recall@1000 0/1}
    \label{fmnist-rand-fig:b}
    \end{subfigure}
\end{subfigure}
\begin{subfigure}[b]{1.0\textwidth}
    \begin{subfigure}[b]{0.24\textwidth}
    \includegraphics[width=1.0\textwidth]{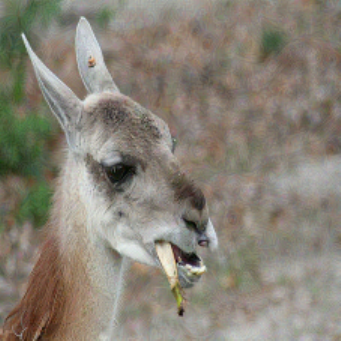}
    \caption{Perturbed Image}
    \label{mnist-rand-fig:a}
    \end{subfigure}
    \begin{subfigure}[b]{0.24\textwidth}
    \includegraphics[width=1.0\textwidth]{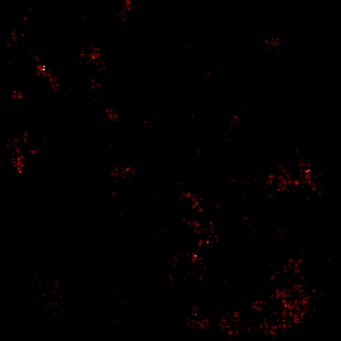}
    \caption{top-1000 of Perturbed IG}
    \label{fmnist-rand-fig:b}
    \end{subfigure}    
    \begin{subfigure}[b]{0.24\textwidth}
    \includegraphics[width=1.0\textwidth]{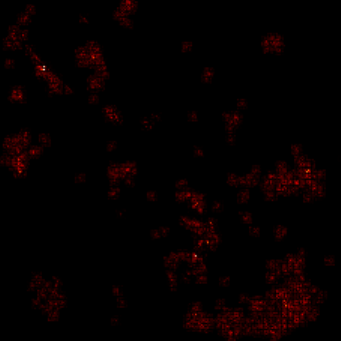}
    \caption{3-LENS-prec@1000}
    \label{gtsrb-rand-fig:c}
    \end{subfigure}    
        \begin{subfigure}[b]{0.24\textwidth}
    \includegraphics[width=1.0\textwidth]{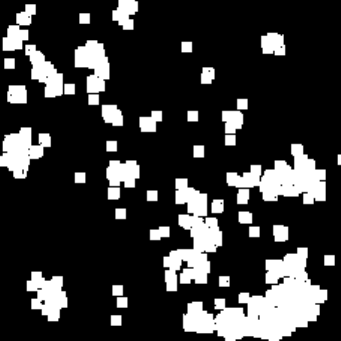}
    \caption{3-LENS-prec@1000 0/1}
    \label{gtsrb-rand-fig:c}
    \end{subfigure}
\end{subfigure}    
\caption{Example based on locality(LENS) with sample Integrated Gradients(IG) map from ImageNet dataset (a) the original image and (b) being the perturbed image. (c) and (d) show top-$k$ highlighted and (e) and (f) are the corresponding maps with LENS. Maps (g), (h) are the maps corresponding to (e), (f), respectively with non-zero value pixels shown as white. top-k:0.108, 3-LENS-recall@k:0.254, 3-LENS-prec@k:0.433, top-k-div:0.090, 3-LENS-recall@k-div:0.758, 3-LENS-prec@k-div:0.807}
\label{app:imagenet-topk-highlight-loc-19}     
\end{figure*}

\begin{figure*}[!ht]
\centering
\begin{subfigure}[b]{1.0\textwidth}
    \begin{subfigure}[b]{0.24\textwidth}
    \includegraphics[width=1.0\textwidth]{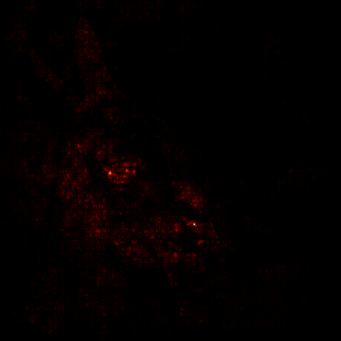}
    \caption{Original IG}
    \label{mnist-rand-fig:a}
    \end{subfigure}
    \begin{subfigure}[b]{0.24\textwidth}
    \includegraphics[width=1.0\textwidth]{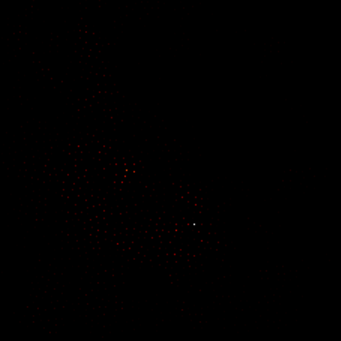}
    \caption{top-1000-div of Original IG}
    \label{mnist-rand-fig:a}
    \end{subfigure}
    \begin{subfigure}[b]{0.24\textwidth}
    \includegraphics[width=1.0\textwidth]{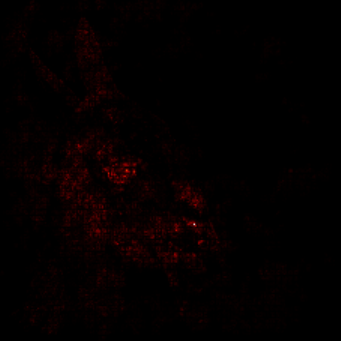}
    \caption{3-LENS-recall@1000-div}
    \label{fmnist-rand-fig:b}
    \end{subfigure}
    \begin{subfigure}[b]{0.24\textwidth}
    \includegraphics[width=1.0\textwidth]{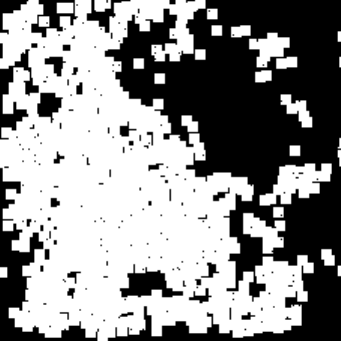}
    \caption{3-LENS-recall@1000-div 0/1}
    \label{fmnist-rand-fig:b}
    \end{subfigure}
\end{subfigure}
\begin{subfigure}[b]{1.0\textwidth}
    \begin{subfigure}[b]{0.24\textwidth}
    \includegraphics[width=1.0\textwidth]{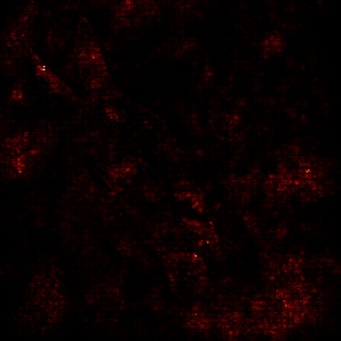}
    \caption{Perturbed IG}
    \label{mnist-rand-fig:a}
    \end{subfigure}    
    \begin{subfigure}[b]{0.24\textwidth}
    \includegraphics[width=1.0\textwidth]{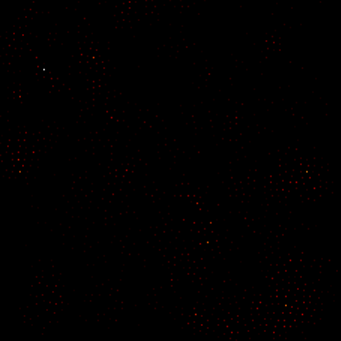}
    \caption{top-1000-div of Perturbed IG}
    \label{fmnist-rand-fig:b}
    \end{subfigure}
    \begin{subfigure}[b]{0.24\textwidth}
    \includegraphics[width=1.0\textwidth]{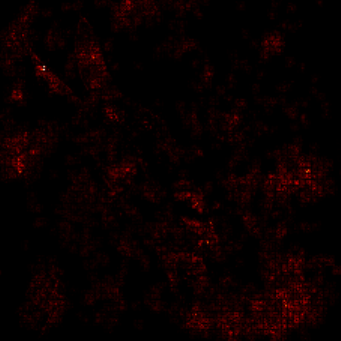}
    \caption{3-LENS-prec@1000-div}
    \label{gtsrb-rand-fig:c}
    \end{subfigure}    
    \begin{subfigure}[b]{0.24\textwidth}
    \includegraphics[width=1.0\textwidth]{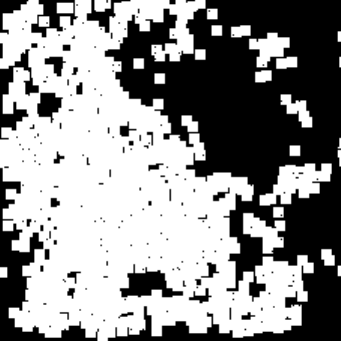}
    \caption{3-LENS-prec@1000-div 0/1}
    \label{gtsrb-rand-fig:c}
    \end{subfigure}
\end{subfigure}
\caption{Example based on diversity with 7$\times$7 window size with sample Integrated Gradients(IG) map from ImageNet dataset (a) the original IG of unperturbed image and (b) perturbed IG obtained with top-$k$ attack\citep{ghorbani2018interpretation}. (c) and (d) show top-$k$-div highlighted and (e) and (f) are the corresponding maps with LENS. Maps (g), (h) are the maps corresponding to (e), (f), respectively with non-zero value pixels shown as white. top-k:0.108, 3-LENS-recall@k:0.254, 3-LENS-prec@k:0.433, top-k-div:0.090, 3-LENS-recall@k-div:0.758, 3-LENS-prec@k-div:0.807}
\label{app:imagenet-topk-highlight-div-win-19}     
\end{figure*}

Figure \ref{app:flower-topk-highlight} highlights the top-100 pixels in the unperturbed and perturbed maps of a sample image. Figure \ref{app:flower-topk-highlight-lens-19}, \ref{app:flower-topk-highlight-div-19} show the top-1000 pixels and top-1000 diverse pixels along with $w$-LENS-recall@$k$ and $w$-LENS-prec@$k$ on them. Figure \ref{app:flower-topk-highlight-div-win-19} visualize the top-1000 diverse pixels obtained with different window sizes eg. 3 and 5, respectively. Figure \ref{app:imagenet-topk-highlight-loc-19} elaborates the use of LENS(locality) and Figure \ref{app:imagenet-topk-highlight-div-win-19} shows LENS-div(diversity) for the same example from ImageNet. Zoom in required to observe the finer details.

\clearpage

\section{Additional results for PGD-trained and IG-SUM-NORM trained models} \label{app:pgd-rar-results}

Figure \ref{dataset-topk-rand-diff-k-new-metric-pgd} and Figure \ref{dataset-topk-rand-diff-k-new-metric-rar} shows the impact of $k$ in top-$k$ for adversarially(PGD) trained and attributional(IG-SUM-NORM) trained network, respectively. But an important point to be noticed is that even with small number of features LENS is able to cross 70-80\% which supports the observation of sparsity and stability of attributions achieved by adversarially(PGD) trained models by \citet{ChalasaniC00J20}. Similarly, the experiments with different $w$ value for $w$-LENS-top-$k$ in Figure \ref{flower-topk-new-metric-training-method} clearly incidates that due to the stability properties at lower window sizes LENS is able to cross the 80\% intersection quickly. Supporting that our metric nicely captures local stability very well. 

While above we observed only the top-$k$ version of LENS. Figure \ref{dataset-smooth-sp-ken-nat-pgd-rar} further extends the observation to LENS-Spearman and LENS-Kendall who to show that with LENS with a smoothing of $3 \times 3$ the maps from PGD-trained and IG-SUM-NORM trained models have a higher top-$k$ intersection above 70\% in comparison to natural trained model across all datasets used in our experiments, which further strengthen the conclusions from previous papers that IG on PGD-trained and IG-SUM-NORM trained models give better attributions.

Appendix \ref{app:more-lens-results} and \ref{app:extra-div-results} provide results on PGD-trained and IG-SUM-NORM trained models along with naturally trained models for compact presentation of results.

\section{Additional results with top-$k$-div} \label{app:extra-div-results}
Table \ref{table-topk-div-results-mnist-fmnist-ig}  provides detailed LENS and diversity with LENS results on MNIST, Fashion-MNIST, Table \ref{table-topk-div-results-flower-ig} on Flower with natural, PGD and IG-SUM-NORM trained networks and Table \ref{table-topk-div-results-imagenet-ig} on naturally trained network for ImageNet. All results with top-$k$, center of mass attack proposed by \citet{ghorbani2018interpretation} and random sign perturbation using Integrated Gradients(IG).

Table \ref{table-topk-div-results-rand-mnist-fmnist} show the LENS and diversity results with random sign perturbation on MNIST, Fashion MNIST and Table \ref{table-topk-div-results-rand-imagenet-app} on ImageNet with different explanation methods like Simple Gradients(SG), Image$\times$Gradient, DeepLIFT\citep{shrikumar2017just,shrikumar2019learning}, LRP \citep{bach2015lrp}, GradShap\citep{lundberg2017unified} and Integrated Gradients(IG) \citep{SundararajanTY17}. 

\begin{table*}[!ht]
\begin{center}
\resizebox{1.0\textwidth}{!}{ 
\begin{tabular}{|l|l|c|c|c|c|c|c|c|c|}
\hline 
{\bf Dataset} & {\bf Train Type} & {\bf Attack Type} & {\bf Attribution Method} & {\bf top-$k$ intersection} & {\bf 1-LENS-recall@$k$} &  {\bf 1-LENS-prec@$k$} & {\bf top-$k$-div} & {\bf 1-LENS-recall@$k$-div} &  {\bf 1-LENS-prec@$k$-div} \\
\hline
MNIST & Nat & top-$k$ & IG & 0.4635 & 0.6209 & 0.9427 & 0.1680 & 0.6592 & 0.8126 \\ 
      & Nat & mass center   & IG & 0.5779 & 0.7146 & 0.9572 & 0.1708 & 0.6183 & 0.7191 \\ 
      & Nat & random & IG & 0.7384 & 0.8790 & 0.9808 & 0.2046 & 0.6774 & 0.8140 \\ 
\hline
MNIST & PGD & top-$k$ & IG & 0.5378 & 0.6127 & 0.9690 & 0.2300 & 0.6567 & 0.8114 \\ 
      & PGD & mass center   & IG & 0.6077 & 0.7133 & 0.9700 & 0.1907 & 0.5894 & 0.6801 \\ 
      & PGD & random & IG & 0.6867 & 0.8220 & 0.9835 & 0.1751 & 0.6726 & 0.7995 \\ 
\hline
MNIST & IG-SUM-NORM & top-$k$ & IG & 0.6406 & 0.7817 & 0.9736 & 0.2327 & 0.6701 & 0.7995 \\ 
      & IG-SUM-NORM & mass center   & IG & 0.7075 & 0.8387 & 0.9695 & 0.2002 & 0.6293 & 0.7294 \\ 
      & IG-SUM-NORM & random & IG & 0.7746 & 0.9129 & 0.9846 & 0.1783 & 0.6788 & 0.8142 \\ 
\hline
Fashion-MNIST & Nat & top-$k$ & IG & 0.3841 & 0.9118 & 0.9361 & 0.2132 & 0.8404 & 0.8857 \\ 
      & Nat & mass center   & IG & 0.5624 & 0.9333 & 0.9370 & 0.2961 & 0.8034 & 0.8277 \\ 
      & Nat & random & IG & 0.5378 & 0.9377 & 0.9500 & 0.2716 & 0.8337 & 0.8825 \\ 
\hline
Fashion-MNIST & PGD & top-$k$ & IG & 0.7440 & 0.9571 & 0.9696 & 0.3770 & 0.8486 & 0.8608 \\ 
      & PGD & mass center   & IG & 0.8543 & 0.9659 & 0.9721 & 0.4323 & 0.8249 & 0.8357 \\ 
      & PGD & random & IG & 0.8816 & 0.9863 & 0.9892 & 0.3806 & 0.8451 & 0.8695 \\ 
\hline
Fashion-MNIST & IG-SUM-NORM & top-$k$ & IG & 0.6953 & 0.9608 & 0.9697 & 0.3671 & 0.8437 & 0.8655 \\ 
      & IG-SUM-NORM & mass center   & IG & 0.8630 & 0.9666 & 0.9752 & 0.4485 & 0.8278 & 0.8437 \\ 
      & IG-SUM-NORM & random & IG & 0.8733 & 0.9837 & 0.9900 & 0.3459 & 0.8444 & 0.8766 \\ 
\hline
\end{tabular}
}
\end{center}
    \caption{Table with top-$k$ intersection and top-$k$-div results for MNIST and Fashion MNIST with LeNet based model using Integrated Gradients(IG) \citep{SundararajanTY17}. The columns first contain locality results: top-$k$ intersection, 1-LENS-recall@$k$, 1-LENS-prec@$k$ followed by diversity results: top-$k$-div, 1-LENS-recall@$k$-div, 1-LENS-prec@$k$-div. Models trained naturally, PGD and IG-SUM-NORM are used. Results include top-$k$, center of mass attack of \citet{ghorbani2018interpretation} as well as random sign perturbation.}
\label{table-topk-div-results-mnist-fmnist-ig}
\end{table*}

\begin{table*}[!ht]
\begin{center}
\resizebox{1.0\textwidth}{!}{ 
\begin{tabular}{|l|l|c|c|c|c|c|c|c|c|}
\hline 
{\bf Dataset} & {\bf Train Type} & {\bf Attack Type} & {\bf Attribution Method} & {\bf top-$k$ intersection} & {\bf 2-LENS-recall@$k$} &  {\bf 2-LENS-prec@$k$} & {\bf top-$k$-div} & {\bf 2-LENS-recall@$k$-div} &  {\bf 2-LENS-prec@$k$-div} \\
\hline
Flower & Nat & top-$k$ & IG & 0.1789 & 0.4091 & 0.5128 & 0.2355 & 0.9482 & 0.9560 \\ 
      & Nat & mass center   & IG & 0.4248 & 0.7196 & 0.6664 & 0.2645 & 0.9360 & 0.9420\\ 
      & Nat & random   & IG & 0.8206 & 0.9709 & 0.9747 & 0.4613 & 0.9741 & 0.9778\\ 
\hline
Flower & PGD & top-$k$ & IG & 0.5941 & 0.8444 & 0.9165 & 0.4078 & 0.9733 & 0.9737 \\ 
      & PGD & mass center   & IG & 0.6983 & 0.9223 & 0.9303 & 0.4247 & 0.9656 & 0.9655\\ 
      & PGD & random   & IG & 0.9033 & 0.9929 & 0.9934 & 0.5948 & 0.9847 & 0.9886\\ 
\hline
Flower & IG-SUM-NORM & top-$k$ & IG & 0.6179 & 0.8720 & 0.9299 & 0.3875 & 0.9747 & 0.9757 \\ 
      & IG-SUM-NORM & mass center   & IG & 0.7053 & 0.9303 & 0.9435 & 0.4031 & 0.9667 & 0.9672\\ 
      & IG-SUM-NORM & random   & IG & 0.8874 & 0.9917 & 0.9927 & 0.5707 & 0.9841 & 0.9869\\ 
\hline
\end{tabular}
}
\end{center}
    \caption{Table with top-$k$ intersection and top-$k$-div results for Flower with ResNet based model using Integrated Gradients(IG) \citep{SundararajanTY17}. The columns first contain locality results: top-$k$ intersection, 2-LENS-recall@$k$, 2-LENS-prec@$k$ followed by diversity results: top-$k$-div, 2-LENS-recall@$k$-div, 2-LENS-prec@$k$-div. Models trained naturally, PGD and IG-SUM-NORM are used. Results include top-$k$, center of mass attack of \citet{ghorbani2018interpretation} as well as random sign perturbation.}
\label{table-topk-div-results-flower-ig}
\end{table*}

\begin{table*}[!ht]
\begin{center}
\resizebox{1.0\textwidth}{!}{ 
\begin{tabular}{|l|l|c|c|c|c|c|c|c|c|}
\hline 
{\bf Dataset} & {\bf Train Type} & {\bf Attack Type} & {\bf Attribution Method} & {\bf top-$k$ intersection} & {\bf 3-LENS-recall@$k$} &  {\bf 3-LENS-prec@$k$} & {\bf top-$k$-div} & {\bf 3-LENS-recall@$k$-div} &  {\bf 3-LENS-prec@$k$-div} \\
\hline
ImageNet & Nat & top-$k$ & IG & 0.0544 & 0.2822 & 0.3189 & 0.0684 & 0.7303 & 0.7458 \\ 
      & Nat & mass center   & IG & 0.0869 & 0.3994 & 0.2082 & 0.0914 & 0.7221 & 0.7114\\ 
      & Nat & random & IG & 0.3133 & 0.8463 & 0.8460 & 0.2157 & 0.8713 & 0.8689 \\ 
\hline
\end{tabular}
}
\end{center}
    \caption{Table with top-$k$ intersection and top-$k$-div results for ImageNet with SqueezeNet model using Integrated Gradients(IG) \citep{SundararajanTY17}. The columns first contain locality results: top-$k$ intersection, 3-LENS-recall@$k$, 3-LENS-prec@$k$ followed by diversity results: top-$k$-div, 3-LENS-recall@$k$-div, 3-LENS-prec@$k$-div. Models naturally trained are used. Results include top-$k$, center of mass attack of \citet{ghorbani2018interpretation} as well as random sign perturbation.}
\label{table-topk-div-results-imagenet-ig}
\end{table*}

\begin{table*}[!ht]
\begin{center}
\resizebox{1.0\textwidth}{!}{ 
\begin{tabular}{|l|l|c|c|c|c|c|c|c|c|}
\hline 
{\bf Dataset} & {\bf Train Type} & {\bf Attack Type} & {\bf Attribution method} & {\bf top-$k$} & {\bf 1-LENS-recall@$k$} &  {\bf 1-LENS-prec@$k$} & {\bf top-$k$-div} & {\bf 1-LENS-recall@$k$-div} &  {\bf 1-LENS-prec@$k$-div} \\
\hline
MNIST 
      & Nat & random & Simple Gradient & 0.6548 & 0.8872 & 0.9355 & 0.2431 & 0.8724 & 0.8942 \\ 
      & Nat & random & Image $\times$ Gradient  & 0.6887 & 0.8590 & 0.9791 & 0.1640 & 0.6725 & 0.7735 \\ 
      & Nat & random & LRP [Bach 2015] & 0.6887 & 0.8590 & 0.9791 & 0.1640 & 0.6725 & 0.7733 \\ 
      & Nat & random & DeepLIFT [Shrikumar 2017]  & 0.6896 & 0.8602 & 0.9792 & 0.1642 & 0.6725 & 0.7724 \\ 
      & Nat & random & GradSHAP [Lundberg 2017] & 0.6950 & 0.8629 & 0.9792 & 0.1646 & 0.6716 & 0.7707 \\ 
      & Nat & random & IG [Sundararajan 2017]  & 0.6978 & 0.8636 & 0.9795 & 0.1654 & 0.6717 & 0.7705 \\ 
\hline
MNIST 
      & PGD & random & Simple Gradient  & 0.4456 & 0.7544 & 0.9712 & 0.1798 & 0.8034 & 0.8719 \\ 
      & PGD & random & Image $\times$ Gradient  & 0.5355 & 0.8102 & 0.9853 & 0.1620 & 0.6788 & 0.8037 \\ 
      & PGD & random & LRP [Bach 2015] & 0.6887 & 0.8590 & 0.9791 & 0.2786 & 0.8669 & 0.9557 \\ 
      & PGD & random & DeepLIFT [Shrikumar 2017]  & 0.5387 & 0.8111 & 0.9855 & 0.1626 & 0.6795 & 0.8056 \\ 
      & PGD & random & GradSHAP [Lundberg 2017] & 0.6831 & 0.9164 & 0.9835 & 0.1611 & 0.6684 & 0.7619 \\ 
      & PGD & random & IG [Sundararajan 2017]  & 0.6729 & 0.9359 & 0.9847 & 0.1671 & 0.6641 & 0.7573 \\ 
\hline
Fashion-MNIST 
      & Nat & random & Simple Gradient    & 0.5216 & 0.8421 & 0.8614 & 0.2691 & 0.8718 & 0.8820 \\ 
      & Nat & random & Image $\times$ Gradient   & 0.5840 & 0.9160 & 0.9317 & 0.2570 & 0.8357 & 0.8719 \\ 
      & Nat & random & LRP [Bach 2015]  & 0.5840 & 0.9160 & 0.9317 & 0.2570 & 0.8357 & 0.8719 \\ 
      & Nat & random & DeepLIFT [Shrikumar 2017]    & 0.5821 & 0.9165 & 0.9311 & 0.2560 & 0.8370 & 0.8716 \\ 
      & Nat & random & GradSHAP [Lundberg 2017]   & 0.6075 & 0.9208 & 0.9377 & 0.2641 & 0.8391 & 0.8717 \\ 
      & Nat & random & IG [Sundararajan 2017]    & 0.6279 & 0.9281 & 0.9443 & 0.2736 & 0.8387 & 0.8713 \\ 
\hline
Fashion-MNIST 
      & PGD & random & Simple Gradient    & 0.6036 & 0.8374 & 0.9362 & 0.2974 & 0.8202 & 0.8696 \\ 
      & PGD & random & Image $\times$ Gradient   & 0.6561 & 0.9236 & 0.9648 & 0.3118 & 0.8472 & 0.8820 \\ 
      & PGD & random & LRP [Bach 2015]  & 0.6561 & 0.9236 & 0.9648 & 0.3118 & 0.8472 & 0.8820 \\
      & PGD & random & DeepLIFT [Shrikumar 2017]    & 0.6628 & 0.9255 & 0.9662 & 0.3124 & 0.8496 & 0.8838 \\ 
      & PGD & random & GradSHAP [Lundberg 2017]   & 0.6678 & 0.9428 & 0.9625 & 0.3023 & 0.8573 & 0.8723 \\ 
      & PGD & random & IG [Sundararajan 2017]    & 0.7103 & 0.9638 & 0.9809 & 0.3242 & 0.8583 & 0.8751 \\ 
\hline
\end{tabular}
}
\end{center}
    \caption{Table with top-$k$ intersection and top-$k$-div results for MNIST and Fashion MNIST with LeNet based model trained naturally and adversarially(PGD), using different explanation methods with random sign perturbation. The columns first contain locality results: top-$k$ intersection, 1-LENS-recall@$k$, 1-LENS-prec@$k$ followed by diversity results: top-$k$-div, 3-LENS-recall@$k$-div, 3-LENS-prec@$k$-div.}
\label{table-topk-div-results-rand-mnist-fmnist}

\end{table*}

\begin{table*}[!ht]
\begin{center}
\resizebox{1.0\textwidth}{!}{ 
\begin{tabular}{|l|l|c|c|c|c|c|c|c|c|}
\hline 
{\bf Dataset} & {\bf Train Type} & {\bf Attack Type} & {\bf Attribution method} & {\bf top-$k$} & {\bf 3-LENS-recall@$k$} &  {\bf 3-LENS-prec@$k$} & {\bf top-$k$-div} & {\bf 3-LENS-recall@$k$-div} &  {\bf 3-LENS-prec@$k$-div} \\
\hline
ImageNet 
& Nat & random & Simple Gradient  & 0.3825 & 0.7875 & 0.7949 & 0.3096 & 0.8290 & 0.8115 \\
& Nat & random & Image $\times$ Gradient  & 0.3316 & 0.7765 & 0.8134 & 0.2905 & 0.8655 & 0.8516 \\
& Nat & random & LRP [Bach 2015] & 0.1027 & 0.2487 & 0.2649 & 0.0970 & 0.7518 & 0.7371 \\
& Nat & random & DeepLIFT [Shrikumar 2017]  & 0.2907 & 0.7641 & 0.7637 & 0.2547 & 0.8504 & 0.8578 \\
& Nat & random & GradSHAP [Lundberg 2017] & 0.2290 & 0.6513 & 0.6778 & 0.1885 & 0.8099 & 0.7979 \\
& Nat & random & IG [Sundararajan 2017]  & 0.2638 & 0.7148 & 0.7064 & 0.2366 & 0.8380 & 0.8296 \\
\hline 
ImageNet 
& PGD & random & Simple Gradient  & 0.1725 & 0.7245 & 0.7306 & 0.1410 & 0.8004 & 0.8004 \\
& PGD & random & Image $\times$ Gradient  & 0.1714 & 0.7269 & 0.8043 & 0.1332 & 0.8552 & 0.8687 \\
& PGD & random & LRP [Bach 2015] & 0.2374 & 0.4147 & 0.4350 & 0.1240 & 0.8161 & 0.8210 \\
& PGD & random & DeepLIFT [Shrikumar 2017]  & 0.5572 & 0.9746 & 0.9808 & 0.2924 & 0.8977 & 0.9078 \\
& PGD & random & GradSHAP [Lundberg 2017] & 0.1714 & 0.7270 & 0.8044 & 0.1336 & 0.8552 & 0.8688 \\
& PGD & random & IG [Sundararajan 2017]  & 0.1947 & 0.7335 & 0.8036 & 0.1524 & 0.8584 & 0.8696 \\
\hline
\end{tabular}
}
\end{center}
    \caption{Table with top-$k$ intersection and top-$k$-div results for ImageNet with naturally and adversarially(PGD) trained ResNet50 model using different explanation methods with random sign perturbation. The columns first contain locality results: top-$k$ intersection, 1-LENS-recall@$k$, 1-LENS-prec@$k$ followed by diversity results: top-$k$-div, 3-LENS-recall@$k$-div, 3-LENS-prec@$k$-div.}
\label{table-topk-div-results-rand-imagenet-app}
\end{table*}

\clearpage

\section{Details of Survey Conducted to Study Human's Perception of Robustness} \label{app:survey-human}
Detailed description of the survey conducted to study human's interpretation of attribution maps.

\textit{Survey Format:} Each question consisted of an unperturbed image from the Flower dataset and a pair of explanation/attribution maps. The pair can be any combination of original and attribution map obtained with random perturbation or with \citet{ghorbani2018interpretation} attack. We used Integrated Gradients (IG) \citep{SundararajanTY17} to obtain the attribution maps. Perturbed maps were obtained using the \citet{ghorbani2018interpretation} attack and random noise with appropriate $\epsilon$-budget. The questions were presented at random. At no point in the survey we revealed the type of map or perturbation added to obtain the maps. This ensured the user was not biased by this extra information while answering the survey. A sample question presented to the user is as given below : 

Here is an image of {\bf Tigerlily} with two attribution maps.

\begin{figure}[!ht]
    \centering
    \includegraphics[width=0.48\textwidth]{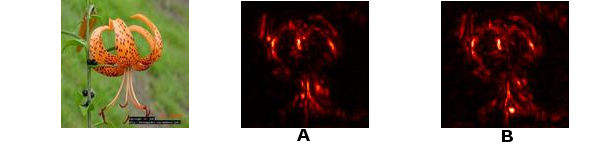}
    \label{fig:survey-sample}
\end{figure}
Do both these attribution maps explain the image well? 
\begin{enumerate} [label=(\arabic*)]
    \item Yes, both are similar and explain the image well.
    \item Yes, both explain the image well, but are dissimilar.
    \item Only A explains the image well, B is different.
    \item Only B explains the image well, A is different.
    \item No, Both the maps do not explain the image well.
\end{enumerate}

\textit{Interpretation of Options:} Options (1), (2) the user is able to relate both the maps to equally represent the image. Option (3) the user finds map A to represent the image over map B. Option (4) the user finds map B relates to the image more closer than map B. Option (5) the user can not relate the maps to the image. 

Table \ref{table-survey-results} in the main paper, we simplify the options by forming 2 categories - \begin{inparaenum}[(1)] \item {\bf Agree with 3-LENS-prec@$k$ metric} : combines Option (1), (2) and (4). These options show that the top-$k$ metric does not match human visual perception when the user either is agnostic to noise in the map or finds the perturbed map more relatable to the image. \item {\bf Agree with top-$k$ metric} : Only Option (3). It shows that top-$k$ metric is sufficient to measure the differences between the attribution maps and closely captures human visual perception.
\end{inparaenum}

\textit{Summary of Survey Results:}
In the above sample question, map {\bf A} is the original IG map of the image and map {\bf B} the IG map of \citet{ghorbani2018interpretation} attacked image with $\epsilon=8/255$. Interestingly the users who chose between (1) to (4), more than 65\% users chose option (1), (2) or (4) (Agree with 3-LENS-prec@$k$ metric), while remaining preferred option (3) (Agree with top-$k$ metric). Showing that despite the top-$k$ intersection being less the 30\% between the maps, users who fall into the {\bf Agree with 3-LENS-prec@$k$ metric} category is large indicating that current top-$k$ based comparison is weak.

\begin{figure}[!ht]
    \centering
    \includegraphics[width=0.48\textwidth]{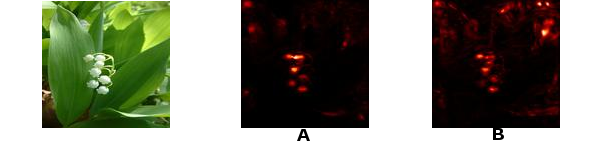}
    \caption{Sample question from survey using {\bf Lily Valley} image with original IG map and IG map with \citet{ghorbani2018interpretation} attack using $\epsilon=8/255$.}
    \label{fig:survey-sample-2}
\end{figure}

In another sample from the survey (Figure \ref{fig:survey-sample-2}), surprisingly more the 50\% who fall in the category {\bf Agree with 3-LENS-prec@k metric} preferred the perturbed map over the original map. top-$k$ and 3-LENS-prec@$k$ values were 36\% and 88\%, respectively.

We did have few questions in the survey to study the effectiveness of random perturbation as an attack as observed by \citet{ghorbani2018interpretation}[Figure 3] with top-$k$ metric. We used $\epsilon=8/255$ for the random perturbation. The results of the survey were very unanimous with users responses overwhelmingly(above 90\%) fell into the {\bf Agree with 3-LENS-prec@$k$ metric}. This strongly indicates that random perturbation considered as an attack under current metrics gives a false sense of attribution robustness. Refer to the last entries in Table \ref{table-survey-results-full}.


\begin{table}[!h]
\begin{center}
\resizebox{0.48\textwidth}{!}{ 
\begin{tabular}{|c|c|c|c|c|}
\hline 
{\bf Attack Type} & {\bf Agree with 3-LENS-prec@$k$ metric(\%)} & {\bf Agree with top-$k$ metric(\%)} & {\bf top-1000 intersection} & {\bf 3-LENS-prec@1000}\\
\hline
top-$k$ & 70.37 & 29.63 & 0.343 & 0.928\\ 
top-$k$ & 81.48 & 18.52 & 0.0805 & 0.521 \\ 
\hline
random & 93.55 & 6.45 & 0.357 & 0.7965 \\ 
\hline
\end{tabular}
}
\end{center}
\caption{Survey results based on humans ability to relate the explanation map to the original image with or without noise using the Flower dataset using Integrated Gradients (IG) \citep{SundararajanTY17} as the explanation method.}
\label{table-survey-results-full}
\end{table}

\end{document}